\documentclass[sigconf,nonacm]{acmart}
\usepackage{multirow}
\usepackage{array} 
\usepackage{subcaption}
\usepackage{amsthm}

\newtheorem{assumption}{Assumption}[section]

\AtBeginDocument{%
  }

\makeatletter
\let\originalfnsymbol\@fnsymbol
\renewcommand{\@fnsymbol}[1]{%
  \ifcase#1
  \or\ensuremath{\dagger}%
  \or\ensuremath{*}%
  \else\originalfnsymbol{#1}%
  \fi}
\makeatother

\begin{document}

\title{Clustering-Friendly Image Representations via Fusion of Global Semantic and Local Relational Embeddings}

\author{Feijiang Li}
\authornote{These authors contributed equally to this work.}
\affiliation{%
  \institution{Key Laboratory of Evolutionary Science Intelligence of Shanxi Province, Shanxi University, Institute of Big Data Science and Industry}
  \city{Taiyuan}
  \state{Shanxi}
  \country{China}
}
\email{fjli@sxu.edu.cn}

\author{Zizheng Jiu}
\authornotemark[1]
\affiliation{%
  \institution{Institute of Big Data Science and Industry, Shanxi University}
  \city{Taiyuan}
  \state{Shanxi}
  \country{China}
}
\email{jiu03699@gmail.com}

\author{Yuhua Qian}
\authornote{Corresponding author.}
\affiliation{%
  \institution{Key Laboratory of Evolutionary Science Intelligence of Shanxi Province, Shanxi University, Institute of Big Data Science and Industry}
  \city{Taiyuan}
  \state{Shanxi}
  \country{China}
}
\email{jinchengqyh@126.com}

\author{Jieting Wang}
\affiliation{%
  \institution{Key Laboratory of Evolutionary Science Intelligence of Shanxi Province, Shanxi University, Institute of Big Data Science and Industry}
  \city{Taiyuan}
  \state{Shanxi}
  \country{China}
}
\email{jtwang@sxu.edu.cn}

\author{Liang Du}
\affiliation{%
  \institution{Key Laboratory of Evolutionary Science Intelligence of Shanxi Province, Shanxi University, Institute of Big Data Science and Industry}
  \city{Taiyuan}
  \state{Shanxi}
  \country{China}
}
\email{duliang@sxu.edu.cn}

\renewcommand{\shortauthors}{Li, Jiu, et al.}

\begin{teaserfigure}
  \centering
  \includegraphics[width=\textwidth]{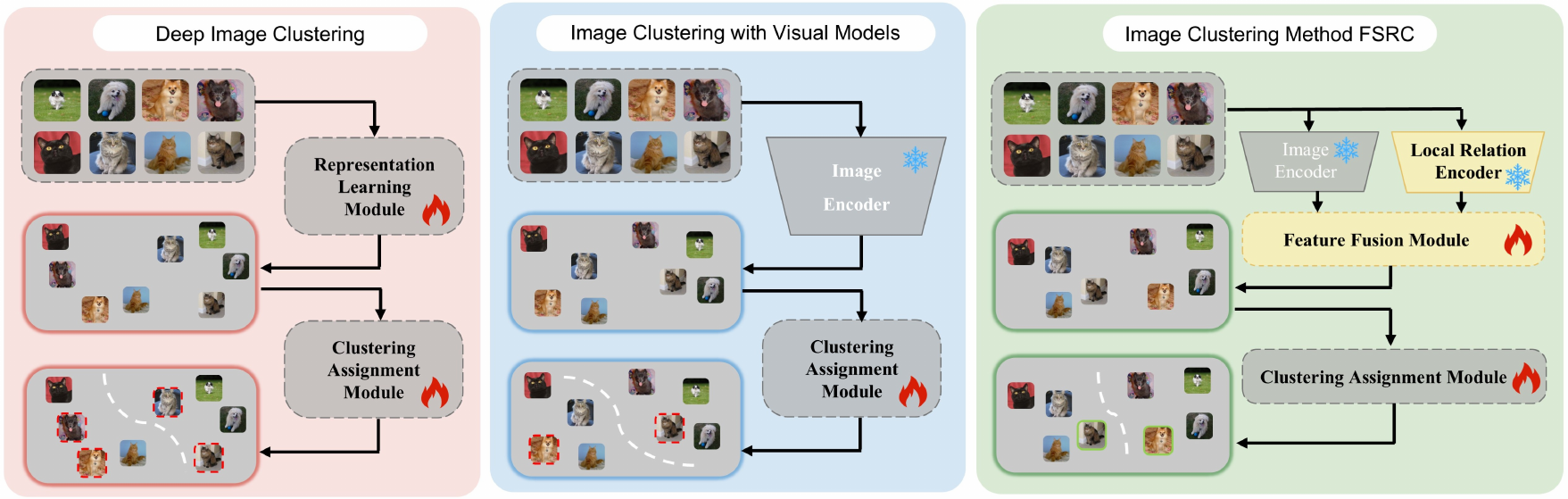}
  \caption{Deep Clustering and FSRC implementation pipelines. 
Left: Conventional deep clustering optimizes the representation learning and clustering assignment modules. 
Middle: Deep clustering with foundation visual models uses a frozen image encoder to extract global semantic features and optimizes the clustering assignment module. 
Right: Built upon the middle pipeline, FSRC introduces a Local Relation Encoder and a Feature Fusion Module to integrate local relation and global semantic embeddings for clustering.}
  \label{fig1}
\end{teaserfigure}

\begin{abstract}
Image–text models such as CLIP provide powerful pre-trained visual features that enable strong zero-shot transfer and have been widely adopted in unsupervised learning. However, for image clustering on specific datasets, representations extracted from foundation models are not always well-suited to the target data distribution, thereby limiting their effectiveness. To address this issue, we propose FSRC, a deep clustering framework that introduces a new paradigm of adapting pretrained representations to unlabeled data structures before optimizing cluster assignments. Specifically, FSRC introduces a Local Relation Encoder (CNN+GAT) to capture relational cues among image patches and generate local relational embeddings, which complement the global semantic embeddings extracted from foundation visual models. These two types of embeddings are integrated through a mutual distillation mechanism that jointly optimizes the global and local clustering heads using distillation, alignment, and entropy losses, promoting consistency and enhancing representational separability. Based on the fused representations, a lightweight clustering classifier is trained with cross-entropy and entropy losses to generate and iteratively refine pseudo-labels, yielding the final clustering assignments. Extensive experiments on 18 benchmark datasets demonstrate that modeling local relational information significantly improves representation separability, as measured by triplet accuracy. FSRC achieves state-of-the-art clustering performance with strong stability across 7 different visual models. 
\textcolor{magenta}{\href{https://zizhjiu.github.io/FSRCPage/}{https://zizhjiu.github.io/FSRCPage/}}.
\end{abstract}

\ccsdesc[500]{Computing methodologies~Cluster analysis}

\keywords{Image Clustering, Unsupervised Learning, Representation Construction}

\maketitle

\begin{figure*}[t]
    \centering
    \begin{subfigure}[b]{0.31\textwidth}
        \centering
        \includegraphics[width=\textwidth]{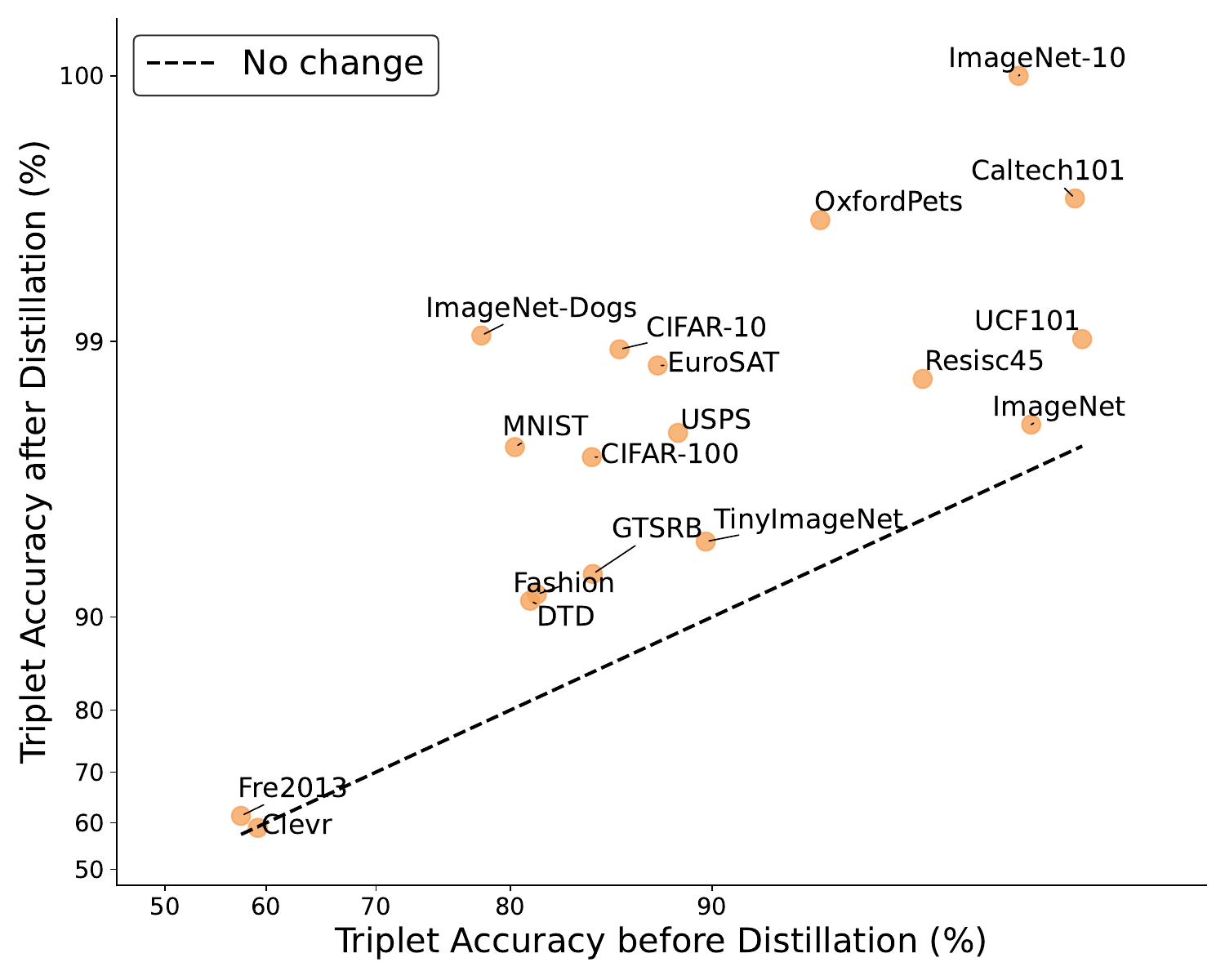}
        \caption{}
    \end{subfigure}\hfill
        \begin{subfigure}[b]{0.31\textwidth}
            \centering
            \includegraphics[width=\textwidth]{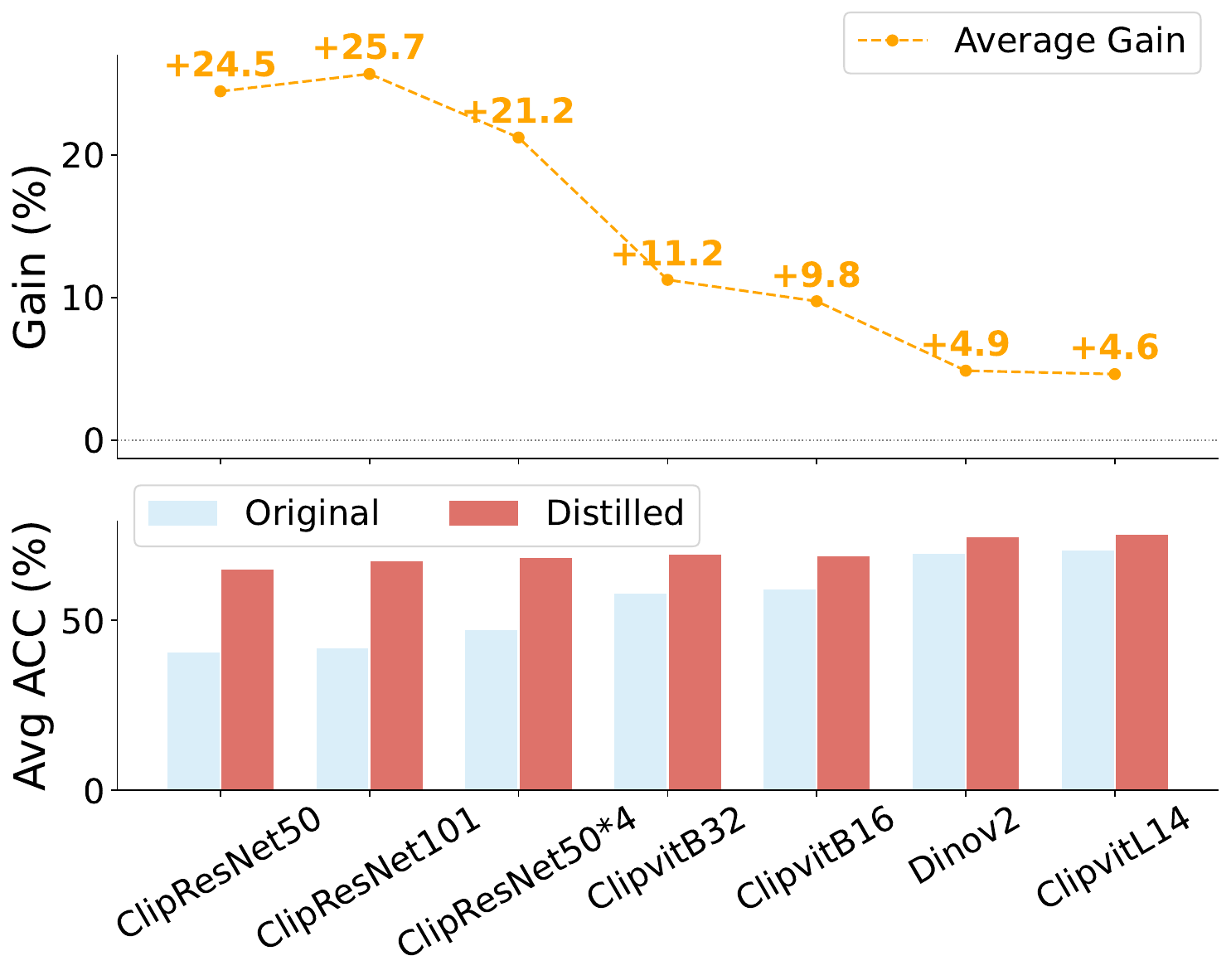}
            \caption{}
        \end{subfigure}\hfill
    \begin{subfigure}[b]{0.31\textwidth}
        \centering
        \includegraphics[width=\textwidth]{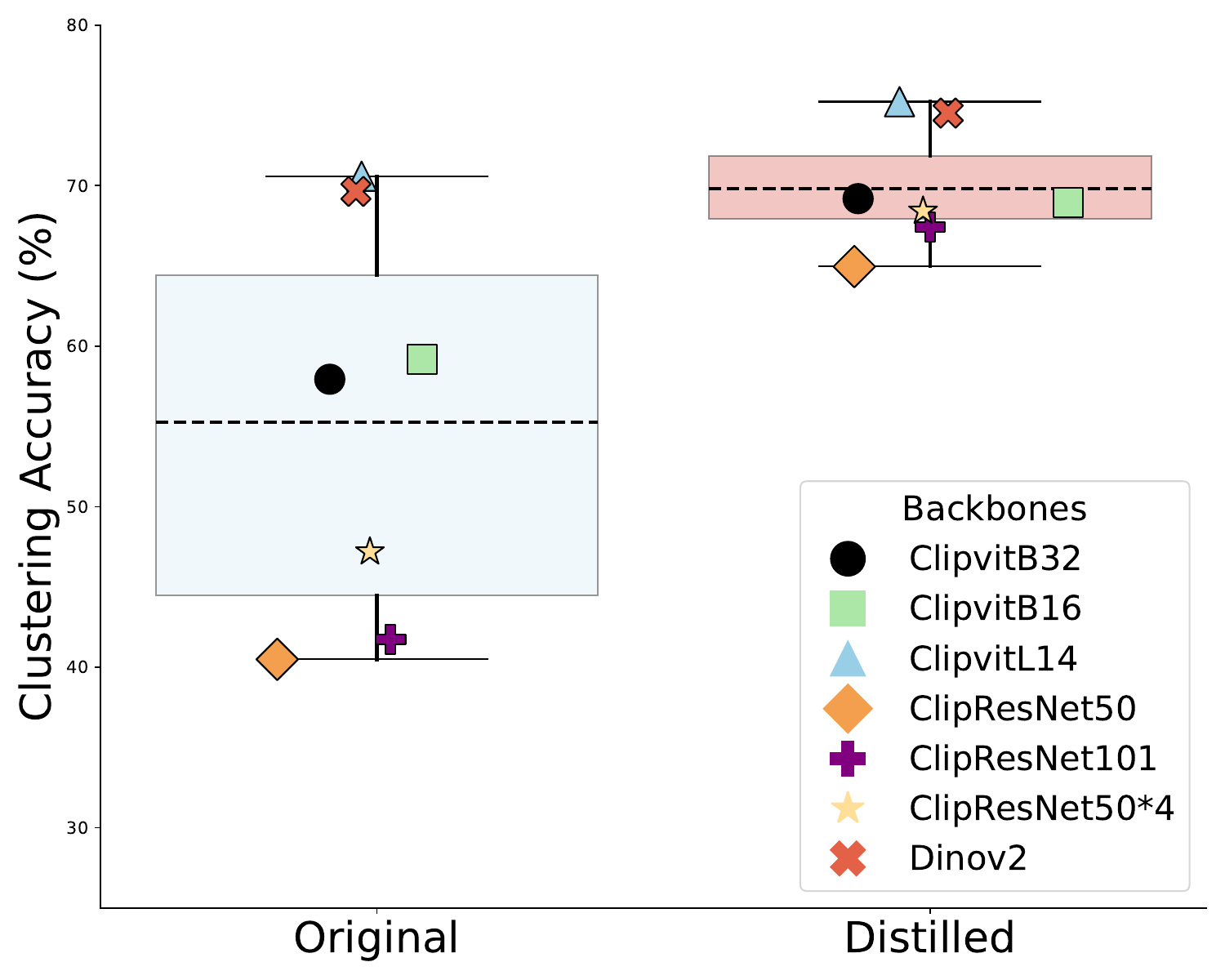}
        \caption{}
    \end{subfigure}
    \caption{Overall performance of FSRC across three key aspects: (a) Triplet accuracy comparison on 18 datasets before and after distillation using ClipVIT-L/14. (b) Effect of Feature Fusion on 7 backbones across 18 datasets. (c) Boxplots depict the distribution of average clustering accuracies across 18 datasets and 7 backbones. Original denotes the original global semantic embeddings, and Distilled denotes the Distilled Representation, which is obtained via mutual distillation between global semantic embeddings and local relational embeddings.}
    \label{fig:combined}
\end{figure*}

\section{Introduction}

Clustering serves as a critical preprocessing step in numerous data mining tasks \citep{zhou2024comprehensive}. Traditional clustering methods assume that data reside in a well-structured feature space, but this assumption often fails for unstructured, high-dimensional data such as images. Conventional deep image clustering addresses this limitation by jointly learning feature representations and clustering assignments in an end-to-end fashion, enabling effective grouping directly from raw data without requiring handcrafted features or human annotations. As illustrated in the left part of Figure~\ref{fig1}, the deep image clustering methods typically optimize the representation learning module and the clustering classifier module jointly. These optimization strategies can be sequential \citep{tao2021clustering}, iterative \citep{niu2022spice}, generative \citep{ji2021decoder}, or simultaneous \citep{tsai2020mice, caron2020unsupervised}. However, without strong semantic priors, the methods are vulnerable to unstable pseudo-labels.

Recently, driven by the increasing availability of large-scale datasets, a variety of foundation models have been developed. Among them, CLIP \citep{radford2021learning} has emerged as a prominent model that learns transferable global semantic representations through contrastive language–image pre-training. Due to its strong generalization capability and inherent semantic alignment, CLIP has been widely adopted as a backbone for image clustering \citep{li2023image,jiu2026spatial}. In this paradigm, as depicted in the middle part of Figure~\ref{fig1}, the pre-trained image encoder is kept frozen to serve as a fixed feature extractor, while only the downstream clustering assignment module is fine-tuned or optimized based on the extracted global semantic features. This pipeline has significantly simplified the training process and contributed to notable advances in deep clustering.

However, despite its strong semantic priors, CLIP is not explicitly optimized for clustering objectives. Consequently, its representations may not align with the intrinsic grouping structure of specific datasets, limiting clustering performance. This limitation is particularly evident in fine-grained datasets, where visually similar categories often differ in subtle local relations rather than global semantics. Therefore, we introduce a new clustering paradigm that first adapts pretrained features to the intrinsic relational structure of unlabeled datasets and subsequently optimizes cluster assignments.

To realize this paradigm, we propose FSRC (Fusing Global Semantic and Local Relational Representations for Deep Clustering), which adapts foundation-model features through local relational modeling and mutual distillation before optimizing cluster assignments.
As illustrated in Figure~\ref{fig1} (right), FSRC first introduces a Local Relation Encoder to capture relational cues among image patches. These patch-level relations provide complementary fine-grained structural cues. For example, visually similar categories may differ in the relative positions of local parts, such as the head and ears across different dog breeds. By modeling such local relational patterns, the Local Relation Encoder produces more clustering-friendly representations. Subsequently, a Mutual Distillation Module integrates these global semantic and local relational features to produce refined distilled representations. To quantify the improvement, we employ triplet accuracy as an intrinsic metric to evaluate the quality of clustering-friendly representations (see Appendix \ref{A} for details). As demonstrated in Figure~\ref{fig:combined} (a), the distilled features achieve substantially higher triplet accuracy, confirming enhanced embedding separability. This improvement directly facilitates downstream clustering assignment and helps discover a more reasonable grouping structure.

To uncover the underlying grouping structure without supervision, inspired by \citet{gadetsky2024let}, we employ a lightweight clustering classifier optimized via iterative self-labeling. This design enables the generation of stable pseudo-labels that converge reliably despite its simplicity. Upon convergence, the classifier directly outputs the final cluster assignments. We evaluate FSRC extensively across 18 benchmark datasets (Table \ref{tab:dataset_summary} in Appendix \ref{B1})  using 7 foundation visual models (Table \ref{tab2} in Appendix \ref{B2}). As shown in Figure~\ref{fig:combined} (b), FSRC consistently improves performance across all backbone architectures, with the detailed results provided in Appendix~\ref{H}. Furthermore, Figure~\ref{fig:combined} (c) demonstrates that FSRC achieves more stable performance, with the detailed quantitative results provided in Appendix~\ref{G}. These results collectively confirm the robustness of the proposed clustering framework.

The main contributions are as follows:
\begin{itemize}
    \item We propose a Local Relation Encoder coupled with a Mutual Distillation Module to explicitly integrate fine-grained local relational cues from image patches into foundation model features. Furthermore, we introduce triplet accuracy as a principled metric for evaluating the clustering suitability of learned representations, which strongly correlates with, and reflects, enhanced separability.

    \item We present FSRC, an image clustering framework that 
    introduces a new paradigm that adapts pretrained representations to unlabeled data structures before optimizing cluster assignments. Extensive experiments demonstrate that FSRC achieves state-of-the-art clustering performance on benchmark datasets while exhibiting excellent stability across 7 foundation visual models.
\end{itemize}

\section{Related Work}
\label{gen_inst}

\subsection{Deep Image Clustering}

Deep image clustering aims to group unlabeled images into semantically meaningful clusters by jointly learning feature representations and cluster assignments in an end-to-end manner. Existing deep clustering methods typically optimize a representation-learning module together with a clustering assignment, assuming that samples from the same category should be proximate in the embedding space. Representative approaches such as Deep Embedded Clustering (DEC) \citep{xie2016unsupervised} iteratively refine cluster assignments by minimizing the Kullback–Leibler (KL) divergence between the predicted cluster distribution and a sharpened target distribution. Prediction-consistency-based methods, including Invariant Information Clustering (IIC) \citep{Ji_2019_ICCV}, SCAN \citep{van2020scan}, and Contrastive Clustering (CC) \citep{li2021contrastive}, leverage consistency across predictions under strong data augmentation to facilitate unsupervised clustering. Another line of work adopts iterative pseudo-labeling strategies, in which cluster assignments are repeatedly generated and used as supervisory signals to update representations, as exemplified by DeepCluster \citep{caron2018deep} and SPICE \citep{niu2022spice}. To further capture structural relationships among samples, graph-based extensions such as GATCluster \citep{niu2020gatcluster} incorporate graph attention mechanisms to model neighborhood dependencies, thereby improving clustering quality. Clustering ensemble methods also benefit from effective modeling of sample relationships~\cite{Li2026,li2023fuzzy,li2025k}. 

With the rapid advancement of large-scale visual foundation models, recent studies have revisited deep image clustering from a new perspective. In particular, models such as CLIP \citep{radford2021learning}, pre-trained on massive image-text pairs, can extract highly generalizable and semantically aligned global visual representations. Leveraging these strong priors, a growing body of work \citep{li2023image,gadetsky2024let} has adopted a paradigm in which the image encoder is kept frozen as a powerful feature extractor, while only the clustering assignment module is optimized over the extracted embeddings. Compared to conventional deep clustering methods that jointly learn representations and clustering objectives, this paradigm significantly simplifies training and improves robustness across diverse datasets. However, since foundation models like CLIP are not explicitly optimized for clustering tasks, their representations may not align well with the intrinsic grouping structure of specific datasets. This misalignment can limit the effectiveness of clustering and underscores the need to enhance clustering-friendly representations.

\subsection{Feature Fusion}

Feature fusion aims to integrate multiple feature representations that capture complementary information into a unified embedding, thereby enhancing representation expressiveness and task suitability \citep{baltruvsaitis2018multimodal}. The importance of exploiting consistency and integrating complementary information has also been demonstrated in multi-view clustering~\cite{Li2025,10.1145/3770854.3780182} and domain adaptation~\cite{Li2026TMM}. Early fusion methods primarily adopt simple aggregation strategies, such as feature concatenation, to combine features extracted from different network layers or branches \citep{hariharan2015hypercolumns}. While computationally efficient, these approaches treat all features equally and lack adaptive mechanisms to model interactions, often limiting their effectiveness in complex visual tasks.

To address these limitations, subsequent work has introduced learnable fusion mechanisms, including multi-branch architectures and attention-based designs, that dynamically weight and integrate features \citep{lin2017feature,wang2018non}. More recently, knowledge distillation has emerged as a principled alternative for feature fusion, transferring complementary information by enforcing consistency in the feature or prediction space \citep{hinton2015distilling,tian2019contrastive}. This paradigm has been further extended to mutual distillation, enabling bidirectional knowledge exchange between heterogeneous representations and leading to more balanced and effective fusion \citep{zhang2018deep,anil2018large}. In the context of foundation models such as CLIP \citep{radford2021learning}, distillation-based fusion offers a promising approach for integrating strong global semantic priors with complementary local relational cues. However, such fusion strategies have rarely been explored in unsupervised clustering settings, motivating our FSRC framework to fuse global semantic and local relational features via a clustering-oriented mutual distillation mechanism.

\section{The Proposed Method}

\subsection{Problem Statement and Objective}

In this paper, we consider the image clustering problem: given an unlabeled image dataset 
\(\mathcal{D} = \{x_n\}_{n=1}^{N}\) with a known number of categories \(K\), the goal is to learn a mapping 
\(f: \mathcal{D} \rightarrow \{1, \dots, K\}\) that assigns each image to a cluster.

\begin{figure}[t]
  \centering
  \includegraphics[width=1.0\linewidth]{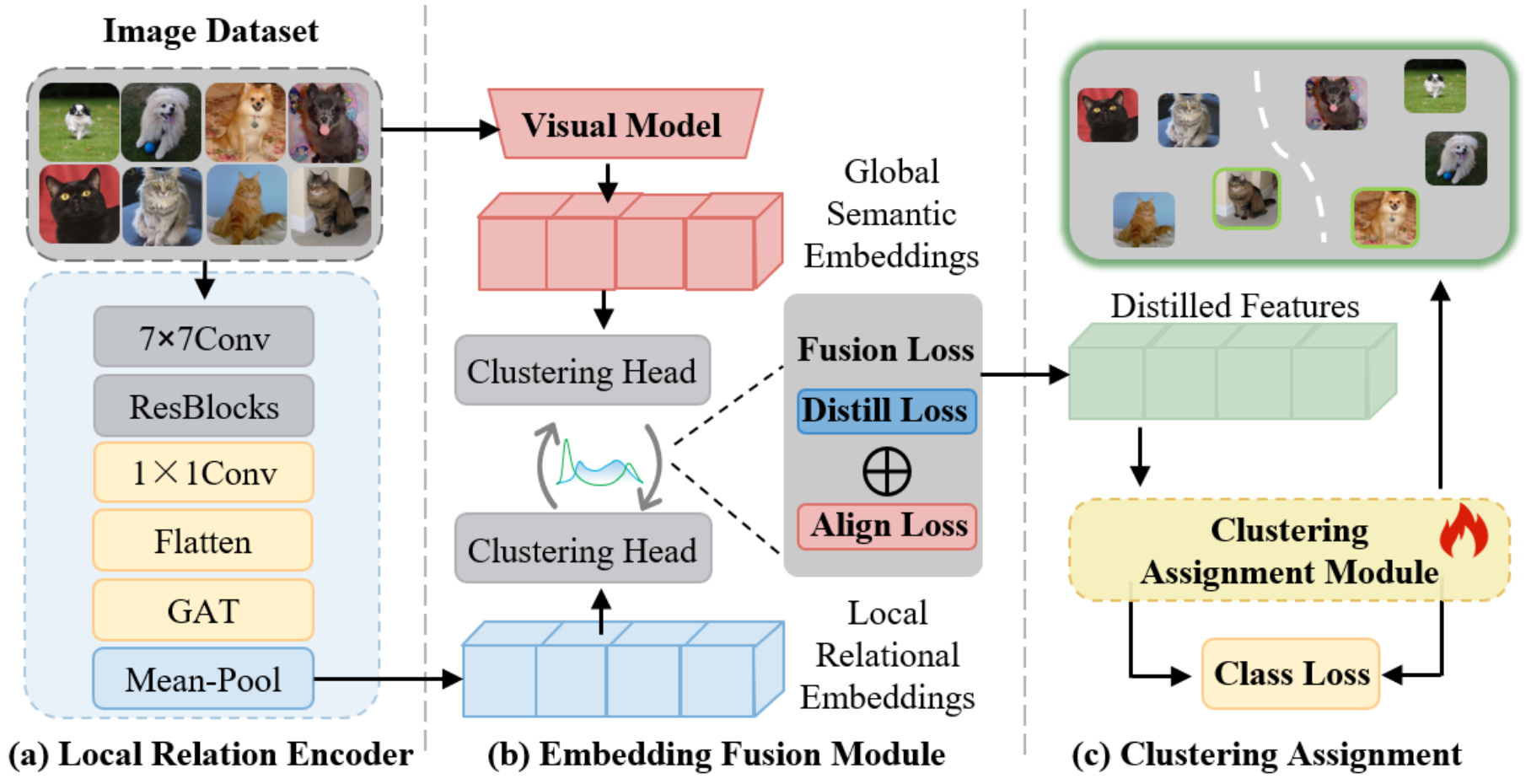}
  \caption{
    Overview of the proposed FSRC framework.
    }
  \label{fig3}
\end{figure}

We address this problem by fusing global semantic and local relational embeddings. Each image \(x\) is represented by a pre-trained visual embedding \(\phi(x) \in \mathbb{R}^{d}\), which is enhanced by a local relational embedding \(g(x)\) through a Mutual Distillation Module, producing a distilled feature $\hat{\phi}(x)$ from the trained global semantic clustering head. Subsequently, \(\hat{\phi}(x)\) is fed into a lightweight clustering assignment module that iteratively refines pseudo-labels, thereby optimizing the final cluster assignments based on the inferred distribution \(\tau_\theta(x)\). An overview of FSRC is shown in Fig.~\ref{fig3}.

Formally, we introduce \(\mathcal{R}\) as an abstract mapping that characterizes the feature update process induced by mutual distillation, integrating \(\phi(x)\) and \(g(x)\) to obtain the distilled feature \(\hat{\phi}(x) = \mathcal{R}(\phi(x), g(x))\). Then, an assignment \(f(x) = w^\top \hat{\phi}(x)\) is trained on $\hat{\phi}(x)$. The overall optimization problem is formulated as:

\begin{align}\label{st1}
\min_{\theta} \mathcal{L}(\theta) = 
\sum_{x \in \mathcal{D}} \mathcal{L}_{ce}\bigl(f(x), \tau_\theta(x)\bigr) - \lambda \mathcal{L}_{ent}(\tau_{\theta}(x)),
\end{align}
\begin{align}\label{st}
\text{s.t.} \quad
\begin{cases}
\text{Embedding Fusion:}\\[4pt]
\quad \hat{\phi}(x) = \mathcal{R}(\phi(x), g(x)),\\[4pt]
\quad \mathcal{R} = \displaystyle \arg\min_{\mathcal{R}'} \sum_{x \in \mathcal{D}}\mathcal{L}_{\mathrm{Fus}}\bigl(\phi(x), g(x); \mathcal{R}'\bigr);\\[6pt]
\text{Clustering Assignment:}\\[4pt]
\quad f(x) = w^\top \hat{\phi}(x),\\[4pt]
\quad w = \displaystyle \arg\min_{w'} \sum_{x \in \mathcal{D}}\mathcal{L}_{\mathrm{Clu}}\bigl(w'^\top \hat{\phi}(x), \tau_\theta(x); w'\bigr).
\end{cases}
\end{align}
Here, \(\lambda\) is a weighting factor for the entropy loss of \(\tau_\theta(x)\), where $\tau_\theta(x)$ denotes the pseudo-label distribution of sample $x$, and $\theta$ represents the learnable parameters used to generate this distribution. \(\mathcal{R}'\) and \(w'\) are auxiliary variables representing candidate solutions during optimization.

\subsection{Embedding Fusion}

\subsubsection{Local Relation Encoder}

While visual models like CLIP excel at extracting global semantic embeddings, their architectures often fail to fully capture the local relational dependencies among image patches, which are critical for clustering tasks. These tasks frequently require distinguishing between visually similar objects based on subtle differences in their spatial configurations. Without capturing these local dependencies, clustering algorithms may struggle to group similar objects effectively. To address this limitation, we introduce a Local Relation Encoder to capture relational dependencies among image patches. The encoder consists of a CNN component initialized from a ResNet-50~\citep{tomasev2022pushing} backbone pretrained on ImageNet-1K, followed by a $7\times7$ convolutional layer, several residual blocks to capture local visual patterns, a $1\times1$ convolution to project features into a compact embedding space, and a Graph Attention Network (GAT)~\citep{velivckovic2017graph} that models local relational dependencies on semantically meaningful patch-level representations.

Formally, given an image \(x \in \mathbb{R}^{3 \times H \times W}\) from the dataset \(\mathcal{D}\), we first extract dense spatial representations using a stack of convolutional layers followed by residual blocks. This convolutional residual network preserves the spatial layout of the image and produces a high-dimensional feature map of size \(C \times H' \times W'\), where \(H'\) and \(W'\) denote the spatial resolution after downsampling and \(C\) is the channel dimension.

To obtain a compact and clustering-friendly representation, we apply a \(1 \times 1\) convolution to project the high-dimensional features from \(C\) to \(d\), where \(d\) matches the embedding dimensionality of the CLIP visual encoder. This results in a feature map of size \(d \times H' \times W'\).

We then treat each spatial location in the projected feature map as a node and flatten the feature map to construct node embeddings \(X = [x_{1}, \ldots, x_{\hat{N}}]^{\top} \in \mathbb{R}^{\hat{N} \times d}\), where \(\hat{N} = H' \times W'\) denotes the number of nodes.

Next, we construct a graph \(G = (V, E)\) based on an 8-neighborhood spatial grid, connecting each node to its spatially adjacent neighbors. This sparse connectivity preserves local structure while avoiding the quadratic complexity of fully connected graphs. Further comparisons with non-local modeling are provided in Appendix~\ref{N}.

The node embeddings \(X\) and edge set \(E\) are then fed into a Graph Attention Network (GAT) to perform relational feature propagation. For each node \(i\), its refined representation is computed as:
\begin{equation}
\label{eq:gat}
\begin{aligned}
x_i' &= \sum_{j \in \mathcal{N}(i)} \alpha_{ij} W x_j, \\
\alpha_{ij} &=
\frac{\exp\left(\mathrm{LeakyReLU}\left(a^{\top} \left[W x_i \,\|\, W x_j\right]\right)\right)}
{\sum_{k \in \mathcal{N}(i)} \exp\left(\mathrm{LeakyReLU}\left(a^{\top} \left[W x_i \,\|\, W x_k\right]\right)\right)},
\end{aligned}
\end{equation}
where \(\mathcal{N}(i)\) denotes the neighbors of node \(i\) defined by the 8-neighborhood graph, \(W\) and \(a\) are learnable parameters of the GAT, and \(\|\) denotes vector concatenation.

After refinement, we aggregate all node representations via global average pooling to obtain the local relational embedding:
\begin{equation}
\label{eq:spatial_embedding}
g(x) = \frac{1}{\hat{N}} \sum_{i=1}^{\hat{N}} x_i' \in \mathbb{R}^{d}.
\end{equation}

\subsubsection{Mutual Distillation}

We perform mutual distillation by fusing the local relational embedding $g(x)$ extracted by the Local Relation Encoder with the global semantic embedding $\phi(x)$. This fusion is conducted from the perspective of
underlying structural alignment, enabling effective integration of fine-grained
local relations and global semantic information, and yielding a refined
distilled representation that is highly amenable to clustering.

Given an input image \(x\), we first pass \(\phi(x)\) and \(g(x)\) through separate clustering heads:
\begin{equation}
p_I = h_I(\phi(x)), \qquad
p_S = h_S(g(x)),
\end{equation}
where \(h_I\) and \(h_S\) are MLP-based clustering heads for the global semantic and local relational embeddings, respectively. The distilled feature \(\hat{\phi}(x)\), obtained from the penultimate hidden layer of the global semantic clustering head after mutual distillation:
\begin{align}
\hat{\phi}(x) = \mathcal{R}(\phi(x), g(x)), \quad
\mathcal{R} = \arg\min_{\mathcal{R}'}  \sum_{x \in \mathcal{D}}\mathcal{L}_{\mathrm{Fus}}\bigl(\phi(x), g(x); \mathcal{R}'\bigr).
\end{align}
The mutual distillation loss \(\mathcal{L}_{\mathrm{Fus}}\) comprises three components:
\begin{equation}
\mathcal{L}_{\mathrm{Fus}} = \mathcal{L}_{\text{Distill}} + \mathcal{L}_{\text{Align}} - \lambda \mathcal{L}_{\text{Entropy}},
\end{equation}
where $\lambda$ is the weight of the entropy loss.

The distillation loss encourages mutual learning between the global and local features:
\begin{align}
\mathcal{L}_{\text{Distill}}
&= \mathcal{L}_{S \rightarrow I} + \mathcal{L}_{I \rightarrow S},\\
\mathcal{L}_{S \rightarrow I}
&= -\frac{1}{K}\sum_{k=1}^{K}
\log \frac{
e^{\mathrm{sim}(p_{I,k},p_{S,k}^{\mathcal{N}})/T}
}{
\sum_{j=1}^{K}e^{\mathrm{sim}(p_{I,k},p_{S,j}^{\mathcal{N}})/T}
+\sum_{j\neq k}e^{\mathrm{sim}(p_{I,k},p_{I,j})/T}
},\\
\mathcal{L}_{I \rightarrow S}
&= -\frac{1}{K}\sum_{k=1}^{K}
\log \frac{
e^{\mathrm{sim}(p_{S,k},p_{I,k}^{\mathcal{N}})/T}
}{
\sum_{j=1}^{K}e^{\mathrm{sim}(p_{S,k},p_{I,j}^{\mathcal{N}})/T}
+\sum_{j\neq k}e^{\mathrm{sim}(p_{S,k},p_{S,j})/T}
},
\end{align}
where $K$ is the number of clusters, and $\mathcal{N}$ indicates randomly sampled neighbors from precomputed top-50 neighborhoods. $T$ is the temperature parameter, and $\mathrm{sim}(\cdot,\cdot)$ denotes cosine similarity.

The alignment loss enforces consistency between the two predictions:
\begin{equation}
\mathcal{L}_{\text{Align}} = - \frac{1}{N} \sum_{i=1}^N \log \left( p_I^{i\top} p_S^i \right).
\end{equation}

The entropy loss prevents clustering collapse:
\begin{align}
\mathcal{L}_{\text{Entropy}} &= -\sum_{i=1}^{K}\left(\bar{p}_i \log \bar{p}_i + \bar{q}_i \log \bar{q}_i\right), \\
\bar{p} &= \frac{1}{N}\sum_{i=1}^{N}p_I^i \in \mathbb{R}^K, \quad \bar{q} = \frac{1}{N}\sum_{i=1}^{N}p_S^i \in \mathbb{R}^K,
\end{align}
where \(\bar{p}\) and \(\bar{q}\) denote the average cluster assignment distributions from the two clustering heads.

\subsection{Clustering Assignment}

The distilled feature \(\hat{\phi}(x)\) is first obtained after optimizing the embedding fusion module through mutual distillation. The embedding fusion module is then frozen, and the \(\hat{\phi}(x)\) are used to train a clustering assignment module. The objective is to refine the pseudo-labels \(\tau_\theta(x)\) by minimizing the classification loss \(\mathcal{L}_{\text{Class}}\), which encourages consistent cluster assignments. Specifically, we define a classifier \(f(x) = w^\top \hat{\phi}(x)\) with parameters learned as:
\begin{equation}
w = \arg\min_{w'} \mathcal{L}_{\text{Class}}\bigl(w'^\top \hat{\phi}(x), \tau_\theta(x); w'\bigr),
\end{equation}
where \(\mathcal{L}_{\text{Class}}\) is defined as:
\begin{align}
\mathcal{L}_{\text{Class}} &= -\frac{1}{N} \sum_{i=1}^{N} \sum_{k=1}^{K} \tau_{\theta}^{k}(x_i) \log \hat{y}^{k}(x_i) + \lambda \sum_{k=1}^{K} \bar{\tau}_k \log \bar{\tau}_k,\\[4pt]
\hat{y}^{k}(x_i) &= \operatorname{softmax}\bigl(w^\top \hat{\phi}(x_i)\bigr)_k, \qquad 
\bar{\tau} = \frac{1}{N} \sum_{i=1}^{N} \tau_\theta(x_i) \in \mathbb{R}^K,
\end{align}
where $\bar{\tau}$ denotes the average pseudo-label distribution. After training converges, the module output \(f(x)\) provides the final cluster assignments for each input image.

\section{Theoretical Analysis}

We provide a theoretical justification showing that mutual distillation between \(\phi(x)\) and \(g(x)\) produces a fused representation \(\hat{\phi}(x)\) which could improve clustering performance.

\begin{definition}
Let the training data consist of \(N\) i.i.d. samples \((X_i, Y_i) \sim \mathcal{P}_{XY}\), where \(Y_i\) denotes the class labels (unavailable during training), and each \(X_i\) is an image. Each image \(X_i\) is mapped into two distinct representations: a global semantic feature \(\phi(X_i) \in \mathbb{R}^d\) and a local relational feature \(g(X_i) \in \mathbb{R}^d\). These are fused via mutual distillation into a distilled representation \(\hat{\phi}(X_i) = \mathcal{R}(\phi(X_i), g(X_i))\).

Given a proximity threshold \(\tau > 0\), we define the following proximity events for each representation:
\begin{align}
\mathcal{E}_V(\tau) &= \{\| \phi(X_i) - \phi(X_j) \| \leq \tau \}, \\
\mathcal{E}_S(\tau) &= \{\| g(X_i) - g(X_j) \| \leq \tau \}, \\
\mathcal{E}_Z(\tau) &= \{\| \hat{\phi}(X_i) - \hat{\phi}(X_j) \| \leq \tau \},
\end{align}
where \(\|\cdot\|\) denotes the Euclidean norm. These events correspond to pairs of images that are within distance \(\tau\) in their respective feature spaces.
\end{definition}

\begin{assumption}[Partial Discriminability]
\label{assump:weak-discriminability}
Each modality provides partial class information. Formally, there exist constants \(\rho_V, \rho_S \in (0,1)\) such that for any pair of samples from different classes:
\begin{equation}
\mathbb{P}(\mathcal{E}_V(\tau) \mid Y_i \neq Y_j) \leq \rho_V, \quad
\mathbb{P}(\mathcal{E}_S(\tau) \mid Y_i \neq Y_j) \leq \rho_S.
\end{equation}
\end{assumption}

\begin{assumption}[Approximate Complementarity]
\label{assump:weak-complementary}
Due to the distinct modeling mechanisms of the two representations, their errors are not perfectly correlated. Specifically, the global semantic feature $\phi(\cdot)$ focuses on global semantic information, while the local relational feature $g(\cdot)$ captures local relations. As a result, they tend to make different types of errors, leading to approximately complementary behaviors.

Formally, there exists a constant $\eta \in (0,1)$ controlling the degree of correlation between their errors. Because of the complementary modeling mechanisms described above, this correlation is weak, so $\eta$ is expected to be close to zero. Consequently, for $Y_i \neq Y_j$, the following holds:
\begin{align}
&\mathbb{P}(\mathcal{E}_V(\tau)\cap
\mathcal{E}_S(\tau)\mid Y_i\neq Y_j)
\notag\\
&\quad\le
\mathbb{P}(\mathcal{E}_V(\tau)\mid Y_i\neq Y_j)
\mathbb{P}(\mathcal{E}_S(\tau)\mid Y_i\neq Y_j)
+\eta.
\end{align}
\end{assumption}

The two assumptions indicate that each modality provides useful but imperfect information, and the errors made by each modality are not perfectly aligned.

\begin{theorem}[Inter-Class Confusion Reduction]
\label{thm:simplified}
Under Assumptions~\ref{assump:weak-discriminability}--\ref{assump:weak-complementary} and the condition $\eta < \min\{\rho_V(1-\rho_S), \ \rho_S(1-\rho_V)\}$, the distilled feature yields an inter-class confusion probability satisfying:
\begin{equation}
\mathbb{P}(\mathcal{E}_Z(\tau) \mid Y_i \neq Y_j)
\leq \rho_V \rho_S + \eta
<
\min\{\rho_V, \rho_S\}.
\end{equation}
\end{theorem}

\begin{theorem}[Preservation of Useful Structure]
\label{thm:simplified-intra}
The fused embedding \(\hat{\phi}(x)\) preserves intra-class proximity at least as well as the better of the two individual modalities:
\begin{align}
&\mathbb{P}(\mathcal{E}_Z(\tau) \mid Y_i = Y_j) \\ \notag
\geq &
\max\Bigl\{
\mathbb{P}(\mathcal{E}_V(\tau) \mid Y_i = Y_j),\;
\mathbb{P}(\mathcal{E}_S(\tau) \mid Y_i = Y_j)
\Bigr\} - \delta,
\end{align}
where \(\delta \geq 0\) is a small constant arising from the fusion process.

\end{theorem}

Theorem~\ref{thm:simplified} shows that mutual distillation reduces inter-class confusion, while Theorem~\ref{thm:simplified-intra} ensures that intra-class structure is preserved. They demonstrate that the fused representation improves cluster separability. Complete proofs and empirical validations are in Appendix~\ref{C}.

\begin{table*}[t]
  \centering
  \small
  \renewcommand{\arraystretch}{1.02}
  \caption{Clustering performance of different approaches evaluated by ACC\%, NMI\%, and ARI\%. ``--'' indicates unavailable results. \textbf{Bold} numbers indicate the best performance across all methods. \underline{Underlined} numbers indicate the second-best performance.}
\begin{tabular*}{\hsize}{@{\extracolsep{\fill}}clcccccccccccc}
\toprule
\multirow{2}{*}{\textbf{Category}} &
\multirow{2}{*}{\textbf{Method}} &
\multicolumn{3}{c}{ImageNet-10} &
\multicolumn{3}{c}{ImageNet-Dogs} &
\multicolumn{3}{c}{CIFAR-100} &
\multicolumn{3}{c}{CIFAR-10} \\
\cmidrule(lr){3-5}
\cmidrule(lr){6-8}
\cmidrule(lr){9-11}
\cmidrule(lr){12-14}
&& ACC & NMI & ARI
& ACC & NMI & ARI
& ACC & NMI & ARI
& ACC & NMI & ARI \\
\midrule

\multirow{10}{*}{\shortstack[c]{Deep\\Image\\Clustering}}

& DSEC (PAMI18) \citep{chang2018deep}
& 67.4 & 58.3 & 52.2
& 26.4 & 23.6 & 12.4
& 25.5 & 21.2 & 11.0
& 47.8 & 43.8 & 34.0 \\

& DCCM (ICCV19) \citep{wu2019deep}
& 71.0 & 60.8 & 55.5
& 38.3 & 32.1 & 18.2
& 32.7 & 28.5 & 17.3
& 62.3 & 49.6 & 40.8 \\

& GATC (ECCV20) \citep{niu2020gatcluster}
& 76.2 & 60.9 & 57.2
& 33.3 & 32.2 & 20.0
& 28.1 & 21.5 & 11.6
& 61.0 & 47.5 & 40.2 \\

& PICA (CVPR20) \citep{huang2020deep}
& 87.0 & 80.2 & 76.1
& 35.2 & 35.2 & 20.1
& 33.7 & 31.0 & 17.1
& 69.6 & 59.1 & 51.2 \\

& IDFD (ICLR21) \citep{tao2021clustering}
& 95.4 & 89.8 & 90.1
& 59.1 & 54.6 & 41.3
& 42.5 & 42.6 & 26.4
& 81.5 & 71.1 & 66.3 \\

& CC (AAAI21) \citep{li2021contrastive}
& 89.3 & 85.9 & 82.2
& 42.9 & 44.5 & 27.4
& 42.9 & 43.1 & 26.6
& 79.0 & 70.5 & 63.7 \\

& TCL (IJCV22) \citep{li2022twin}
& 89.5 & 87.5 & 83.7
& 64.4 & 62.3 & 51.6
& 53.1 & 52.9 & 35.7
& 88.7 & 81.9 & 78.0 \\

& ConCUR (ML22) \citep{deshmukh2022representation}
& 95.8 & 90.7 & 90.9
& 69.5 & 63.0 & 53.1
& 47.9 & 46.8 & 30.3
& 84.6 & 76.2 & 71.5 \\

& SPICE (TIP22) \citep{niu2022spice}
& 96.9 & 92.7 & 93.3
& 67.5 & 62.7 & 52.6
& 58.4 & 58.3 & 42.2
& 91.8 & 85.0 & 83.6 \\

& DPAC (ECCV24) \citep{yan2024deep}
& 97.0 & 92.5 & 93.5
& 72.6 & 66.7 & 59.8
& 55.5 & 54.2 & 39.3
& 93.4 & 87.0 & 86.6 \\

\midrule

\multirow{9}{*}{\shortstack[c]{Clustering\\with\\Visual\\Models}}

& SIC (AAAI23) \citep{cai2023semantic}
& 98.2 & 97.0 & 96.1
& 69.7 & 69.0 & 55.8
& 58.3 & 59.3 & 43.9
& 92.6 & 84.7 & 84.4 \\

& TAC (ICML24) \citep{li2023image}
& 99.5 & 98.5 & 98.8
& 84.4 & 77.4 & 72.0
& 59.0 & 68.6 & 42.7
& 92.2 & 83.7 & 83.6 \\

& TURTLE (ICML24) \citep{gadetsky2024let}
& \underline{99.8} & 99.3 & 99.5
& 74.4 & 71.0 & 62.5
& 63.4 & 71.9 & 51.9
& \underline{97.6} & \underline{93.6} & 94.0 \\

& EC (Info.Sci25) \citep{li2025exploring}
& 98.9 & 96.9 & 97.5
& 54.9 & 53.1 & 37.3
& 56.2 & 58.6 & 41.3
& 88.1 & 81.8 & 77.8 \\

& GradNorm (ICCV25) \citep{peng2025provable}
& 99.4 & 98.7 & 98.7
& 81.2 & 81.0 & 70.9
& 60.6 & 61.3 & 43.6
& 91.1 & 82.6 & 81.5 \\

& CDC-Cal (ICLR25) \citep{jia2025towards}
& 97.3 & 93.2 & 94.1
& 79.2 & 76.5 & 70.0
& 61.7 & 60.9 & 46.6
& 94.9 & 89.3 & 89.5 \\

& SANE (ICLR26) \citep{jiao2026samples}
& 97.2 & 92.9 & 93.8
& 84.3 & 78.2 & 73.8
& 62.7 & 61.5 & 46.6
& 94.7 & 88.7 & 89.0 \\

& CAE (AAAI26) \citep{zhu2026hierarchical}
& 98.8 & 98.1 & 97.4
& 77.5 & 77.9 & 66.3
& 60.9 & 62.8 & 45.9
& 90.9 & 81.7 & 80.5 \\

& NTK-SC (ICLR26) \citep{peng2026delving}
& 99.2 & 97.8 & 98.4
& 84.9 & 82.4 & 71.4
& 59.6 & 63.3 & 43.5
& 92.0 & 83.3 & 83.0 \\

\midrule

\multirow{2}{*}{Ours}

& FSRC (no train)
& 99.5 & \underline{99.4} & \underline{99.6}
& \underline{89.4} & \underline{88.2} & \underline{84.2}
& \underline{64.8} & \underline{76.7} & \underline{52.4}
& 97.3 & 93.3 & \underline{94.1} \\

& FSRC
& \textbf{99.9} & \textbf{99.5} & \textbf{99.7}
& \textbf{90.0} & \textbf{88.6} & \textbf{84.9}
& \textbf{70.2} & \textbf{77.4} & \textbf{57.2}
& \textbf{97.8} & \textbf{93.9} & \textbf{94.3} \\

\bottomrule
\end{tabular*}
\label{tab3}
\end{table*}

\begin{table}[t]
  \centering
  \small
  \caption{Clustering performance on two complex datasets (ACC\%, NMI\%, ARI\%). Best and second-best results are \textbf{bold} and \underline{underlined}, respectively.}
  \begin{tabular*}{\hsize}{@{\extracolsep{\fill}}ccccccc}
    \toprule
    \multirow{3}{*}{\centering\arraybackslash\textbf{Method}}
    & \multicolumn{6}{c}{\textbf{Dataset}} \\
    \cmidrule(lr){2-7}
    & \multicolumn{3}{c}{UCF101} 
    & \multicolumn{3}{c}{ImageNet-1K} \\
    \cmidrule(lr){2-4} \cmidrule(lr){5-7}
    & ACC & NMI & ARI & ACC & NMI & ARI \\
    \midrule
    SIC (AAAI23) \citep{cai2023semantic} & 61.9& 81.0 & 53.6 & 47.0 & 77.2 & 34.3 \\
    Turtle (ICML24) \citep{gadetsky2024let} &\underline{77.8} & 86.7 & 69.0 & 56.6 & 76.9 & 45.5 \\
    GradNorm (ICCV25)~\citep{peng2025provable}   & 62.7 & 82.9 & 53.2 & 52.6 & 79.2 & 39.1 \\

NTK-SC (ICLR26) \citep{peng2026delving}
& 67.9 & 83.0 & 59.4
& 56.3 & 79.2 & 39.4 \\

    \midrule
    FSRC(no train) & 75.9 & \underline{87.3} & \underline{70.5} & \underline{60.3} & \underline{81.7} & \underline{50.1} \\
    FSRC            & \textbf{80.4} & \textbf{89.3} & \textbf{71.5} 
                    & \textbf{67.9} & \textbf{82.4} & \textbf{56.6} \\
    \bottomrule
  \end{tabular*}
  \label{tab4}
\end{table}

\section{Experimental Analysis}

\subsection{Experimental settings}

\subsubsection{Datasets and Evaluation metric} The experimental analysis is conducted on 18 benchmark datasets spanning object recognition, fine-grained classification, and remote sensing. The detailed description of each dataset is provided in Appendix \ref{B1}. Train and test splits follow the original dataset configurations. Clustering training is conducted on the training split, while evaluation is performed on the test split using the corresponding ground-truth labels, which are used only for evaluation. To evaluate the performance, we adopted Accuracy (ACC) \citep{xie2016unsupervised}, Normalized Mutual Information (NMI) \citep{estevez2009normalized}, and Adjusted Rand Index (ARI) \citep{steinley2004properties}. The statistical stability of FSRC across 10 random seeds is evaluated and reported in Appendix~\ref{M}.

\subsubsection{Foundation Visual Models}

We employ CLIP \citep{radford2021learning} representations across different architectures and model sizes, including three ResNets (ResNet50, ResNet101, ResNet50x4), three Vision Transformers (VIT-B/32, VIT-B/16, and VIT-L/14), and DINOv2 VIT-G/14 \citep{oquab2023dinov2}. The specific backbones employed in each experiment are indicated in the corresponding sections. Additional details on the models and the procedure for preparing representations are provided in Appendix \ref{B2}.

\subsubsection{Implementation Details}

The training procedure consists of two stages. In the first stage, we implement mutual distillation with two MLP-based cluster heads (d--d--$K$). The cluster heads are trained using Adam (learning rate $=0.001$, $\beta_1=(0.9,0.99)$) for 20 epochs, with a batch size of 512 and a temperature of $T=1.0$. In the second stage, the distilled features are frozen, and the clustering classifier is trained for 6,000 epochs (batch size 10,000) using Adam (learning rate $=0.001$, $\beta_2=(0.9,0.999)$). We employ early stopping only in the second stage, which terminates training if the loss does not decrease for 10 consecutive epochs. All experiments are conducted on a single GPU: ImageNet-1K is trained on an NVIDIA H200 GPU (50 minutes in total), while the remaining datasets are trained on an NVIDIA RTX 4060 Ti GPU.

\subsection{Comparative Analysis}

\subsubsection{Clustering performance results}

We compare FSRC with a range of image clustering methods, including conventional deep image clustering methods: DSEC \citep{chang2018deep}, DCCM \citep{wu2019deep}, GATCluster \citep{niu2020gatcluster}, PICA \citep{huang2020deep}, IDFD \citep{tao2021clustering}, CC \citep{li2021contrastive}, TCL \citep{li2022twin}, ConCUR \citep{deshmukh2022representation}, SPICE \citep{niu2022spice}, and DPAC \citep{yan2024deep}; and image clustering methods with visual models: SIC \citep{cai2023semantic}, TAC \citep{li2023image}, TURTLE \citep{gadetsky2024let}, EC \citep{li2025exploring}, GradNorm \citep{peng2025provable}, CDC-Cal \citep{jia2025towards}, SANE \citep{jiao2026samples}, CAE \citep{zhu2026hierarchical}, and NTK-SC \citep{peng2026delving}.
For a fair comparison, both FSRC and TURTLE use features extracted from the CLIP ViT-L/14 backbone. Here, FSRC (no train) denotes K-Means clustering applied directly to the distilled representations. The evaluation is performed on 4 widely used benchmark datasets and 2 challenging image clustering datasets. The FSRC results are reported as the mean over 10 independent runs.

Tables~\ref{tab3} and~\ref{tab4} summarize the clustering performance. In particular, FSRC  consistently outperforms previous state-of-the-art methods on both standard and complex datasets. For datasets where the original features are less well aligned with the intrinsic data structure, such as ImageNet-Dogs, UCF101, and ImageNet-1K, FSRC achieves significant improvements. For relatively simple datasets where foundation-model features are already highly discriminative, such as ImageNet-10, FSRC still maintains comparable performance. Notably, even without further training, FSRC (no train) yields competitive performance, indicating that the distilled representations are already discriminative and can directly support clustering. 

\begin{figure}[t]
  \centering
  \includegraphics[width=\linewidth]{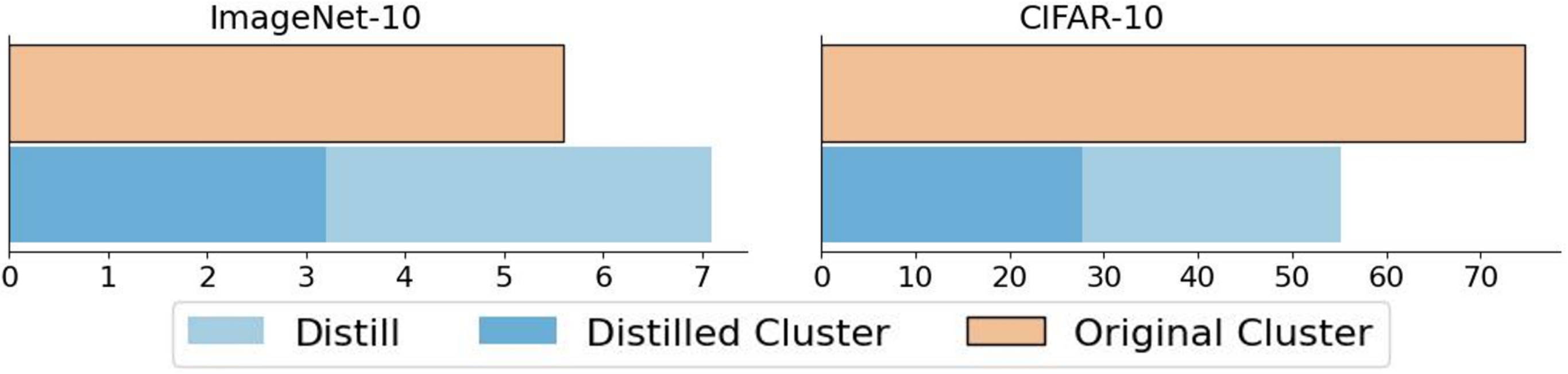}
\caption{Total training time (seconds) of FSRC (including embedding fusion (Distill) and clustering on distilled features (Distilled Cluster)) versus clustering on original features (Original Cluster), evaluated with CLIP ViT-L/14.}
  \label{fig4}
\end{figure}

\subsubsection{Training Efficiency Analysis}

We report the total training time of FSRC, including embedding fusion and clustering assignment. For comparison, we also report the total training time of clustering assignment performed directly on the original features. Experiments are conducted on two datasets (ImageNet-10 and CIFAR-10) using the CLIP-ViT-L/14 backbone. As shown in Figure~\ref{fig4}, clustering assignment trained on distilled features generally converges faster than that trained on original features. Although embedding fusion introduces additional computational overhead, it refines the group structure and accelerates the subsequent clustering assignment, thereby enhancing the overall efficiency of FSRC. Further computational efficiency analyses are provided in Appendix~\ref{O}.

\subsection{Effectiveness Analysis of Embedding Fusion}

We analyze the effectiveness of embedding fusion from two aspects to demonstrate that it yields clustering-friendly image representations. First, we evaluate the triplet accuracy of the representations, which directly reflects their separability. Second, we assess the performance on downstream clustering tasks, validating the practical utility of the fused embeddings.

\subsubsection{Result on Triplet Accuracy}

Evaluating the discriminability of learned representations is fundamental to understanding their suitability for downstream tasks~\cite{wangstabilizing}. To evaluate the effectiveness of FSRC in enhancing the separability of fused representations, we adopt triplet accuracy as a metric. 

Triplet accuracy measures the ability to correctly distinguish between similar and dissimilar image pairs within an embedding space. Specifically, given an anchor image, a positive image from the same class, and a negative image from a different class, the triplet is considered correct if the distance between the anchor and the positive image is smaller than the distance between the anchor and the negative image. Formally, for a triplet \((x_i, x_j, x_k)\), where \(x_i\) and \(x_j\) belong to the same class, and \(x_i\) and \(x_k\) belong to different classes, let the corresponding embeddings be \(f(x_i), f(x_j), f(x_k)\). The triplet is considered correct if:

\[
\operatorname{dist}\!\bigl(f(x_i), f(x_j)\bigr) < \operatorname{dist}\!\bigl(f(x_i), f(x_k)\bigr),
\]
where \(\operatorname{dist}(\cdot,\cdot)\) denotes a distance function, which in this case is cosine distance.

Figure~\ref{fig:combined} (a) presents the comparison of triplet verification accuracy on 10,000 randomly sampled triplets across 18 datasets using the ClipVIT-L/14 backbone. As shown in Figure~\ref{fig:combined} (a), the distilled features achieve substantially higher triplet accuracy. Additional results across six other backbone architectures are provided in Table~\ref{tab:triplet_accuracy} in Appendix~\ref{A}, consistently demonstrating that distilled representations achieve higher triplet accuracy and are more clustering-friendly. Furthermore, improvements in triplet accuracy are strongly positively correlated with downstream clustering performance, as detailed in Appendix~\ref{A}.

\begin{figure}[t]
  \centering
  \includegraphics[width=0.85\linewidth]{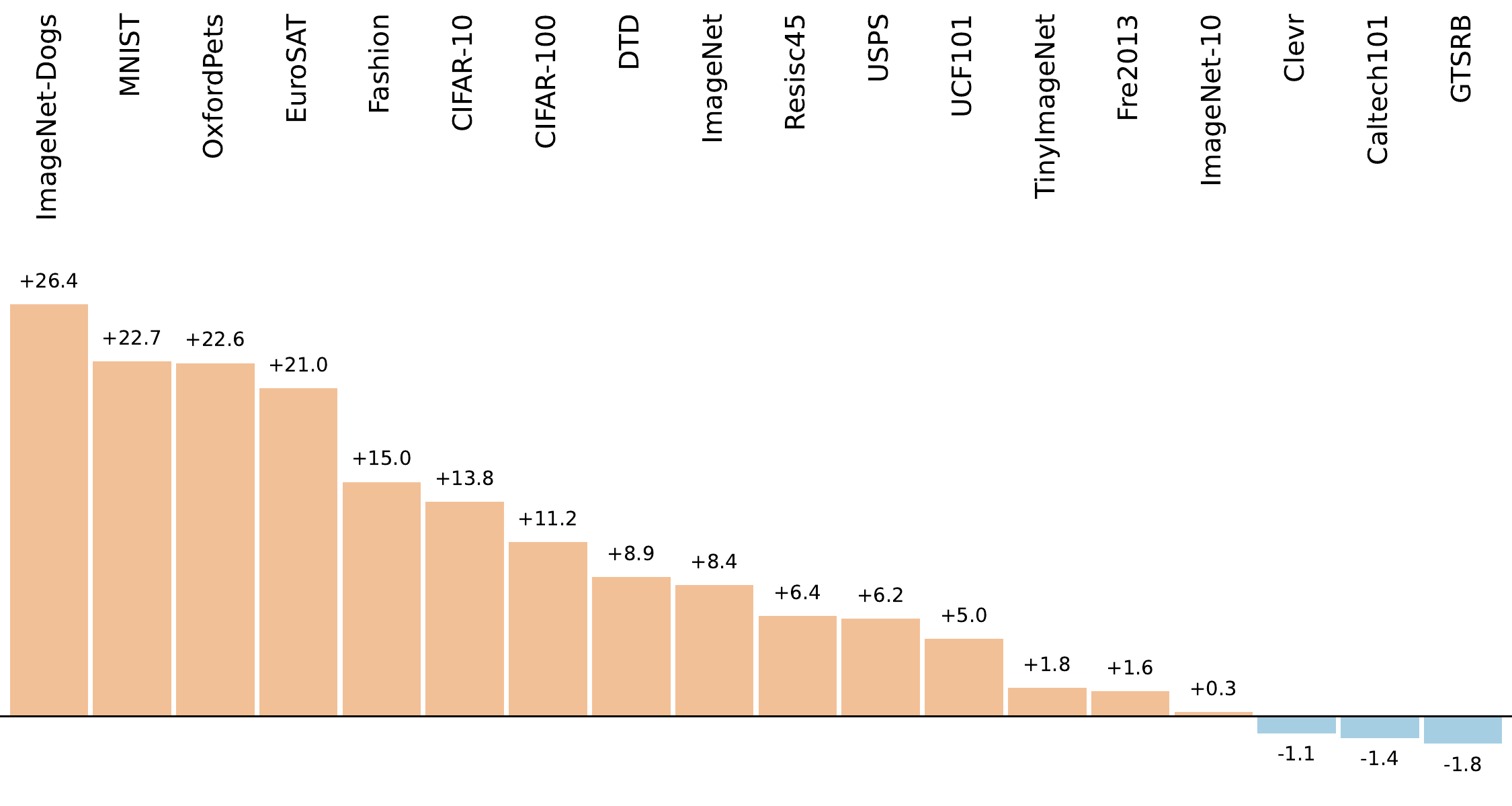}
\caption{K-Means ACC improvement (\%) on 18 datasets: distilled vs. original ViT-L/14 features.}
  \label{fig44}
\end{figure}

\subsubsection{Result on Clustering Performance}

\begin{figure}[t]
  \centering
  \includegraphics[width=0.85\linewidth]{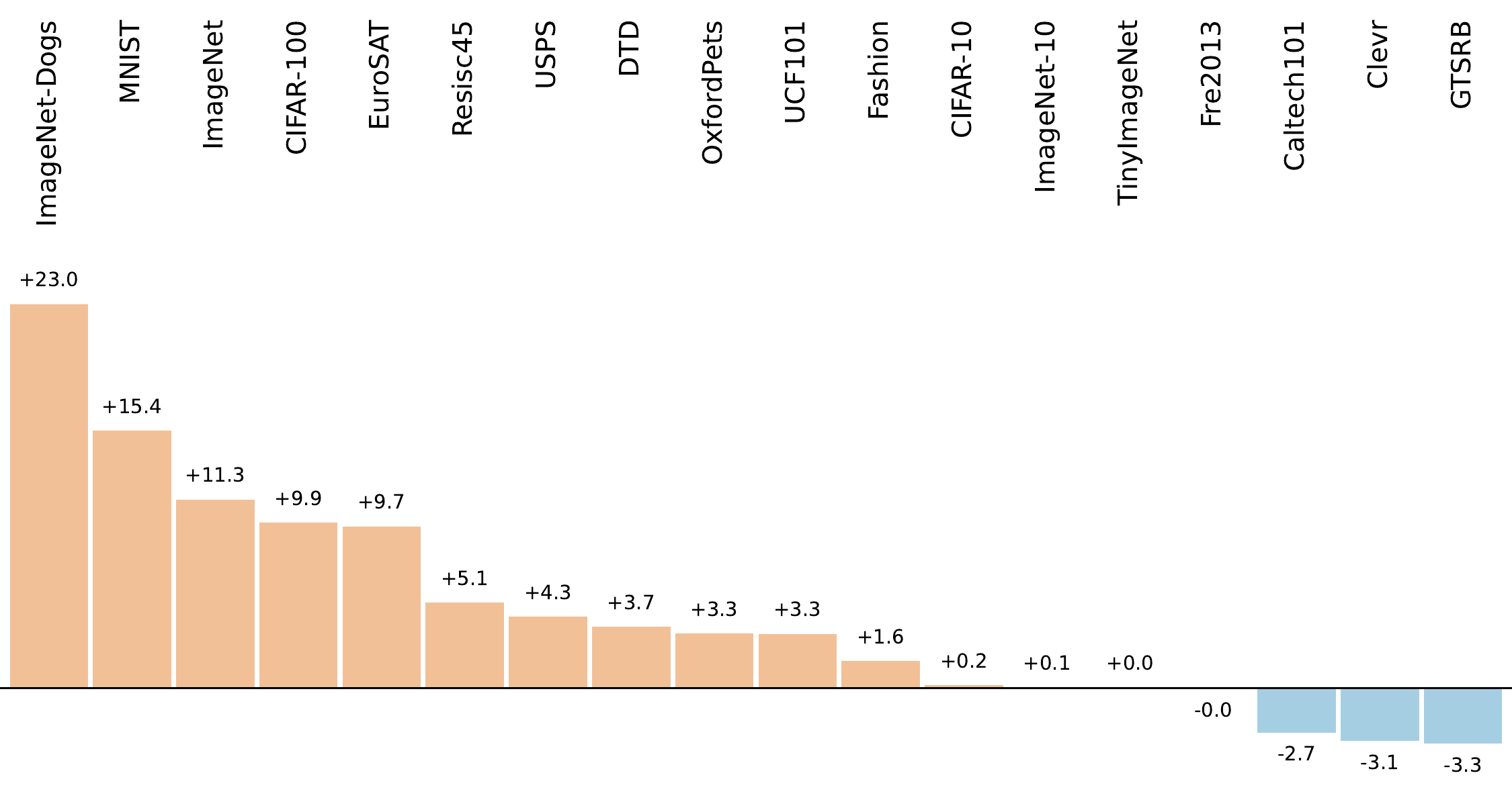}
\caption{Assignment ACC improvement (\%) on 18 datasets: distilled vs. original ViT-L/14 features.}
  \label{fig5}
\end{figure}

\begin{figure*}[t]
  \centering
  \begin{tabular}{cccc}
    \includegraphics[width=0.23\linewidth]{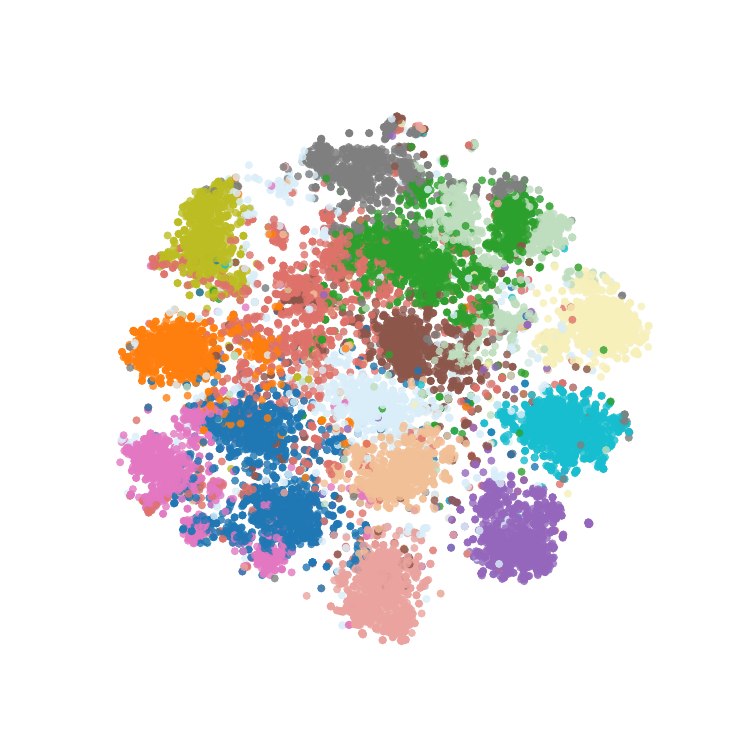} &
    \includegraphics[width=0.23\linewidth]{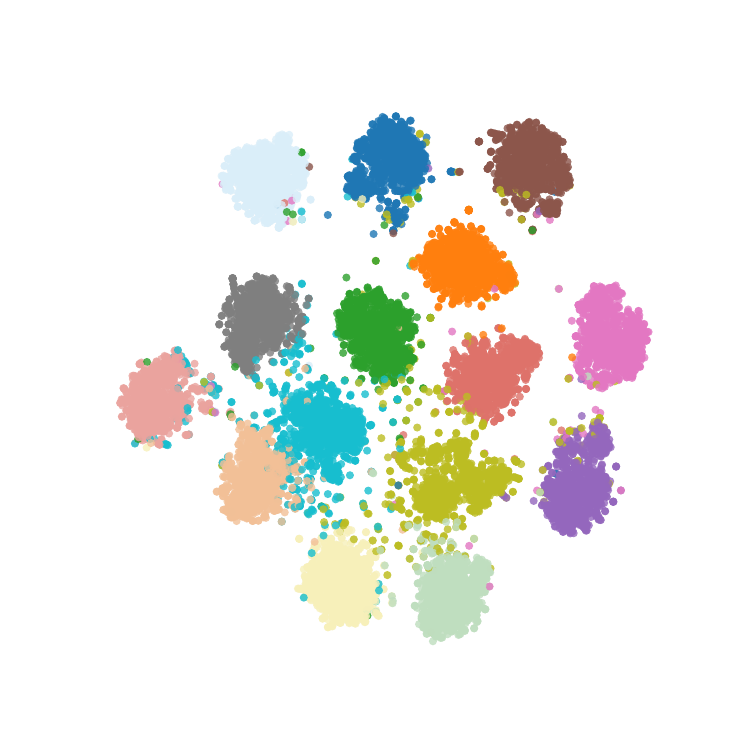} &
    \includegraphics[width=0.23\linewidth]{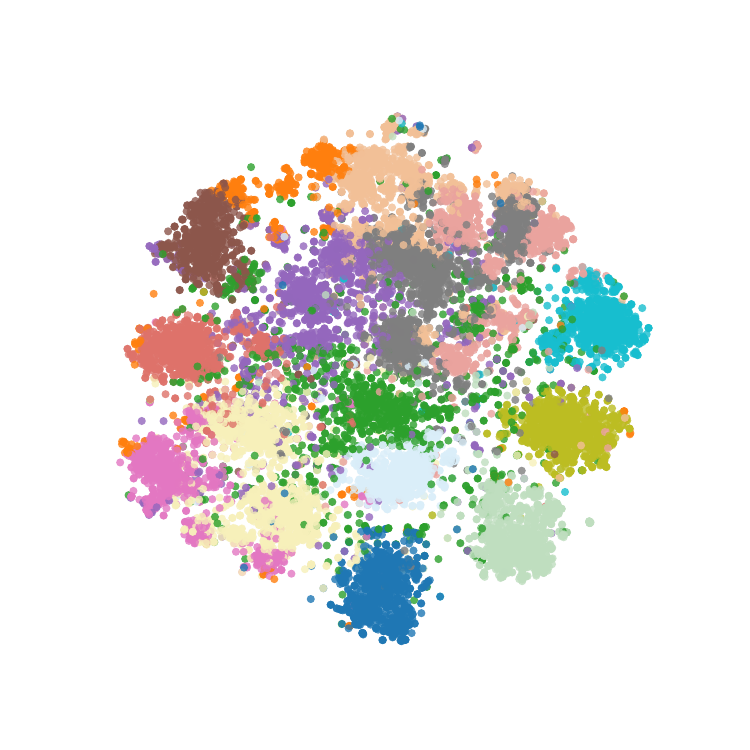} &
    \includegraphics[width=0.23\linewidth]{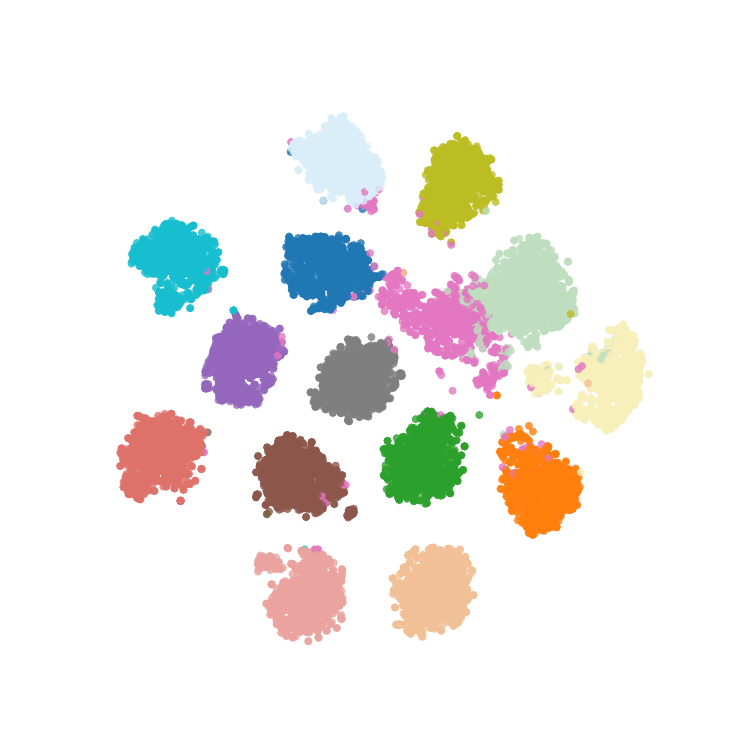} \\
    A. Global Features & B. Local Features & C. Simple Concatenate & D. FSRC
  \end{tabular}
  \caption{t-SNE visualizations of the ImageNet-Dogs dataset under four experimental setups.}
  \label{fig7}
\end{figure*}

To further evaluate the effectiveness of the fused features, we perform both K-Means clustering and a clustering assignment model on the distilled embeddings and compare the clustering accuracy (ACC) with that obtained from the original global semantic embeddings. This experiment examines whether integrating local relational features with global semantic features improves feature separability and enhances clustering performance. We conduct the analysis on all 18 datasets using the CLIP ViT-L/14 backbone, providing a complementary perspective to the triplet-accuracy evaluation.

The K-Means ACC improvements are shown in Figure~\ref{fig44}. The distilled embeddings achieve substantial ACC gains on fine-grained datasets such as MNIST, ImageNet-Dogs, and OxfordPets, indicating that feature fusion effectively captures discriminative details beneficial for clustering. Conversely, a slight performance drop is observed on datasets such as CLEVR, Caltech101, and GTSRB, which can be attributed to their strong reliance on global semantic cues, where local relational features contribute less. Successful and failed cases are further analyzed in Appendix~\ref{P}.
Overall, the trends in K-Means ACC changes are consistent with the triplet-accuracy results, suggesting a strong correlation between embedding separability. Additional results of the K-Means performance improvement on six backbones are provided in Appendix~\ref{D}.

We evaluate clustering performance using a clustering-assignment model on both the original global semantic features and the distilled features. As shown in Figure~\ref{fig5}, ACC trends remain consistent with triplet accuracy and K-Means results. This indicates that improvements in embedding separability through feature fusion generally lead to better clustering results. Overall, these observations highlight that FSRC's feature fusion primarily benefits clustering when local relational cues complement global semantic embeddings. Additional results across 6 other backbone architectures, along with further analysis, are provided in Appendix~\ref{E}, which also confirm the clustering-friendly nature of the fused features.

\subsubsection{Feature Contributions and Fusion Mechanisms Analysis}

To disentangle the contributions of local relational information and feature fusion, we conduct component-wise ablations on ImageNet-10, ImageNet-Dogs, and CIFAR-10. Global Only uses CLIP ViT-L/14 features, while Local Only uses features extracted by the Local Relation Encoder. Global + Local directly concatenates the two embeddings without mutual distillation. FSRC(w/o GAT) removes only the GAT, leaving all other components unchanged. Single-backbone FSRC extracts both global and local features from CLIP ViT-L/14 without an additional ResNet-50. All variants are evaluated using the same downstream clustering assignment module.

\begin{table}[t]
\centering
\caption{Component-wise analysis on three representative datasets. Results are reported by ACC (\%).}
\label{component_ablation}
\renewcommand{\arraystretch}{1.12}
\setlength{\tabcolsep}{5pt}
\resizebox{\columnwidth}{!}{
\begin{tabular}{lccc}
\toprule
Method & ImageNet-10 & ImageNet-Dogs & CIFAR-10 \\
\midrule
Global Only & 99.8 & 74.4 & 97.6 \\
Local Only & 59.1 & 54.0 & 56.6 \\
Global + Local & 99.8 & 74.0 & 97.6 \\
FSRC(w/o GAT) & 99.8 & 76.5 & 87.5 \\
Single-backbone FSRC & 99.9 & 87.9 & 97.7 \\
\textbf{FSRC (Ours)} & \textbf{99.9} & \textbf{90.0} & \textbf{97.8} \\
\bottomrule
\end{tabular}
}
\end{table}

As shown in Table~\ref{component_ablation}, local relational features alone do not outperform global semantic features, while their simple concatenation yields no improvement. In contrast, FSRC achieves superior performance, demonstrating that its gains arise from the effective integration of complementary representations. Removing GAT substantially degrades performance, highlighting the importance of spatial relation modeling within the fusion framework. Moreover, single-backbone FSRC, which extracts local patch features directly from CLIP without an additional ResNet-50, remains competitive, indicating that the improvements are not solely attributable to additional backbone capacity. 

\subsection{Visualization Analysis}

To visually illustrate the representation structure learned by FSRC, we conduct a t-SNE visualization study \cite{maaten2008visualizing} on the ImageNet-Dogs dataset under K-Means clustering with four experimental setups: (1) using only global semantic ViT-L/14 features, (2) using only local relational features, (3) a naive fusion strategy that directly concatenates the two features, and (4) the distilled features obtained by mutual distillation of global and local features within FSRC. 

As shown in Figure~\ref{fig7}, global semantic features exhibit relatively scattered distributions with limited separation between visually similar categories, while local relational features capture structural cues that differ from those represented by global semantic features. However, simply concatenating the two feature types does not substantially improve the clustering structure. In contrast, FSRC effectively integrates their complementary information through mutual distillation, producing more compact and well-separated clusters. Additional t-SNE comparisons with representative clustering methods are provided in Appendix J.

\subsection{Additional Results}

Appendix \ref{J} shows the convergence of FSRC. Appendix \ref{K0} presents parameter analysis, demonstrating FSRC's robustness to hyperparameters. Appendix \ref{K} further provides semantic compactness results showing that local relations improve cluster compactness.

\section{Conclusion}
The visual features extracted from pre-trained image-text models, such as CLIP, are not always well aligned with the intrinsic structure of specific datasets, which limits their effectiveness for challenging clustering tasks. To address this, we propose FSRC, an image clustering framework that effectively fuses global semantic and local relational representations through a mutual distillation mechanism. By capturing patch-level relational cues through a Local Relation Encoder and integrating them with rich global semantics, FSRC produces more separable and clustering-friendly embeddings that significantly and consistently enhance clustering performance. Comprehensive experiments on 18 diverse benchmark datasets demonstrate that FSRC consistently outperforms existing state-of-the-art methods and remains highly stable across 7 different backbone models, clearly confirming its effectiveness, robustness, and broad applicability.

\bibliographystyle{ACM-Reference-Format}
\bibliography{sample-base}

\appendix
\onecolumn
\section*{Appendix}

The appendix is organized as follows:

\begin{itemize}

    \item \ref{B1} Dataset Description
    \item \ref{B2} Visual Foundation Models
    \item \ref{A} Additional Triplet Accuracy Results
    \item \ref{C} Theoretical Proofs and Empirical Validation
    \item \ref{D} Embedding Fusion Effectiveness Analysis on K-Means
    \item \ref{E} Embedding Fusion Effectiveness Analysis on Clustering Assignment 
    \item \ref{P} Successful and Failed Cases Analysis
    \item \ref{G} Clustering Stability on 18 Datasets Across 7 Backbones
    \item \ref{H} Clustering Accuracy on 18 Datasets Across 7 Backbones
    \item \ref{I} Visualization Comparison with Representative Clustering Methods
    \item \ref{J} Training Objective Convergence Analysis
    \item \ref{K0} Parameter Analysis
    \item \ref{K} Additional Semantic Compactness Results
    \item \ref{N} Effectiveness Analysis of Local Relational Graph Construction
    \item \ref{M} Statistical Stability Analysis
     \item \ref{O} Computational Cost Analysis

\end{itemize}

\section{Dataset Description}
\label{B1}

We evaluate our framework on 18 benchmark datasets, covering a wide range of vision tasks. These include general object classification datasets CIFAR-10 \citep{krizhevsky2009learning}, CIFAR-100 \citep{krizhevsky2009learning}, TinyImageNet \citep{le2015tiny}, ImageNet-10 \citep{chang2017deep}, Caltech101 \citep{fei2004learning}, and ImageNet \citep{deng2009imagenet}; fine-grained object classification datasets ImageNet-Dogs \citep{chang2017deep} and OxfordPets \citep{parkhi2012cats}; grayscale image datasets USPS \citep{hull1994database} and MNIST \citep{lecun1998gradient}; the fashion image dataset Fashion-MNIST \citep{xiao2017fashion}; the texture dataset DTD \citep{cimpoi2014describing}; the facial emotion recognition dataset FER2013 \citep{goodfellow2013challenges}; the satellite image classification datasets EuroSAT \citep{helber2019eurosat} and RESISC45 \citep{cheng2017remote}; the German Traffic Sign Recognition Benchmark (GTSRB) \citep{stallkamp2012man}; the CLEVR counting dataset \citep{johnson2017clevr}; and the video dataset UCF101 \citep{soomro2012ucf101}.

For the CLEVR dataset, we randomly sample 2,000 training and 500 test images. For UCF101, we extract the middle frame from each video clip for input. The key statistics of all datasets are summarized in Table~\ref{tab:dataset_summary}.

\begin{table}[htbp]
  \centering
  \renewcommand{\arraystretch}{1.15}
  \small
  \caption{Summary statistics of the 18 datasets, including number of classes, train size, and test size.}
  \begin{tabular}{lcccc}
    \toprule
    \textbf{No.} & \textbf{Dataset} & \textbf{Num Classes} & \textbf{Train Size} & \textbf{Test Size} \\
    \midrule
    1  & CLEVR \citep{johnson2017clevr}           & 8    & 2,000    & 500 \\
    2  & Caltech101 \citep{fei2004learning}     & 102  & 3,060    & 6,084 \\
    3  & OxfordPets \citep{parkhi2012cats}          & 37   & 3,680    & 3,669 \\
    4  & DTD \citep{cimpoi2014describing}           & 47   & 3,760    & 1,880 \\
    5  & USPS \citep{hull1994database}          & 10   & 7,291    & 2,007 \\
    6  & ImageNet-10 \citep{chang2017deep}   & 10   & 10,500   & 2,630 \\
    7  & ImageNet-Dogs \citep{chang2017deep} & 15   & 15,600   & 3,900 \\
    8  & EuroSAT \citep{helber2019eurosat}        & 10   & 10,000   & 5,000 \\
    9  & GTSRB \citep{stallkamp2012man}         & 43   & 26,640   & 12,630 \\
    10 & FER2013 \citep{goodfellow2013challenges}       & 7    & 28,709   & 7,178 \\
    11 & CIFAR-10 \citep{krizhevsky2009learning}       & 10   & 50,000   & 10,000 \\
    12 & CIFAR-100 \citep{krizhevsky2009learning}      & 100  & 50,000   & 10,000 \\
    13 & MNIST \citep{lecun1998gradient}            & 10   & 60,000   & 10,000 \\
    14 & TinyImageNet \citep{le2015tiny}  & 200  & 100,000  & 10,000 \\
    15 & Fashion \citep{xiao2017fashion}        & 10   & 60,000   & 10,000 \\
    16 & UCF101 \citep{soomro2012ucf101}        & 101  & 9,537    & 3,783 \\
    17 & RESISC45 \citep{cheng2017remote}      & 45   & 25,200   & 6,300 \\
    18 & ImageNet-1K \citep{deng2009imagenet}        & 1000 & 1,281,167  & 50,000 \\
    \bottomrule
  \end{tabular}
  \label{tab:dataset_summary}
\end{table}

\section{Visual Foundation Models}
\label{B2}

FSRC leverages features from various pre-trained visual foundation models. To comprehensively assess its effectiveness and robustness, we evaluate FSRC using seven prominent models that differ in pre-training data, architecture, and scale: three CLIP ResNets (ResNet50, ResNet101, ResNet50x4) and three CLIP Vision Transformers (ViT-B/32, ViT-B/16, ViT-L/14), all pre-trained on WebImageText-400M \citep{radford2021learning}, as well as DINOv2 ViT-G/14 pre-trained on LVD-142M \citep{oquab2023dinov2}. The specific backbone used in each experiment is specified in the corresponding section. For all models, image features are pre-extracted using their standard preprocessing pipelines. Key characteristics of these foundation models are summarized in Table~\ref{tab2}.

\begin{table}[t]
\centering
\small
\caption{Visual foundation models, including architecture, number of parameters,  and textual supervision status.}
\label{tab2}
\begin{tabular}{ccccc}
\toprule
\textbf{Model} & \textbf{Architecture} & \textbf{Parameters} & \textbf{Trained on} & \textbf{Textual} \\
\midrule
\multirow{6}{*}{CLIP\citep{radford2021learning} } 
& RN50     & 100M & \multirow{6}{*}{WebImageText-400M} & \multirow{6}{*}{$\checkmark$} \\
& RN101    & 120M & & \\
& RN50x4   & 180M & & \\
\cmidrule(lr){2-3}
& VIT-B/32 & 150M & & \\
& VIT-B/16 & 150M & & \\
& VIT-L/14 & 430M & & \\
\midrule
DINOv2 \citep{oquab2023dinov2}    & VIT-G/14 & 1.1B & LVD-142M     & $\times$ \\
\bottomrule
\end{tabular}
\end{table}

\begin{table}[t]
\centering
\caption{Triplet accuracy (\%) on 10,000 randomly sampled triplets. \underline{Underlined} numbers indicate the higher value within each Original/Distilled pair for the same architecture. Original denotes the original global semantic embeddings, and Distilled denotes the Distilled Representation, which is obtained via mutual distillation between global semantic embeddings and local relational embeddings.}
\renewcommand{\arraystretch}{1.5}
\resizebox{\textwidth}{!}{
\begin{tabular}{c|cc|cc|cc|cc|cc|cc|cc}
\toprule
 & \multicolumn{2}{c}{ClipResNet50} & \multicolumn{2}{c}{ClipResNet101} & \multicolumn{2}{c}{ClipResNet50*4} & \multicolumn{2}{c}{ClipVITB/32} & \multicolumn{2}{c}{ClipVITB/16} & \multicolumn{2}{c}{DINOv2} & \multicolumn{2}{c}{ClipVITL/14} \\
\cmidrule(lr){2-3} \cmidrule(lr){4-5} \cmidrule(lr){6-7} \cmidrule(lr){8-9}\cmidrule(lr){10-11}\cmidrule(lr){12-13}\cmidrule(lr){14-15}
  & Original \hspace{0.5pt}& Distilled \hspace{1pt}& Original \hspace{0.5pt}& Distilled \hspace{1pt}& Original \hspace{0.5pt}& Distilled \hspace{1pt}& Original \hspace{0.5pt}& Distilled \hspace{1pt}& Original \hspace{0.5pt}& Distilled \hspace{1pt}& Original \hspace{0.5pt}& Distilled \hspace{1pt}& Original \hspace{0.5pt}& Distilled\\

\midrule
MNIST        & 73.34 & \underline{96.78} & 74.41 & \underline{96.23} & 73.08 & \underline{96.63} & 75.14 & \underline{97.09} & 74.98 & \underline{97.29} & 75.13 & \underline{91.87} & 80.30 & \underline{97.53} \\
GTSRB        & 75.62 & \underline{83.99} & 77.14 & \underline{85.80} & 76.71 & \underline{86.43} & 74.64 & \underline{82.43} & 77.94 & \underline{88.38} & 69.13 & \underline{82.91} & 84.79 & \underline{92.90} \\
ImageNet-10  & 97.13 & \underline{99.95} & 97.30 & \underline{99.93} & 97.23 & \underline{99.94} & 97.35 & \underline{99.92} & 97.36 & \underline{99.95} & 96.46 & \underline{99.92} & 96.86 & \underline{99.95} \\
OxfordPets   & 87.93 & \underline{98.91} & 91.11 & \underline{99.10} & 91.20 & \underline{99.40} & 88.37 & \underline{99.35} & 65.14 & \underline{84.21} & 83.23 & \underline{95.65} & 93.29 & \underline{99.65} \\
ImageNet-Dogs& 69.74 & \underline{95.94} & 75.42 & \underline{97.28} & 76.50 & \underline{97.62} & 71.26 & \underline{97.01} & 73.98 & \underline{97.30} & 97.75 & \underline{99.42} & 78.08 & \underline{99.05} \\
DTD          & 80.48 & \underline{89.36} & 81.22 & \underline{90.11} & 82.42 & \underline{90.40} & 81.04 & \underline{89.65} & 81.97 & \underline{90.27} & 86.33 & \underline{91.06} & 81.25 & \underline{91.20} \\
CIFAR-10     & 81.00 & \underline{92.87} & 85.32 & \underline{95.77} & 84.67 & \underline{95.66} & 87.97 & \underline{97.60} & 87.93 & \underline{98.45} & 89.54 & \underline{99.56} & 86.12 & \underline{98.93} \\
CIFAR-100    & 80.07 & \underline{90.64} & 84.83 & \underline{92.61} & 83.67 & \underline{92.47} & 86.82 & \underline{95.43} & 86.44 & \underline{96.03} & 96.01 & \underline{98.66} & 84.74 & \underline{97.31} \\
TinyImageNet & 84.00 & \underline{88.25} & 86.94 & \underline{90.15} & 86.27 & \underline{89.75} & 90.68 & \underline{93.15} & 90.47 & \underline{93.59} & \underline{97.20} & 95.81 & 89.77 & \underline{94.55} \\
Caltech101   & 95.49 & \underline{98.46} & 96.86 & \underline{99.22} & 97.22 & \underline{98.99} & 97.09 & \underline{99.15} & 97.25 & \underline{99.31} & 98.99 & \underline{99.61} & 97.48 & \underline{99.71} \\
FER      & 56.27 & \underline{59.59} &57.80 &  \underline{58.91} & 57.93 & \underline{59.79} & 57.96 & \underline{59.18} & 58.05 & \underline{59.39} & 61.16 & \underline{61.42} & 57.56 & \underline{61.45} \\
EuroSAT      & 80.51 & \underline{94.15} & 81.18 & \underline{96.27} & 82.89 & \underline{96.03} & 81.64 & \underline{94.56} & 84.26 & \underline{97.79} & 84.09 & \underline{98.01} & 87.86 & \underline{98.77} \\
RESISC45     & 91.18 & \underline{96.63} & 93.02 & \underline{97.49} & 92.89 & \underline{97.24} & 93.55 & \underline{98.01} & 95.20 & \underline{98.30} & 92.38 & \underline{97.35} & 95.45 & \underline{98.62} \\
UCF101       & 92.70 & \underline{96.52} & 94.63 & \underline{96.69} & 94.24 & \underline{97.51} & 94.57 & \underline{97.33} & 95.88 & \underline{97.49} & 98.22 & \underline{98.40} & 97.55 & \underline{99.02} \\
CLEVR        & 60.10 & \underline{60.10} & \underline{62.00} & 58.90 & 58.75 & \underline{60.75} & 59.90 & \underline{61.35} & 58.70 & \underline{60.55} & \underline{55.45} & 54.80 & \underline{59.20} & 58.95 \\
USPS         & 85.01 & \underline{96.23} & 81.17 & \underline{96.80} & 82.49 & \underline{96.54} & 87.30 & \underline{97.83} & 87.97 & \underline{97.52} & 83.23 & \underline{95.65} & 88.70 & \underline{97.81} \\
Fashion      & 81.12 & \underline{90.72} & 84.10 & \underline{91.80} & 82.20 & \underline{91.88} & 84.82 & \underline{92.48} & 82.64 & \underline{91.58} & 87.48 & \underline{93.94} & 81.66 & \underline{91.64} \\
ImageNet-1K      & 94.47 & \underline{96.65} & 94.61 & \underline{97.24} & 95.36 & \underline{97.69} & 94.70 & \underline{97.15} & 95.56 & \underline{97.99} & 97.77 & \underline{98.25} & 97.01 & \underline{97.96} \\
\hline
\textbf{AVG.}& 81.45 & \underline{90.32} & 83.28 & \underline{91.13} & 83.10 & \underline{91.37} & 83.60 & \underline{91.59} & 82.87 & \underline{91.41} & 86.09& \underline{91.79}  & 85.43 & \underline{93.06} \\
\bottomrule
\end{tabular}}
\label{tab:triplet_accuracy}
\end{table}

\section{Additional Triplet Accuracy Results}
\label{A}
\subsection{Triplet Accuracy Across Backbones}
We use triplet accuracy to evaluate the improvement in representation separability achieved by FSRC. Table~\ref{tab:triplet_accuracy} reports the triplet verification accuracy on 10,000 randomly sampled triplets across all 18 datasets, comparing the original global semantic embeddings (\textit{Original}) with the distilled embeddings (\textit{Distilled}) obtained through mutual distillation. Results are provided for multiple backbone networks, including CLIP ResNets (ResNet50, ResNet101, ResNet50x4), CLIP Transformers (ViT-B/32, ViT-B/16, ViT-L/14), and DINOv2.

Overall, the distilled embeddings consistently outperform the original representations across nearly all datasets and backbones in triplet accuracy. Particularly pronounced gains are observed on fine-grained datasets with large intra-class variation, such as \textit{OxfordPets} and \textit{ImageNet-Dogs}, where improvements exceed 20\% in several cases, highlighting the effectiveness of incorporating local relational cues to complement global semantics. On average, triplet accuracy increases across different backbones.

Higher triplet accuracy indicates enhanced embedding separability, in which semantically similar samples are drawn closer while dissimilar ones are pushed further apart. This improved separability strongly correlates with better clustering performance, resulting in more accurate, stable, and robust groupings. Consequently, the proposed mutual distillation strategy effectively mitigates the intrinsic limitations of foundation-model features in clustering tasks.

\subsection{Correlation Between Triplet Accuracy and Clustering Performance}

To further investigate whether improvements in triplet accuracy are associated with downstream clustering performance, we conduct a quantitative correlation analysis across all 18 datasets using CLIP ViT-L/14. For each dataset, we compute the improvements in triplet accuracy and K-Means ACC between the original and distilled representations.

Across all 18 datasets, we obtain a Pearson correlation coefficient of $r=0.759$ ($p<0.001$) and a Spearman rank correlation coefficient of $\rho=0.725$ ($p<0.001$). Both coefficients indicate a strong positive association between improvements in triplet accuracy and gains in downstream clustering performance. These results provide empirical evidence that the enhanced discriminability of distilled representations is closely associated with their improved clustering effectiveness.

\section{Theoretical Proofs and Empirical Validation}
\label{C}

\subsection{Theoretical Proofs}
THEOREM \ref{thm:simplified} (INTER-CLASS CONFUSION REDUCTION)
\emph{Under Assumptions~\ref{assump:weak-discriminability}--\ref{assump:weak-complementary} and the condition $\eta < \min\{\rho_V(1-\rho_S), \ \rho_S(1-\rho_V)\}$, the distilled feature yields an inter-class confusion probability satisfying:
\begin{equation*} \mathbb{P}(\mathcal{E}_Z(\tau) \mid Y_i \neq Y_j) \leq \rho_V \rho_S + \eta < \min\{\rho_V, \rho_S\}. \end{equation*}}

\begin{proof}
We analyze the probability that two samples from different classes are mapped
close to each other in the fused embedding space.

The mutual distillation process aligns the two modalities $\phi(x)$ and $g(x)$
and enforces an \emph{agreement-preserving} fusion mechanism. In particular, for
two samples $X_i, X_j$ drawn from different latent classes ($Y_i \neq Y_j$),
similarity in the fused representation $\hat{\phi}$ can occur only if
\emph{both} original modalities erroneously deem the samples to be similar.
That is, proximity in the fused space requires simultaneous proximity in the
individual modality spaces.

Formally, for the fused-space proximity event $\mathcal{E}_Z(\tau)$ to occur
under $Y_i \neq Y_j$, it is \emph{necessary} that the corresponding proximity
events $\mathcal{E}_V(\tau_V)$ and $\mathcal{E}_S(\tau_S)$ also occur for some
thresholds $\tau_V$ and $\tau_S$ related to $\tau$. This necessity is captured
by the set inclusion
\[
\mathcal{E}_Z(\tau) \subseteq \mathcal{E}_V(\tau_V) \cap \mathcal{E}_S(\tau_S).
\]
For simplicity of exposition, and without loss of generality, we assume that the feature spaces are appropriately rescaled so that a common threshold $\tau$ can be used across modalities, yielding
\[
\mathcal{E}_Z(\tau) \subseteq \mathcal{E}_V(\tau) \cap \mathcal{E}_S(\tau).
\]

Conditioning on $Y_i \neq Y_j$ and taking probabilities on both sides gives
\begin{equation}
\mathbb{P}(\mathcal{E}_Z(\tau) \mid Y_i \neq Y_j)
\;\leq\;
\mathbb{P}(\mathcal{E}_V(\tau) \cap \mathcal{E}_S(\tau) \mid Y_i \neq Y_j).
\label{eq:Z-to-joint}
\end{equation}

By Assumption~\ref{assump:weak-complementary} (Approximate Complementarity), the
joint ``failure'' probability---that is, two samples from different classes
being close in both modalities---admits an upper bound in terms of the product
of the marginal failure probabilities, up to an additive deviation $\eta$:
\begin{equation}
\mathbb{P}(\mathcal{E}_V(\tau) \cap \mathcal{E}_S(\tau) \mid Y_i \neq Y_j)
\;\leq\;
\mathbb{P}(\mathcal{E}_V(\tau) \mid Y_i \neq Y_j)
\,
\mathbb{P}(\mathcal{E}_S(\tau) \mid Y_i \neq Y_j)
+ \eta .
\label{eq:joint-bound}
\end{equation}

Moreover, by Assumption~\ref{assump:weak-discriminability} (Partial
Discriminability), each individual modality exhibits bounded inter-class
confusion:
\begin{equation}
\mathbb{P}(\mathcal{E}_V(\tau) \mid Y_i \neq Y_j) \leq \rho_V,
\qquad
\mathbb{P}(\mathcal{E}_S(\tau) \mid Y_i \neq Y_j) \leq \rho_S .
\label{eq:marginal-bounds}
\end{equation}

Substituting~\eqref{eq:marginal-bounds} into~\eqref{eq:joint-bound}, and then
combining with~\eqref{eq:Z-to-joint}, we obtain
\[
\mathbb{P}(\mathcal{E}_Z(\tau) \mid Y_i \neq Y_j)
\;\leq\;
\rho_V \rho_S + \eta .
\]

Furthermore, whenever the complementarity deviation satisfies
\[
\eta < \min(\rho_V, \rho_S) - \rho_V \rho_S ,
\]
it follows that
\[
\mathbb{P}(\mathcal{E}_Z(\tau) \mid Y_i \neq Y_j)
\;\leq\;\rho_V \rho_S + \eta < \min(\rho_V, \rho_S),
\]
implying that the fused embedding $\hat{\phi}$ exhibits strictly lower
inter-class confusion than either individual modality. This completes the
proof of Theorem~\ref{thm:simplified}.
\end{proof}

THEOREM \ref{thm:simplified-intra} (PRESERVATION OF USEFUL STRUCTURE)
\emph{The fused embedding $\hat{\phi}(x)$ preserves intra-class proximity at least
as well as the better of the two individual modalities:
\begin{align*}
\mathbb{P}(\mathcal{E}_Z(\tau) \mid Y_i = Y_j)
\ge
\max\Bigl\{
\mathbb{P}(\mathcal{E}_V(\tau) \mid Y_i = Y_j),\;
\mathbb{P}(\mathcal{E}_S(\tau) \mid Y_i = Y_j)
\Bigr\}
- \delta,
\end{align*}
where $\delta \ge 0$ is a constant induced by the fusion error and the
threshold rescaling.}

\begin{proof}
We assume the fused embedding $\hat{\phi}(X)$ approximates each
individual modality with a bounded fusion error $\epsilon \ge 0$:
\[
\| \hat{\phi}(X) - \Pi_V(\phi(X)) \| \le \epsilon,
\qquad
\| \hat{\phi}(X) - \Pi_S(g(X)) \| \le \epsilon,
\]
where $\Pi_V$ and $\Pi_S$ denote the projection operators for the global semantic and
local relational modalities, respectively. Both $\Pi_V$ and $\Pi_S$ are Lipschitz
continuous with constants $L_V$ and $L_S$.

By the triangle inequality,
\[
\| \hat{\phi}(X_i) - \hat{\phi}(X_j) \|
\le \| \Pi_V(\phi(X_i)) - \Pi_V(\phi(X_j)) \| + 2\epsilon.
\]
Using the $L_V$-Lipschitz continuity of $\Pi_V$,
\[
\| \Pi_V(\phi(X_i)) - \Pi_V(\phi(X_j)) \|
\le L_V \| \phi(X_i) - \phi(X_j) \|.
\]
Hence,
\begin{align}\label{use}
\| \hat{\phi}(X_i) - \hat{\phi}(X_j) \|
\le L_V \| \phi(X_i) - \phi(X_j) \| + 2\epsilon.
\end{align}

Under the condition 
\(
\| \phi(X_i) - \phi(X_j) \| \le \dfrac{\tau - 2\epsilon}{L_V},
\) then (\ref{use}) gives
\[
\| \hat{\phi}(X_i) - \hat{\phi}(X_j) \| \le \tau.
\]
Thus,
\begin{align}\label{use_1}
\mathcal{E}_V\!\left( \frac{\tau - 2\epsilon}{L_V} \right)
\subseteq \mathcal{E}_Z(\tau).
\end{align}

Conditioning on $Y_i = Y_j$ and taking probabilities in (\ref{use_1}) yields
\begin{align}\label{use_2}
\mathbb{P}(\mathcal{E}_Z(\tau) \mid Y_i = Y_j)
\ge \mathbb{P}\!\left(
\mathcal{E}_V\!\left( \frac{\tau - 2\epsilon}{L_V} \right)
\mid Y_i = Y_j
\right).
\end{align}

Repeating the same argument for $g(X)$ with Lipschitz constant $L_S$ gives
\begin{align}\label{use_3}
\mathbb{P}(\mathcal{E}_Z(\tau) \mid Y_i = Y_j)
\ge \mathbb{P}\!\left(
\mathcal{E}_S\!\left( \frac{\tau - 2\epsilon}{L_S} \right)
\mid Y_i = Y_j
\right).
\end{align}

Define for modality $\ast \in \{V,S\}$ the maximal probability drop due to
threshold rescaling:
\[
\delta_\ast := \sup_{\tau>0}\,
\Bigl[ \,
\mathbb{P}(\mathcal{E}_\ast(\tau) \mid Y_i = Y_j)
- \mathbb{P}\!\left(
\mathcal{E}_\ast\!\left( \frac{\tau - 2\epsilon}{L_\ast} \right)
\mid Y_i = Y_j
\right) \Bigr]_+,
\]
where $[\,\cdot\,]_+ = \max\{0,\cdot\}$. By construction $\delta_\ast \ge 0$, and
\[
\mathbb{P}\!\left(
\mathcal{E}_\ast\!\left( \frac{\tau - 2\epsilon}{L_\ast} \right)
\mid Y_i = Y_j
\right)
\ge \mathbb{P}(\mathcal{E}_\ast(\tau) \mid Y_i = Y_j) - \delta_\ast.
\]

Combining (\ref{use_2}), (\ref{use_3}) and the definitions of $\delta_V,\delta_S$,
\[
\begin{aligned}
\mathbb{P}(\mathcal{E}_Z(\tau) \mid Y_i = Y_j)
&\ge \max\Bigl\{
\mathbb{P}\!\left(
\mathcal{E}_V\!\left( \frac{\tau - 2\epsilon}{L_V} \right)
\mid Y_i = Y_j
\right),\;
\mathbb{P}\!\left(
\mathcal{E}_S\!\left( \frac{\tau - 2\epsilon}{L_S} \right)
\mid Y_i = Y_j
\right) \Bigr\} \\
&\ge \max\Bigl\{
\mathbb{P}(\mathcal{E}_V(\tau) \mid Y_i = Y_j) - \delta_V,\;
\mathbb{P}(\mathcal{E}_S(\tau) \mid Y_i = Y_j) - \delta_S
\Bigr\}.
\end{aligned}
\]

Setting $\delta := \max\{\delta_V, \delta_S\}$ we obtain
\[
\mathbb{P}(\mathcal{E}_Z(\tau) \mid Y_i = Y_j)
\ge \max\Bigl\{
\mathbb{P}(\mathcal{E}_V(\tau) \mid Y_i = Y_j),\;
\mathbb{P}(\mathcal{E}_S(\tau) \mid Y_i = Y_j)
\Bigr\} - \delta,
\]
which completes the proof.
\end{proof}

\subsection{Empirical Validation of Theoretical Analysis}
\label{app:theory_validation}

To examine the practical relevance of the theoretical
conditions, we conduct empirical analyses of the
complementarity between global semantic and local
relational representations, intra-class proximity
preservation, and neighborhood consistency.
Ground-truth labels are used exclusively for
post-training diagnostics and are not involved
in the optimization of FSRC.

\subsubsection{Empirical Examination of Theoretical Conditions}

We evaluate the empirical complementarity deviation $\eta_{\mathrm{emp}}$ associated with Assumption~\ref{assump:weak-complementary}. Specifically, we estimate
\begin{equation}
\begin{aligned}
\eta_{\mathrm{emp}}
=\Big[
&\mathbb{P}(\mathcal{E}_V(\tau)\cap
\mathcal{E}_S(\tau)\mid Y_i\neq Y_j)
-\mathbb{P}(\mathcal{E}_V(\tau)\mid Y_i\neq Y_j)
\mathbb{P}(\mathcal{E}_S(\tau)\mid Y_i\neq Y_j)
\Big]_+,
\end{aligned}
\label{eq:eta_emp}
\end{equation}
where $[a]_+=\max\{a,0\}$. The proximity threshold $\tau$ is determined from the 20th percentile of unlabeled pairwise distances.
Table~\ref{tab:theory_validation} reports $\eta_{\mathrm{emp}}$ and the corresponding upper bound required by Theorem~\ref{thm:simplified}
across all 18 benchmark datasets.

\begin{table}[t]
\centering
\caption{Empirical examination of the complementarity
condition across 18 datasets. The empirical deviation
$\eta_{\mathrm{emp}}$ remains below the corresponding
upper bound required by Theorem~\ref{thm:simplified}.
}
\label{tab:theory_validation}
\renewcommand{\arraystretch}{1.1}
\setlength{\tabcolsep}{6pt}
\begin{tabular}{lcc}
\toprule
Dataset
& $\eta_{\mathrm{emp}}$
& Required Bound \\
\midrule
MNIST         & 0.01 & 0.13 \\
GTSRB         & 0.04 & 0.15 \\
ImageNet-10   & 0.00 & 0.05 \\
OxfordPets    & 0.04 & 0.10 \\
ImageNet-Dogs & 0.02 & 0.06 \\
DTD           & 0.03 & 0.11 \\
CIFAR-10      & 0.01 & 0.10 \\
CIFAR-100     & 0.02 & 0.11 \\
TinyImageNet  & 0.02 & 0.16 \\
Caltech101    & 0.02 & 0.10 \\
FER2013       & 0.00 & 0.16 \\
EuroSAT       & 0.02 & 0.10 \\
RESISC45      & 0.03 & 0.13 \\
UCF101        & 0.03 & 0.16 \\
CLEVR         & 0.04 & 0.12 \\
USPS          & 0.01 & 0.11 \\
Fashion       & 0.04 & 0.11 \\
ImageNet-1K   & 0.01 & 0.10 \\
\bottomrule
\end{tabular}
\end{table}

As shown in Table~\ref{tab:theory_validation},
$\eta_{\mathrm{emp}}$ is below its corresponding
required bound on all 18 datasets.
For example, the more precise estimates on
ImageNet-Dogs, CIFAR-10, and OxfordPets are
0.0171, 0.0138, and 0.0397, respectively,
which are below the corresponding bounds
of 0.0608, 0.0970, and 0.0964.
These results provide empirical support
for the complementarity deviation $\eta$
across diverse datasets.

We further examine the intra-class proximity
preservation characterized by
Theorem~\ref{thm:simplified-intra}.
Specifically, we define the observed
intra-class proximity degradation as
\begin{equation}
\begin{aligned}
\delta_{\mathrm{obs}}(\tau)
=\Big[
\max\Big\{
&\mathbb{P}(\mathcal{E}_V(\tau)\mid Y_i=Y_j),
\mathbb{P}(\mathcal{E}_S(\tau)\mid Y_i=Y_j)
\Big\}
-\mathbb{P}(\mathcal{E}_Z(\tau)\mid Y_i=Y_j)
\Big]_+.
\end{aligned}
\label{eq:delta_obs}
\end{equation}

Under the same evaluation protocol, we obtain
$\delta_{\mathrm{obs}}(\tau)=0$ on all 18 datasets,
indicating that the distilled representations
exhibit no observed degradation in intra-class
proximity relative to the better of the two
source representations.

\subsubsection{Inter-Class Error Complementarity}

To further investigate the complementarity
between the two source representations,
we measure the proportion of inter-class
proximity errors made by the global semantic
representation that are avoided by the
local relational representation.
Specifically, the error avoidance rate is
defined as
\begin{equation}
R_{\mathrm{avoid}}
=
\mathbb{P}\big(
\mathcal{E}_S(\tau)^c
\mid
\mathcal{E}_V(\tau),Y_i\neq Y_j
\big),
\label{eq:error_avoid}
\end{equation}
where $\mathcal{E}_S(\tau)^c$ denotes the complement of the local proximity event.

The resulting avoidance rates are
84.4\%, 81.0\%, and 70.1\% on
ImageNet-Dogs, CIFAR-10, and OxfordPets,
respectively.
These results indicate that the global
semantic and local relational representations
exhibit substantially different inter-class
proximity error patterns, providing additional
empirical evidence for their complementarity.

\subsubsection{Neighborhood Consistency Analysis}

We further examine the closeness relation
used in the theoretical analysis:
\begin{equation}
\mathcal{E}_Z
\subseteq
\mathcal{E}_V\cap\mathcal{E}_S.
\label{eq:closeness_relation}
\end{equation}

This relation expresses the intuition that
sample pairs becoming close in the distilled
representation should also maintain proximity
in both the global semantic and local
relational representation spaces.
To empirically examine this intuition,
we adopt neighborhood trustworthiness \citep{venna2001neighborhood},
which measures whether neighbors formed
in the distilled representation remain
relatively close in the original
representation spaces.
Specifically, $T(V,Z)$ measures the
neighborhood trustworthiness of the distilled
features $Z$ with respect to the global
semantic embeddings $V$, while $T(S,Z)$
measures the corresponding consistency with
the local relational embeddings $S$.
A value closer to 1 indicates fewer instances where samples are close in $Z$ but ranked far apart in the original space $S$.

We evaluate neighborhood trustworthiness using 20 nearest
neighbors on ImageNet-Dogs and CIFAR-10. ImageNet-Dogs
achieves $T(V,Z)=0.980$ and $T(S,Z)=0.980$, while CIFAR-10
achieves $T(V,Z)=0.996$ and $T(S,Z)=0.936$.
These results indicate that the distilled representation
maintains high neighborhood consistency with both source
representations, providing empirical support for the
closeness intuition underlying our theoretical analysis.

\section{Embedding Fusion Effectiveness Analysis on K-Means}
\label{D}

To validate the generality of the K-Means results reported in the main paper, we extend the analysis to 6 other backbone architectures beyond the primary CLIP ViT-L/14: two CLIP Vision Transformers (ViT-B/32, ViT-B/16), three CLIP ResNets (ResNet50, ResNet101, ResNet50x4), and DINOv2 ViT-G/14. For each backbone, we evaluate K-Means clustering performance using both the original global semantic embeddings and the distilled embeddings. This experiment investigates whether fusing local relational features with global semantic ones consistently improves clustering performance across diverse model families, complementing the feature-separability perspective offered by triplet accuracy. The analysis is conducted on the same 18 benchmark datasets. The results reported are obtained from individual runs.

As summarized in Table~\ref{tab6}, the distilled embeddings consistently outperform the original ones across most datasets and backbones. This robust improvement demonstrates that FSRC’s feature fusion is generalizable across different foundation models and is not architecture-specific.

\begin{table}[t]
\centering
\caption{Clustering accuracy comparison across 7 backbones on the K-Means. Results are reported under original embeddings and distilled embeddings. The \underline{underlined} value indicates the best-performing method within each backbone.}
\label{tab6}
\renewcommand{\arraystretch}{1.25}
\setlength{\tabcolsep}{4pt}
\resizebox{\textwidth}{!}{
\begin{tabular}{c|cc|cc|cc|cc|cc|cc|cc}
\toprule
 & \multicolumn{2}{c}{ClipVIT-B/32}
 & \multicolumn{2}{c}{ClipVIT-B/16}
 & \multicolumn{2}{c}{ClipVIT-L/14}
 & \multicolumn{2}{c}{ClipResNet50}
 & \multicolumn{2}{c}{ClipResNet101}
 & \multicolumn{2}{c}{ClipResNet50*4}
 & \multicolumn{2}{c}{DINOV2} \\
\cmidrule(lr){2-3} \cmidrule(lr){4-5} \cmidrule(lr){6-7}
\cmidrule(lr){8-9} \cmidrule(lr){10-11} \cmidrule(lr){12-13}
\cmidrule(lr){14-15}
  & Original & Distilled 
 & Original & Distilled
 & Original & Distilled
 & Original & Distilled
 & Original & Distilled
 & Original & Distilled
 & Original & Distilled
 \\
\midrule
MNIST      & 56.51 & \underline{89.13} & 55.18 & \underline{89.81} & 67.32 & \underline{90.06} & 50.12 & \underline{95.02} & 49.28 & \underline{87.52} & 45.76 & \underline{88.88} & 46.19 & \underline{74.71} \\
GTSRB      & 33.64 & \underline{37.79} & \underline{35.54} & 34.91 & \underline{49.06} & 47.30 & 26.66 & \underline{30.87} & 28.69 & \underline{33.19} & 26.10 & \underline{35.06} & 20.71 & \underline{31.14} \\
ImageNet-10& 98.69 & \underline{99.74} & 99.22 & \underline{99.79} & 99.63 & \underline{99.90} & 96.97 & \underline{99.76} & 98.46 & \underline{99.75} & 98.93 & \underline{99.75} & 91.84 & \underline{99.01} \\
OxfordPets & 52.61 & \underline{87.47} & 30.38 & \underline{47.31} & 69.97 & \underline{92.60} & 47.77 & \underline{83.86} & 57.77 & \underline{86.63} & 60.57 & \underline{84.48} & 82.23 & \underline{88.97} \\
ImageNet-Dogs
           & 39.80 & \underline{90.59} & 42.57 & \underline{91.75} & 62.99 & \underline{89.40} & 33.25 & \underline{88.69} & 46.47 & \underline{88.81} & 47.12 & \underline{88.92} & 87.52 & \underline{97.39} \\
DTD        & 46.38 & \underline{52.61} & 47.77 & \underline{56.52} & 50.88 & \underline{59.80} & 39.65 & \underline{52.63} & 41.86 & \underline{53.62} & 45.88 & \underline{57.55} & 47.55 & \underline{55.90} \\
CIFAR-10   & 76.11 & \underline{93.71} & 78.31 & \underline{95.25} & 83.54 & \underline{97.30} & 55.27 & \underline{81.70} & 69.52 & \underline{87.79} & 65.36 & \underline{87.56} & 74.92 & \underline{98.24} \\
CIFAR-100  & 42.39 & \underline{55.90} & 47.82 & \underline{58.43} & 53.64 & \underline{64.80} & 25.04 & \underline{38.20} & 34.99 & \underline{44.16} & 30.36 & \underline{45.27} & 68.47 & \underline{73.16} \\
TinyImageNet
           & 40.85 & \underline{47.74} & 43.37 & \underline{50.71} & 57.49 & \underline{59.30} & 25.22 & \underline{35.20} & 34.40 & \underline{40.35} & 34.72 & \underline{38.92} & \underline{72.19} & 69.67 \\
Caltech101 & 72.29 & \underline{75.75} & 76.99 & \underline{78.92} & \underline{84.51} & 83.10 & 62.06 & \underline{70.85} & 69.22 & \underline{73.43} & 73.24 & \underline{75.23} & 85.82 & \underline{88.37} \\
Fer2013    & 26.95 & \underline{30.04} & 27.82 & \underline{30.35} & 31.48 & \underline{33.10} & 26.42 & \underline{28.77} & 26.79 & \underline{30.59} & 27.65 & \underline{29.01} & \underline{32.98} & 32.82 \\
EuroSAT    & 64.14 & \underline{78.74} & 73.98 & \underline{93.73} & 74.99 & \underline{96.00} & 56.83 & \underline{70.99} & 52.31 & \underline{91.11} & 53.39 & \underline{85.35} & 62.36 & \underline{95.22} \\
RESISC45   & 68.19 & \underline{77.29} & 73.36 & \underline{80.56} & 73.87 & \underline{80.30} & 54.29 & \underline{74.77} & 62.56 & \underline{75.56} & 66.54 & \underline{76.71} & 62.65 & \underline{74.46} \\
UCF101     & 61.34 & \underline{66.13} & 63.53 & \underline{70.92} & 70.94 & \underline{75.90} & 51.87 & \underline{64.20} & 59.64 & \underline{66.70} & 61.85 & \underline{68.50} & 71.33 & \underline{77.28} \\
CLEVR      &  \underline{24.95} & 23.65 & \underline{24.30} & 23.25 & \underline{23.50} & 22.40 & \underline{24.05} & 22.20 & \underline{26.05} & 21.35 & \underline{25.30} & 25.00 & 20.80 & \underline{20.80} \\
USPS       & 60.76 & \underline{80.46} & 78.47 & \underline{77.55} & 79.66 & \underline{85.90} & 61.42 & \underline{78.07} & 51.28 & \underline{81.02} & 51.83 & \underline{76.01} & 56.38 & \underline{87.03} \\
Fashion    & 64.16 & \underline{70.84} & 58.96 & \underline{72.96} & 53.19 & \underline{68.20} & 58.78 & \underline{74.25} & 63.83 & \underline{72.41} & 56.05 & \underline{67.08} & 65.37 & \underline{71.80} \\
ImageNet-1K
           & 38.24 & \underline{47.03} & 43.09 & \underline{52.83} & 52.19 & \underline{60.30} & 32.64 & \underline{44.75} & 39.28 & \underline{47.54} & 42.06 & \underline{50.91} & 62.12 & \underline{62.94} \\
\midrule
\textbf{AVG.}
 & 53.78 & \underline{66.92}
 & 55.59 & \underline{66.98}
 & 63.27 & \underline{72.54}
 & 46.02 & \underline{63.04}
 & 50.69 & \underline{65.64}
 & 50.71 & \underline{65.57}
 & 61.75 & \underline{72.16} \\
\bottomrule
\end{tabular}}
\end{table}

\section{Embedding Fusion Effectiveness Analysis on Clustering Assignment }
\label{E}

To rigorously assess the generalizability of the clustering assignment module results, we extend the evaluation to six additional backbone architectures beyond the primary CLIP ViT-L/14: CLIP ViT-B/32, CLIP ViT-B/16, CLIP ResNet50, CLIP ResNet101, CLIP ResNet50x4, and DINOv2 ViT-G/14. For each backbone, we apply the same clustering-assignment model with iterative self-labeling to both the original global semantic embeddings and the FSRC-distilled embeddings, maintaining a consistent evaluation protocol across all 18 benchmark datasets. Clustering accuracy (ACC) serves as the primary comparison metric.

As detailed in Table~\ref{tab7}, the distilled embeddings achieve superior clustering accuracy compared to the original embeddings across the vast majority of dataset-backbone combinations. This consistent and substantial improvement reproduces the core performance trend initially observed with CLIP ViT-L/14. The results conclusively demonstrate that the performance gains delivered by FSRC are not model-specific but are generalizable and robust across diverse foundation models.

\begin{table}[t]
\centering
\caption{Clustering accuracy comparison across 7 backbones on the Clustering Assignment Model. Results are reported under original embeddings and distilled embeddings. The \underline{underlined} value indicates the best-performing method within each backbone. The results reported are obtained from individual runs.}
\label{tab7}
\renewcommand{\arraystretch}{1.25}
\setlength{\tabcolsep}{4pt}
\resizebox{\textwidth}{!}{
\begin{tabular}{c|cc|cc|cc|cc|cc|cc|cc}
\toprule
 & \multicolumn{2}{c}{ClipVIT-B/32} 
 & \multicolumn{2}{c}{ClipVIT-B/16} 
 & \multicolumn{2}{c}{ClipVIT-L/14} 
 & \multicolumn{2}{c}{ClipResNet50} 
 & \multicolumn{2}{c}{ClipResNet101} 
 & \multicolumn{2}{c}{ClipResNet50*4} 
 & \multicolumn{2}{c}{DINOV2} \\
\cmidrule(lr){2-3} \cmidrule(lr){4-5} \cmidrule(lr){6-7}
\cmidrule(lr){8-9} \cmidrule(lr){10-11} \cmidrule(lr){12-13} \cmidrule(lr){14-15}
  & Original & Distilled 
 & Original & Distilled
 & Original & Distilled
 & Original & Distilled
 & Original & Distilled
 & Original & Distilled
 & Original & Distilled
 \\
\midrule
MNIST      & 56.53 & \underline{89.61} & 54.74 & \underline{90.43} & 74.76 & \underline{90.18} & 35.16 & \underline{96.48} & 41.44 & \underline{90.11} & 40.18 & \underline{89.19} & 52.30 & \underline{76.63} \\
GTSRB      & 34.50 & \underline{38.56} & \underline{37.59} & 36.49 & \underline{49.07} & 45.33 & 22.73 & \underline{30.52} & 18.95 & \underline{34.62} & 26.37 & \underline{34.77} & 22.22 & \underline{31.85} \\
ImageNet-10& 99.36 & \underline{99.76} & 99.58 & \underline{99.82} & 99.80 & \underline{99.86} & 98.43 & \underline{99.77} & 99.02 & \underline{99.74} & 99.53 & \underline{99.74} & 96.86 & \underline{99.80} \\
OxfordPets & 63.83 & \underline{92.36} & 36.63 & \underline{49.46} & 88.02 & \underline{91.30} & 34.54 & \underline{90.65} & 42.42 & \underline{92.64} & 58.75 & \underline{93.10} & 89.21 & \underline{94.48} \\
ImageNet-Dogs
           & 49.58 & \underline{92.97} & 49.58 & \underline{91.94} & 74.43 & \underline{97.44} & 37.97 & \underline{90.51} & 43.39 & \underline{90.08} & 51.38 & \underline{91.96} & 94.86 & \underline{97.81} \\
DTD        & 52.70 & \underline{57.13} & 55.13 & \underline{56.68} & 56.52 & \underline{60.21} & 42.10 & \underline{49.87} & 39.36 & \underline{55.13} & 52.58 & \underline{57.21} & 53.99 & \underline{55.51} \\
CIFAR-10   & 86.55 & \underline{94.63} & 94.04 & \underline{95.81} & 97.58 & \underline{97.76} & 60.90 & \underline{85.90} & 72.12 & \underline{90.71} & 68.70 & \underline{90.42} & 90.32 & \underline{98.51} \\
CIFAR-100  & 46.44 & \underline{57.85} & 51.31 & \underline{63.46} & 62.27 & \underline{72.18} & 15.13 & \underline{43.46} & 18.25 & \underline{48.74} & 18.56 & \underline{48.74} & 80.90 & \underline{81.16} \\
TinyImageNet
           & 44.15 & \underline{49.56} & 45.65 & \underline{53.52} & 64.66 & \underline{67.47} & 11.17 & \underline{36.54} & 13.97 & \underline{42.51} & 16.11 & \underline{41.42} & \underline{80.91} & 76.17 \\
Caltech101 & 86.11 & \underline{86.93} & 88.73 & \underline{89.51} & \underline{94.87} & 92.22 & 68.63 & \underline{80.82} & 67.42 & \underline{85.29} & 83.56 & \underline{88.04} & 88.89 & \underline{92.81} \\
Fer2013    & \underline{31.38} & 30.96 & 28.99 & \underline{31.22} & \underline{32.93} & 32.91 & \underline{30.28} & 28.81 & 29.12 & \underline{31.30} & 28.58 & \underline{30.07} & 32.81 & \underline{33.56} \\
EuroSAT    & 68.55 & \underline{79.28} & 79.59 & \underline{94.27} & 86.60 & \underline{96.29} & 49.89 & \underline{72.27} & 54.31 & \underline{89.54} & 52.93 & \underline{88.89} & 90.61 & \underline{95.89} \\
RESISC45   & 74.56 & \underline{80.07} & 76.05 & \underline{83.83} & 83.14 & \underline{88.27} & 36.72 & \underline{76.12} & 39.27 & \underline{77.45} & 52.44 & \underline{78.43} & 71.98 & \underline{76.30} \\
UCF101     & 66.66 & \underline{71.79} & 72.15 & \underline{73.44} & 77.75 & \underline{80.38} & 53.09 & \underline{64.71} & 47.89 & \underline{68.59} & 64.54 & \underline{71.72} & 77.26 & \underline{78.10} \\
CLEVR      & 25.10 & \underline{25.45} & 24.60 & \underline{24.63} & \underline{26.50} & 23.35 & 21.90 & \underline{22.90} & \underline{26.05} & 21.90 & \underline{24.55} & 18.70 & 20.80 & \underline{20.80} \\
USPS       & 59.00 & \underline{78.23} & 71.13 & \underline{76.49} & 77.16 & \underline{81.44} & 52.00 & \underline{77.04} & 39.57 & \underline{69.87} & 47.25 & \underline{76.30} & 61.97 & \underline{81.42} \\
Fashion    & 68.43 & \underline{70.01} & 66.31 & \underline{72.69} & 67.93 & \underline{69.56} & 48.10 & \underline{75.09} & 52.27 & \underline{75.77} & 50.29 & \underline{72.00} & 78.71 & \underline{79.72} \\
ImageNet-1K
           & 41.93 & \underline{50.36} & 48.64 & \underline{57.31} & 56.60 & \underline{67.90} & 39.82 & \underline{47.88} & 42.35 & \underline{49.88} & 46.38 & \underline{55.05} & 68.90 & \underline{70.91} \\
\hline
\textbf{AVG.}
 & 58.63 & \underline{69.19}
 & 60.02 & \underline{68.94}
 & 70.59 & \underline{75.23}
 & 42.14 & \underline{64.96}
 & 43.73 & \underline{67.44}
 & 49.04 & \underline{68.10}
 & 69.64 & \underline{74.52} \\
\bottomrule
\end{tabular}}
\end{table}

\section{Successful and Failed Cases Analysis}
\label{P}
To further analyze the successful and failed cases of incorporating local relational features, we conduct a diagnostic analysis from two perspectives: category consistency with ground-truth labels and the ability to correct clustering errors made by global semantic embeddings. Specifically, category consistency is measured by the average proportion of each sample's top-20 nearest neighbors that share its ground-truth label. The local correction rate is defined as the proportion of samples incorrectly clustered using the original global representations that are correctly clustered using the distilled representations obtained after mutual distillation.

As shown in Table~\ref{tab:local_complementarity}, on ImageNet-Dogs and OxfordPets, where local relational features contribute substantially to clustering performance, local representations exhibit higher category consistency than global semantic representations, together with relatively high correction rates. These results indicate that local relational features provide substantial complementary information. On CLEVR and GTSRB, where local features yield limited improvements, local representations struggle to capture category-consistent structures on CLEVR, while global semantic representations already exhibit high category consistency on GTSRB, leaving limited room for additional complementary information. 
Overall, these findings suggest that the effectiveness of local relational features depends on both their ability to capture category-consistent structures and their complementarity to global semantic representations.

\begin{table}[t]
\centering
\caption{Neighborhood category consistency and local correction rate (\%) on four representative datasets.}
\label{tab:local_complementarity}
\renewcommand{\arraystretch}{1.15}
\setlength{\tabcolsep}{5pt}
\begin{tabular}{lccc}
\toprule
Dataset & Global Consistency & Local Consistency & Correction Rate \\
\midrule
ImageNet-Dogs & 79.8 & 84.8 & 80.9 \\
OxfordPets & 76.5 & 84.2 & 77.6 \\
CLEVR & 25.5 & 19.6 & 19.5 \\
GTSRB & 89.4 & 78.0 & 22.7 \\
\bottomrule
\end{tabular}
\end{table}

\section{Clustering Stability on 18 Datasets Across 7 Backbones}
\label{G}

\begin{figure*}[t]
\centering
\begin{subfigure}[b]{0.16\textwidth}
    \centering
    \includegraphics[width=\textwidth]{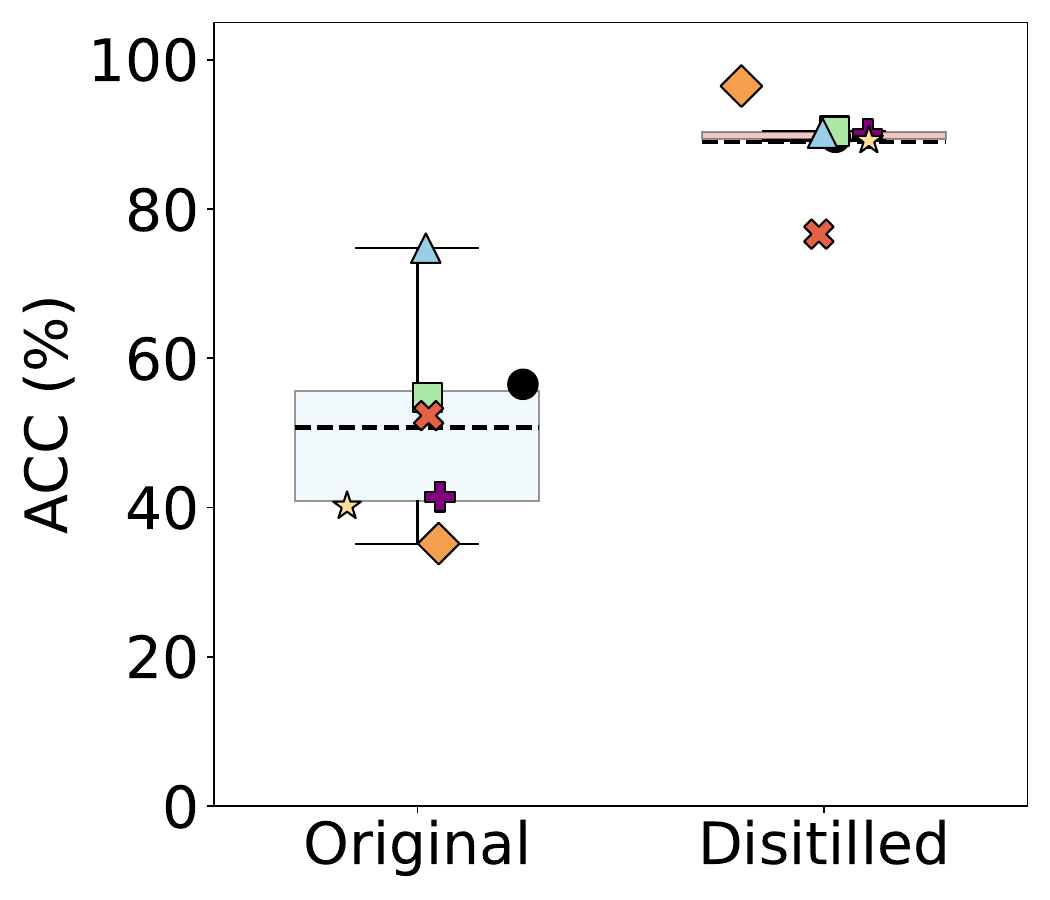}
    \caption{MNIST}
\end{subfigure}
\begin{subfigure}[b]{0.16\textwidth}
    \centering
    \includegraphics[width=\textwidth]{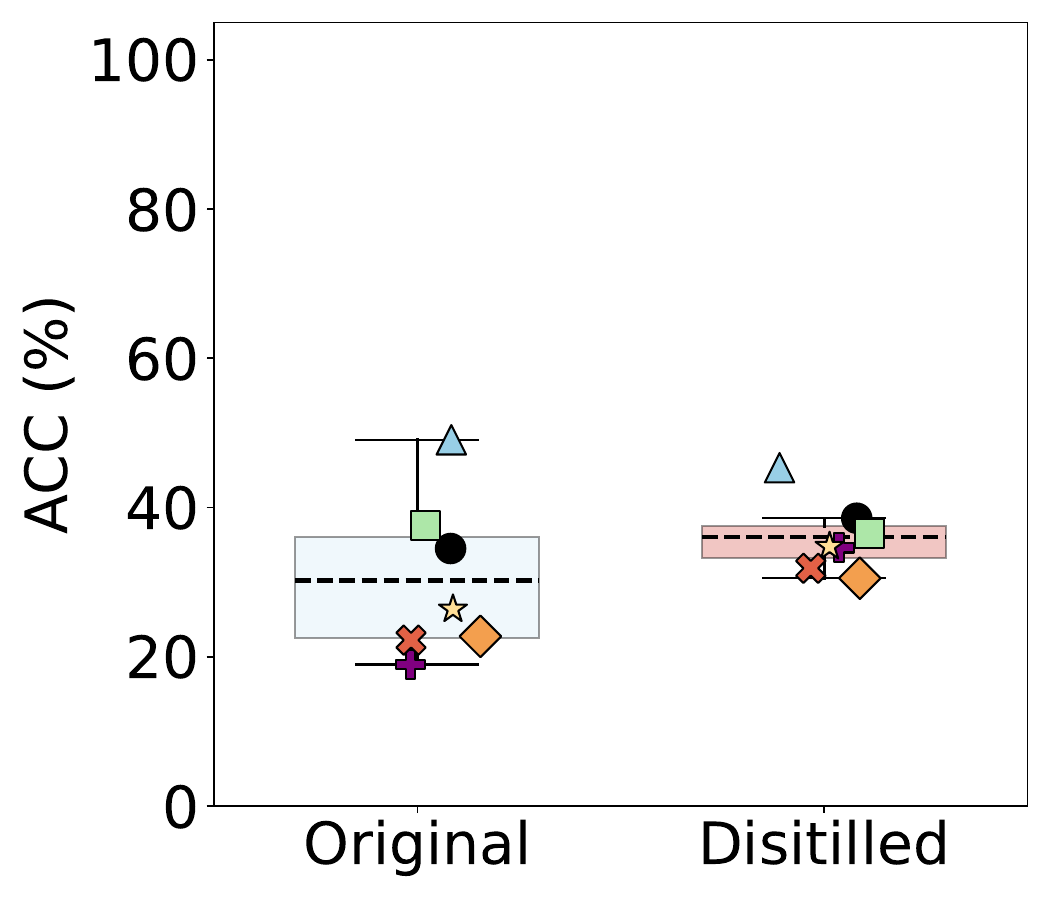}
    \caption{GTSRB}
\end{subfigure}
\begin{subfigure}[b]{0.16\textwidth}
    \centering
    \includegraphics[width=\textwidth]{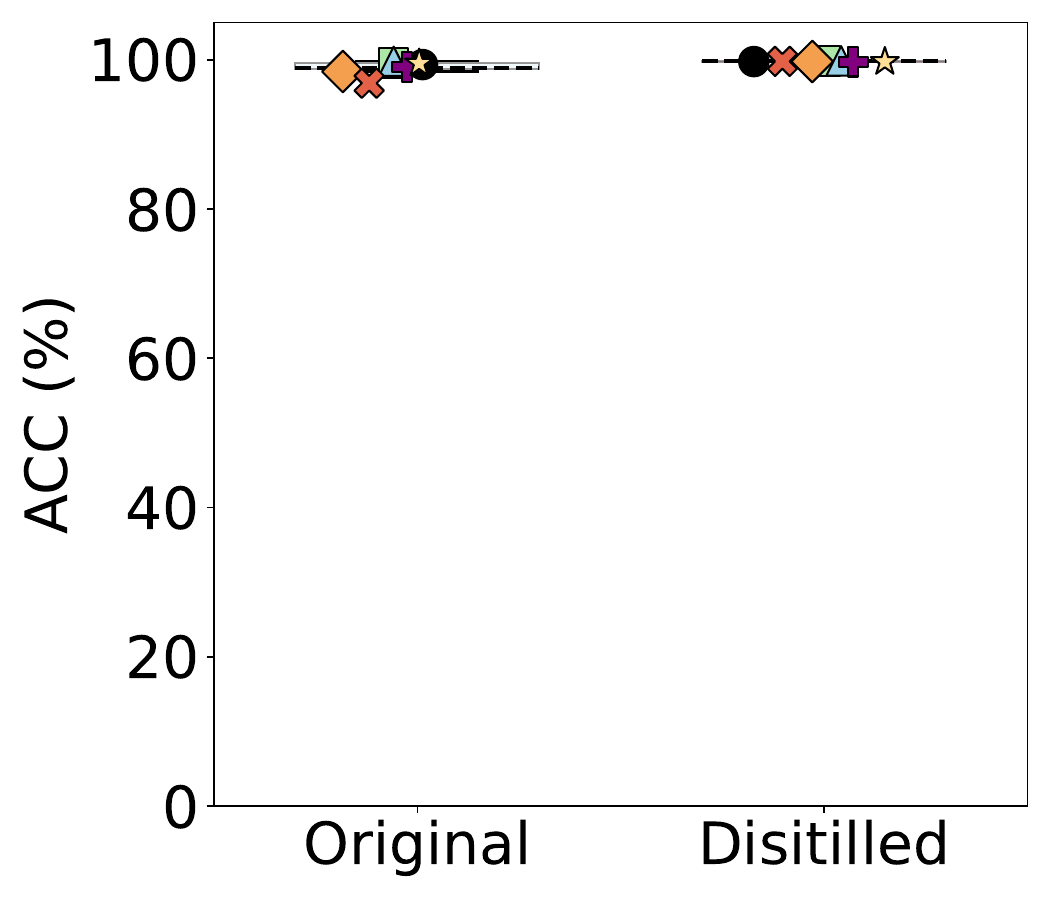}
    \caption{ImageNet-10}
\end{subfigure}
\begin{subfigure}[b]{0.16\textwidth}
    \centering
    \includegraphics[width=\textwidth]{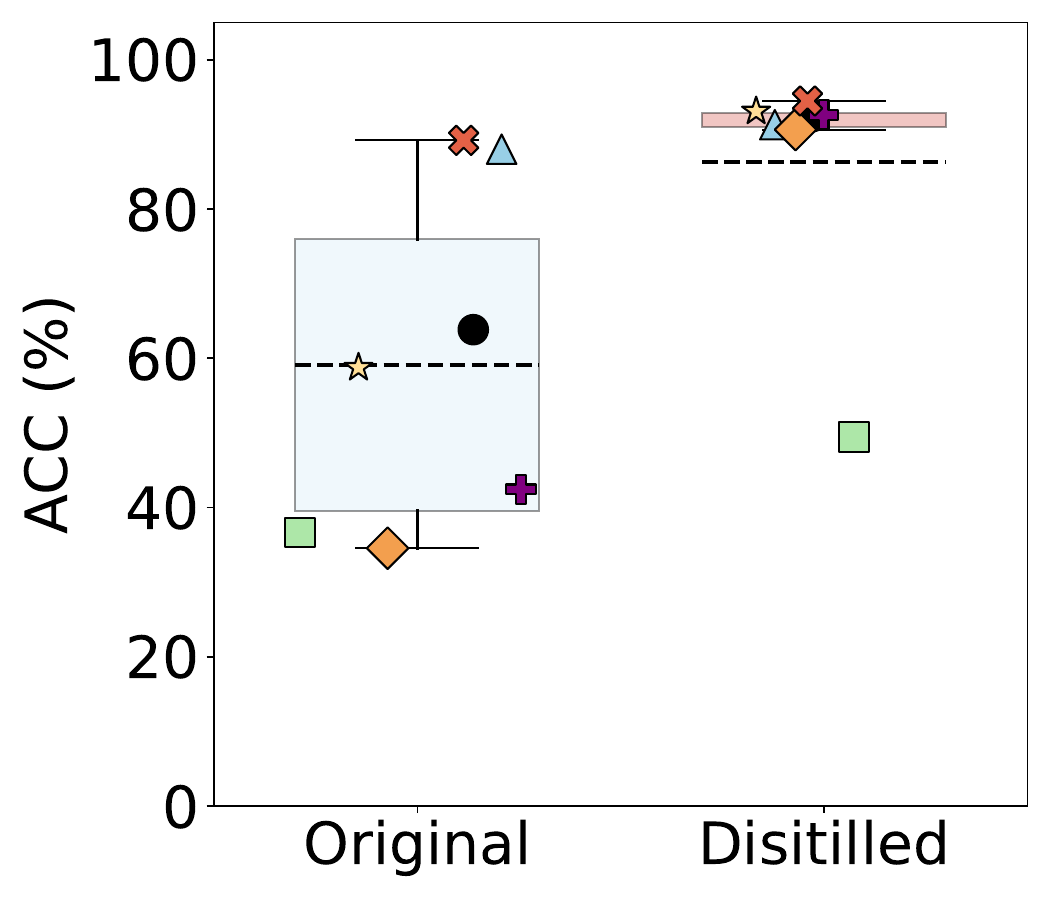}
    \caption{OxfordPets}
\end{subfigure}
\begin{subfigure}[b]{0.16\textwidth}
    \centering
    \includegraphics[width=\textwidth]{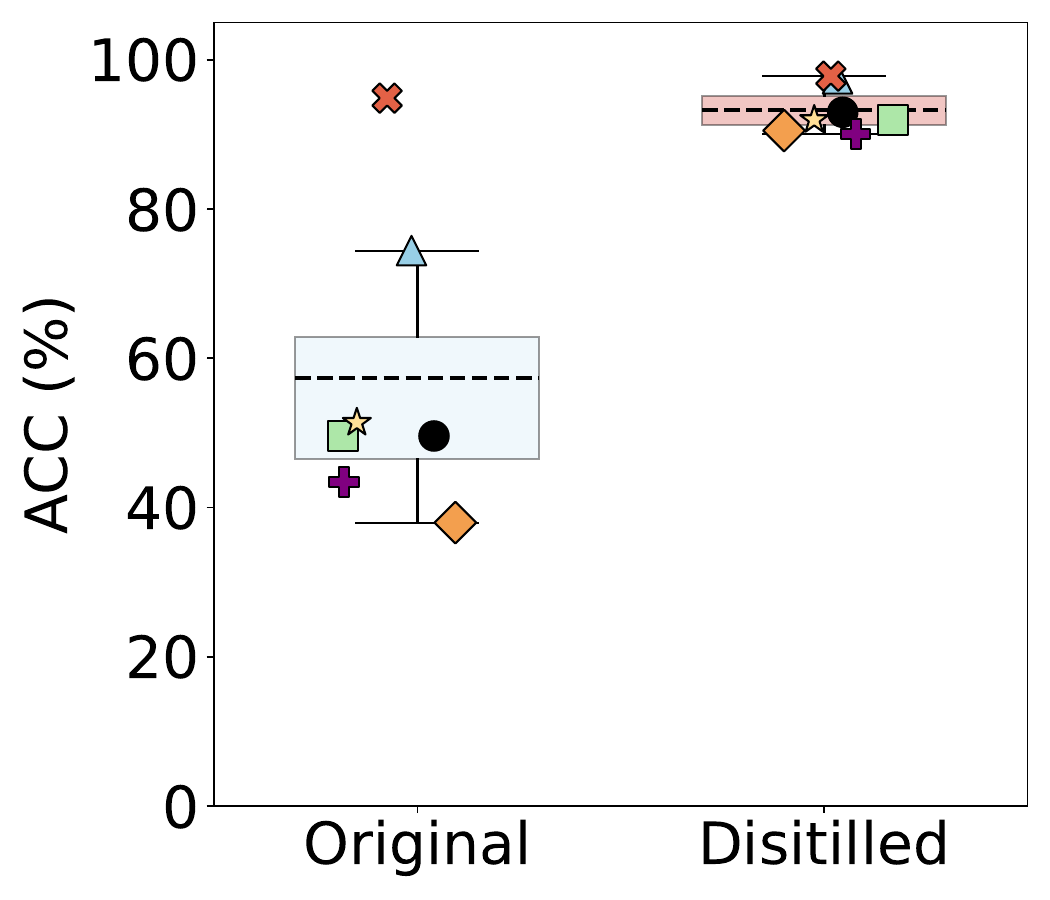}
    \caption{ImageNet-Dogs}
\end{subfigure}
\begin{subfigure}[b]{0.16\textwidth}
    \centering
    \includegraphics[width=\textwidth]{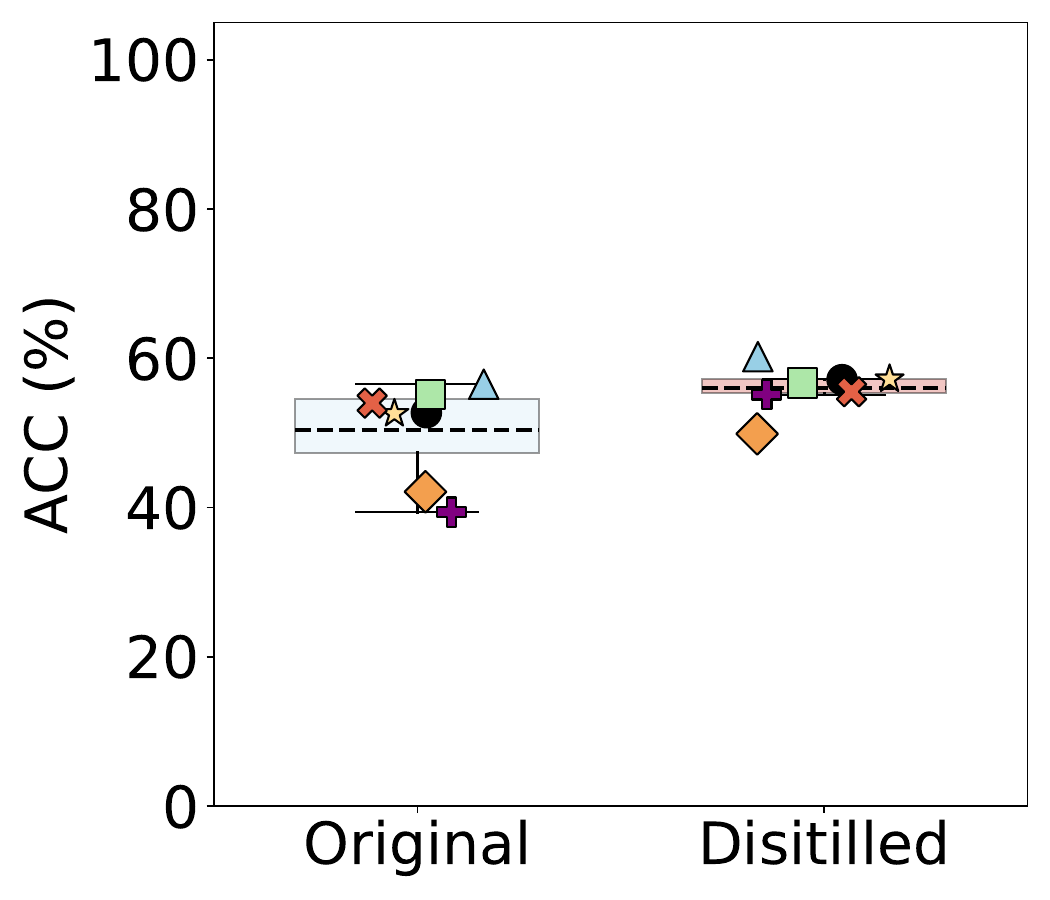}
    \caption{DTD}
\end{subfigure}

\begin{subfigure}[b]{0.16\textwidth}
    \centering
    \includegraphics[width=\textwidth]{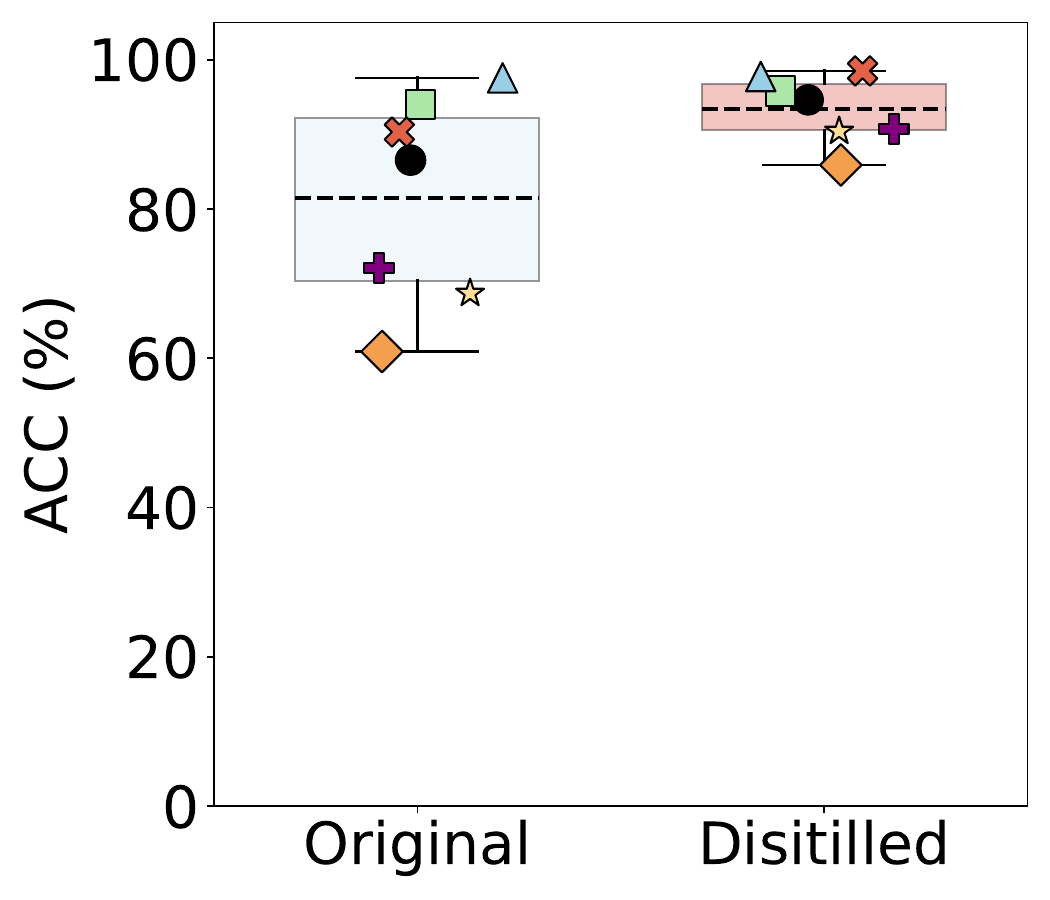}
    \caption{CIFAR-10}
\end{subfigure}
\begin{subfigure}[b]{0.16\textwidth}
    \centering
    \includegraphics[width=\textwidth]{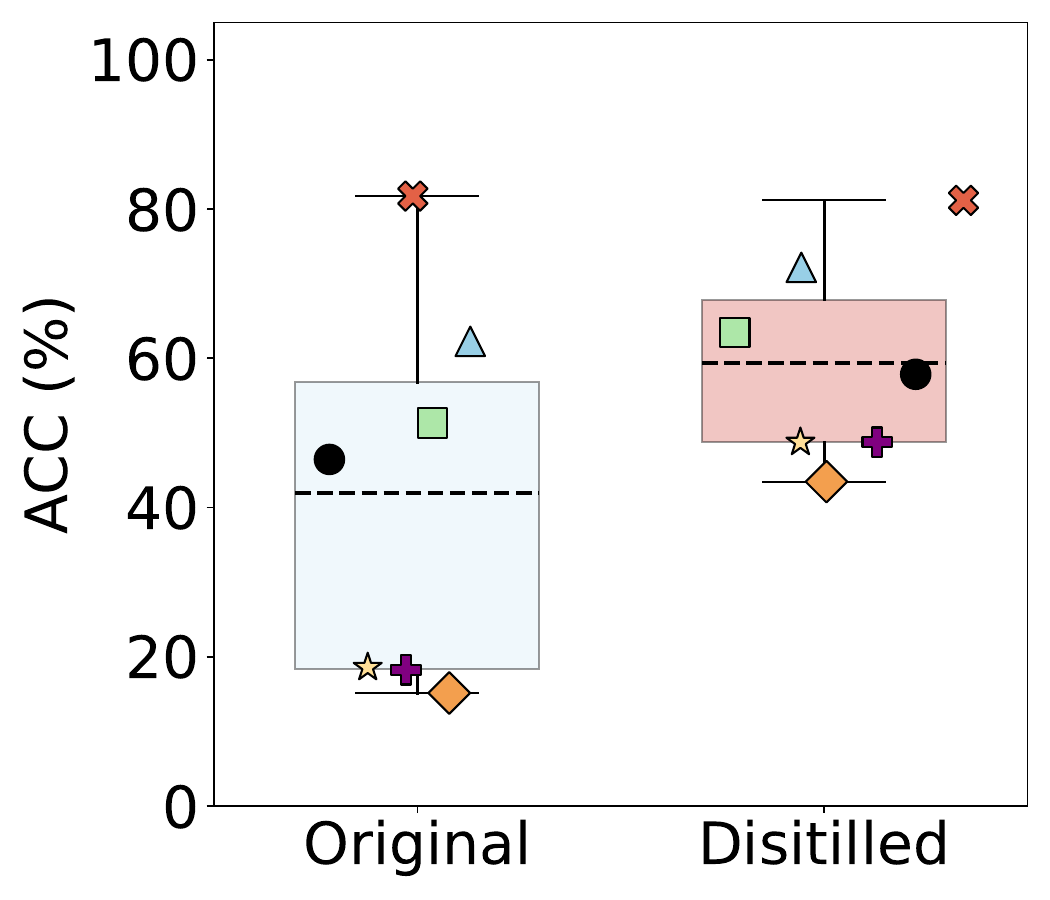}
    \caption{CIFAR-100}
\end{subfigure}
\begin{subfigure}[b]{0.16\textwidth}
    \centering
    \includegraphics[width=\textwidth]{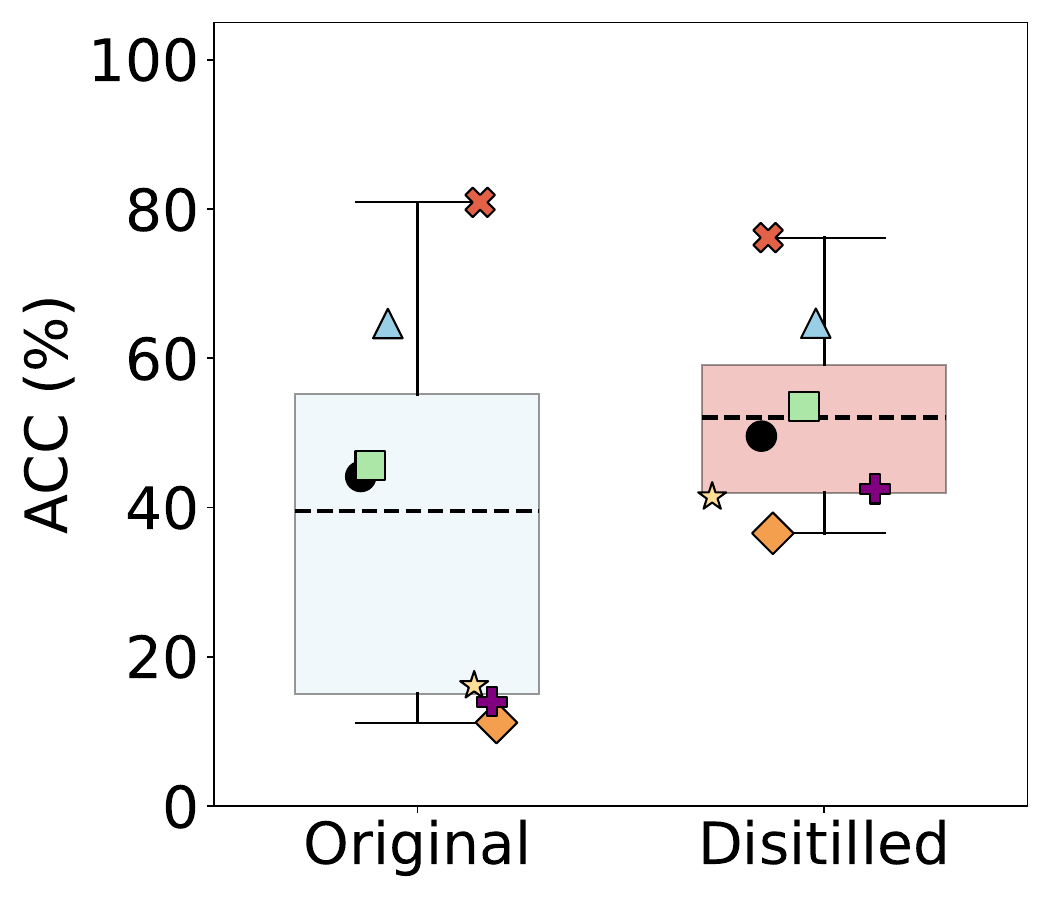}
    \caption{TinyImageNet}
\end{subfigure}
\begin{subfigure}[b]{0.16\textwidth}
    \centering
    \includegraphics[width=\textwidth]{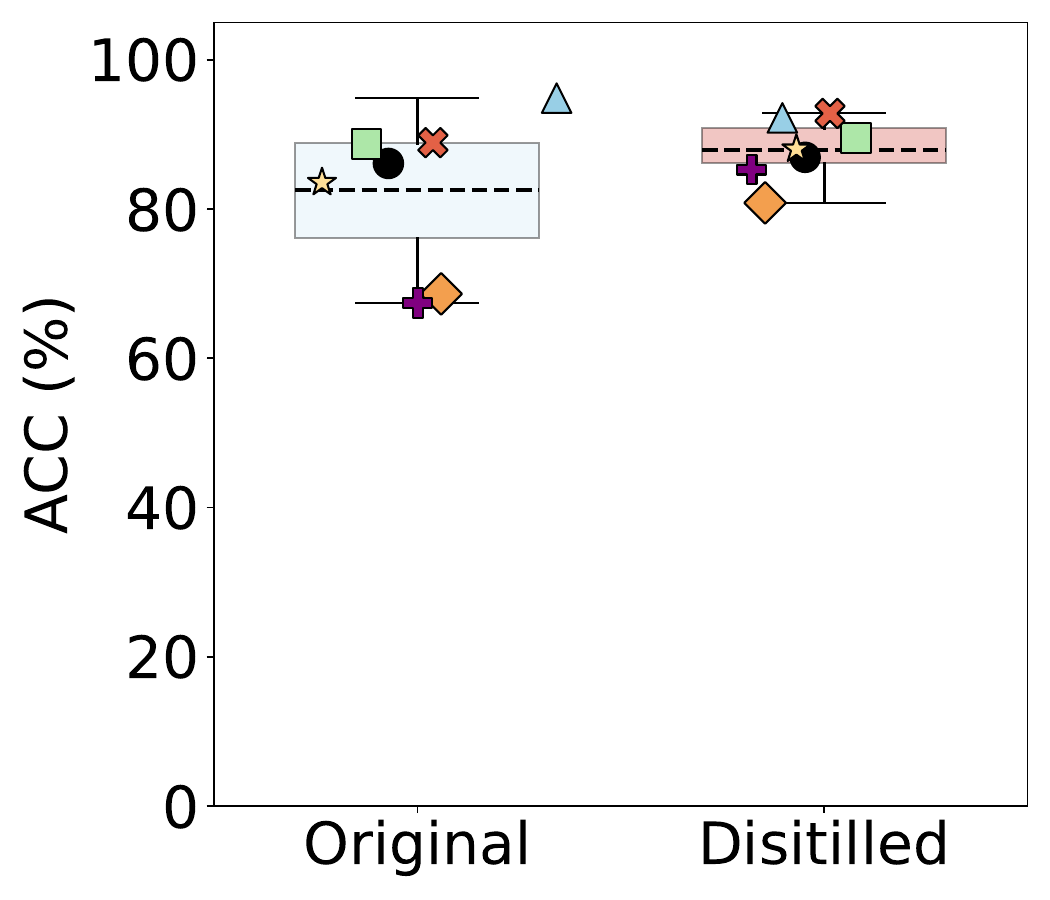}
    \caption{Caltech101}
\end{subfigure}
\begin{subfigure}[b]{0.16\textwidth}
    \centering
    \includegraphics[width=\textwidth]{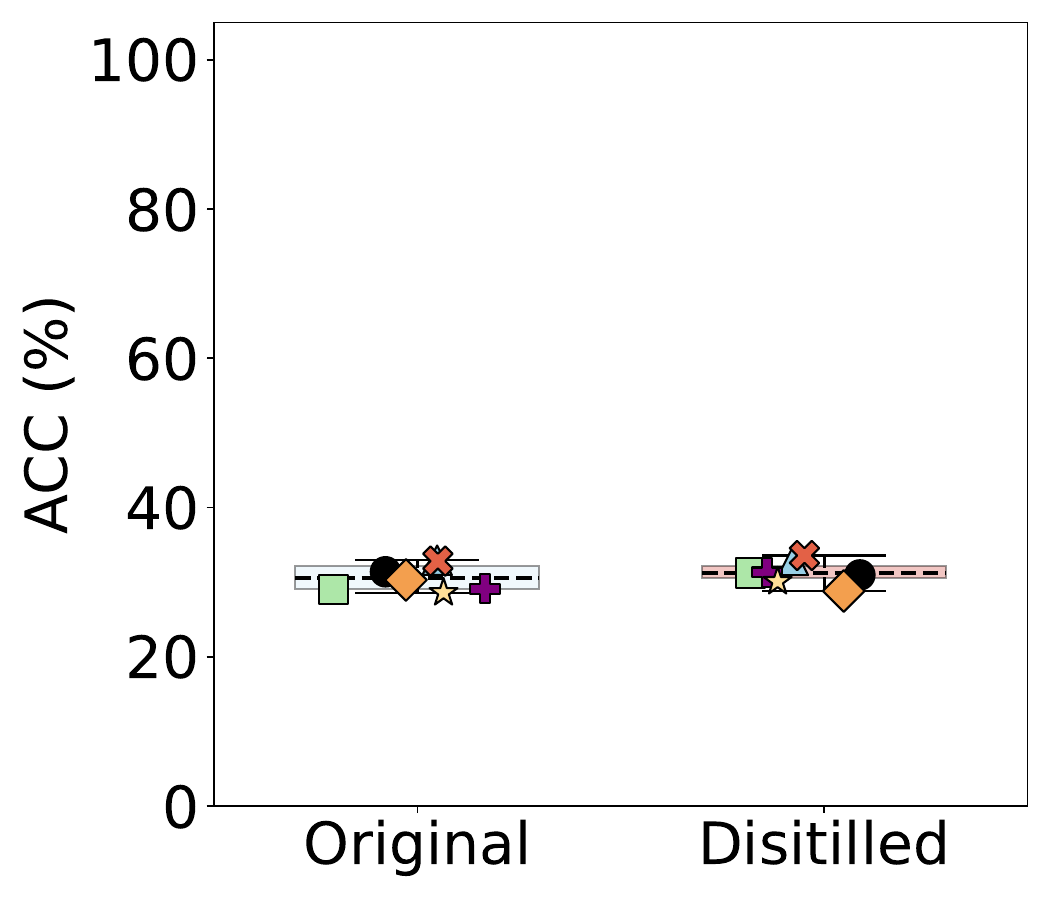}
    \caption{FER2013}
\end{subfigure}
\begin{subfigure}[b]{0.16\textwidth}
    \centering
    \includegraphics[width=\textwidth]{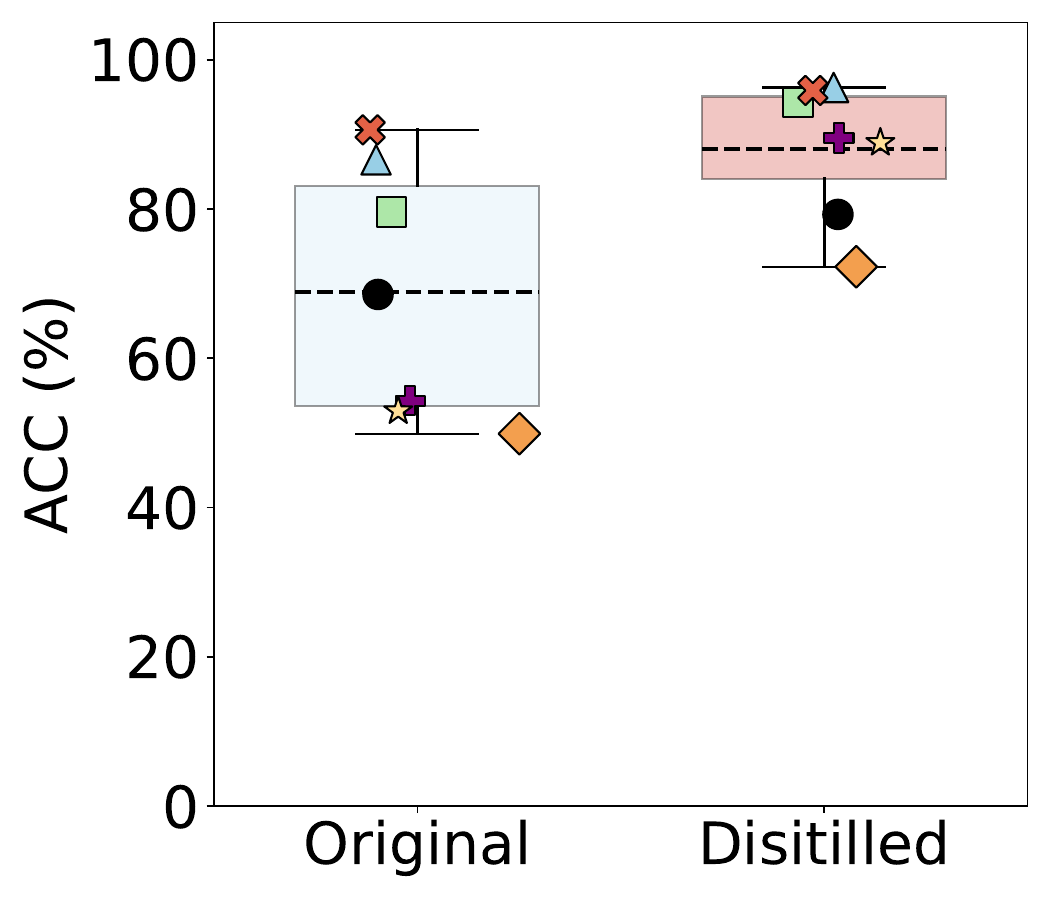}
    \caption{EuroSAT}
\end{subfigure}

\begin{subfigure}[b]{0.16\textwidth}
    \centering
    \includegraphics[width=\textwidth]{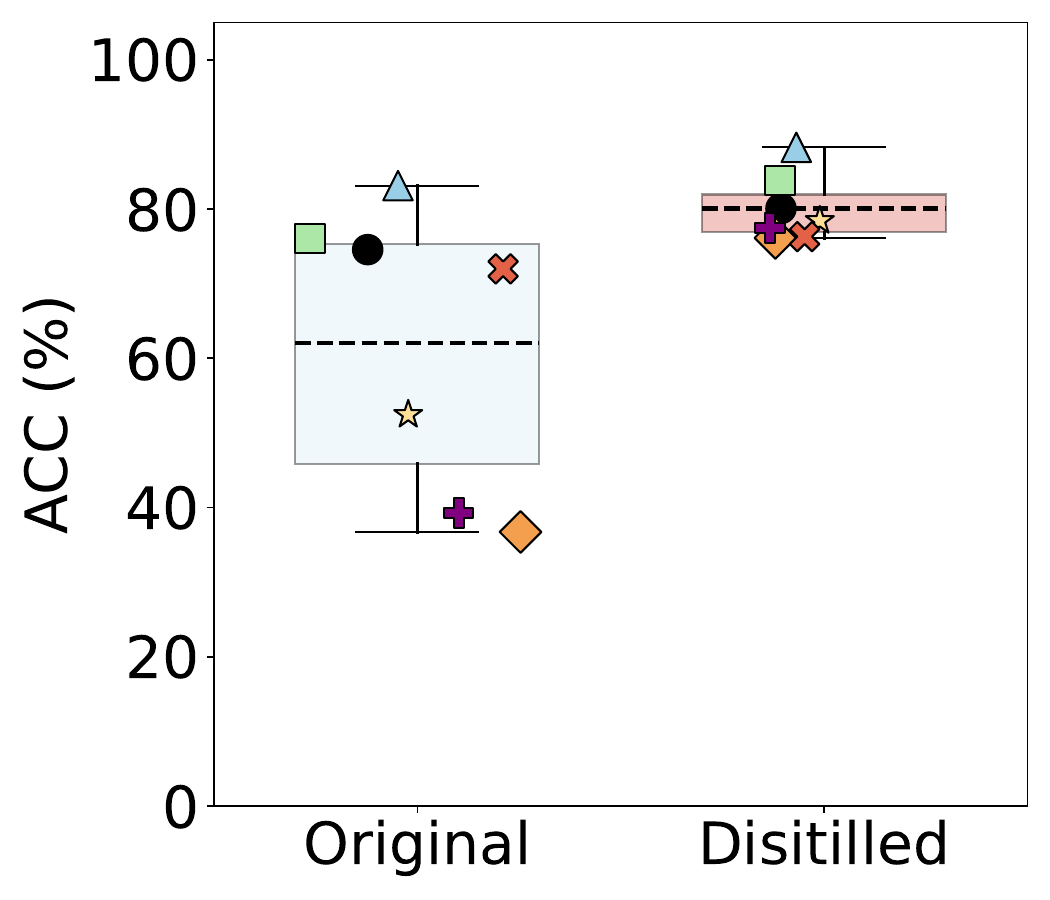}
    \caption{RESISC45}
\end{subfigure}
\begin{subfigure}[b]{0.16\textwidth}
    \centering
    \includegraphics[width=\textwidth]{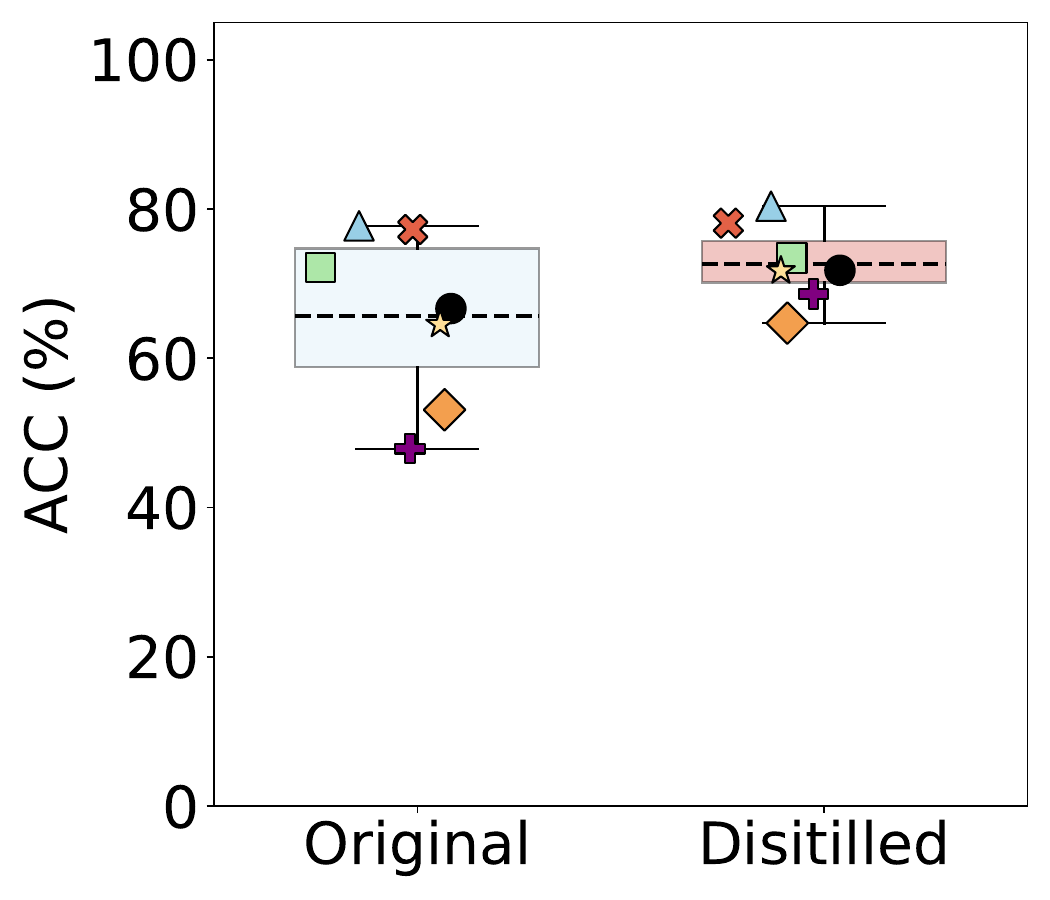}
    \caption{UCF101}
\end{subfigure}
\begin{subfigure}[b]{0.16\textwidth}
    \centering
    \includegraphics[width=\textwidth]{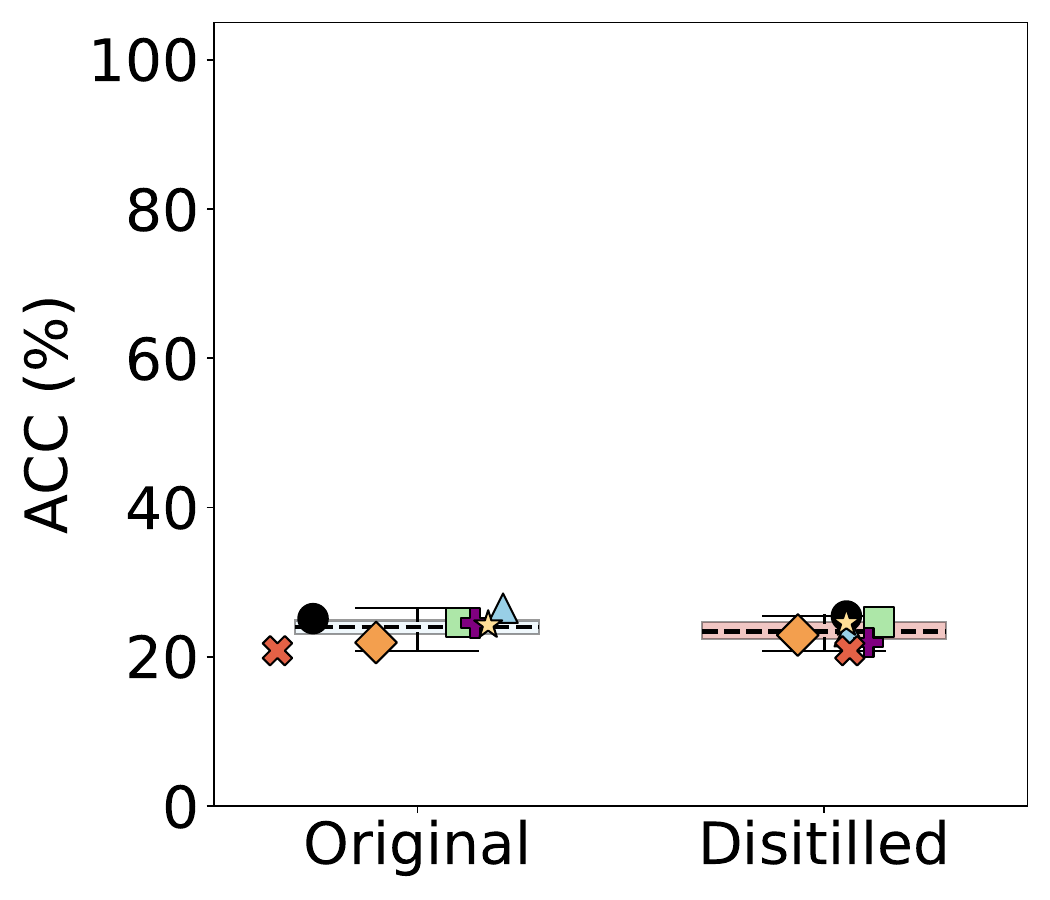}
    \caption{CLEVR}
\end{subfigure}
\begin{subfigure}[b]{0.16\textwidth}
    \centering
    \includegraphics[width=\textwidth]{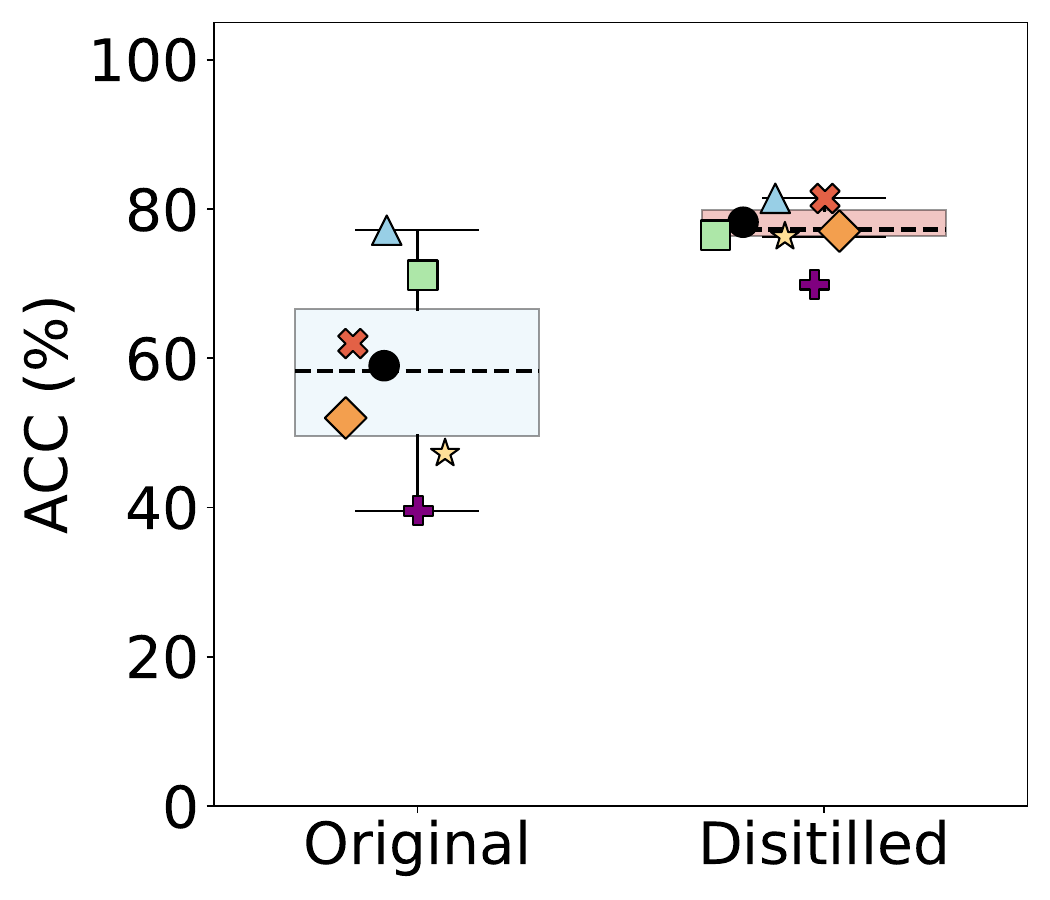}
    \caption{USPS}
\end{subfigure}
\begin{subfigure}[b]{0.16\textwidth}
    \centering
    \includegraphics[width=\textwidth]{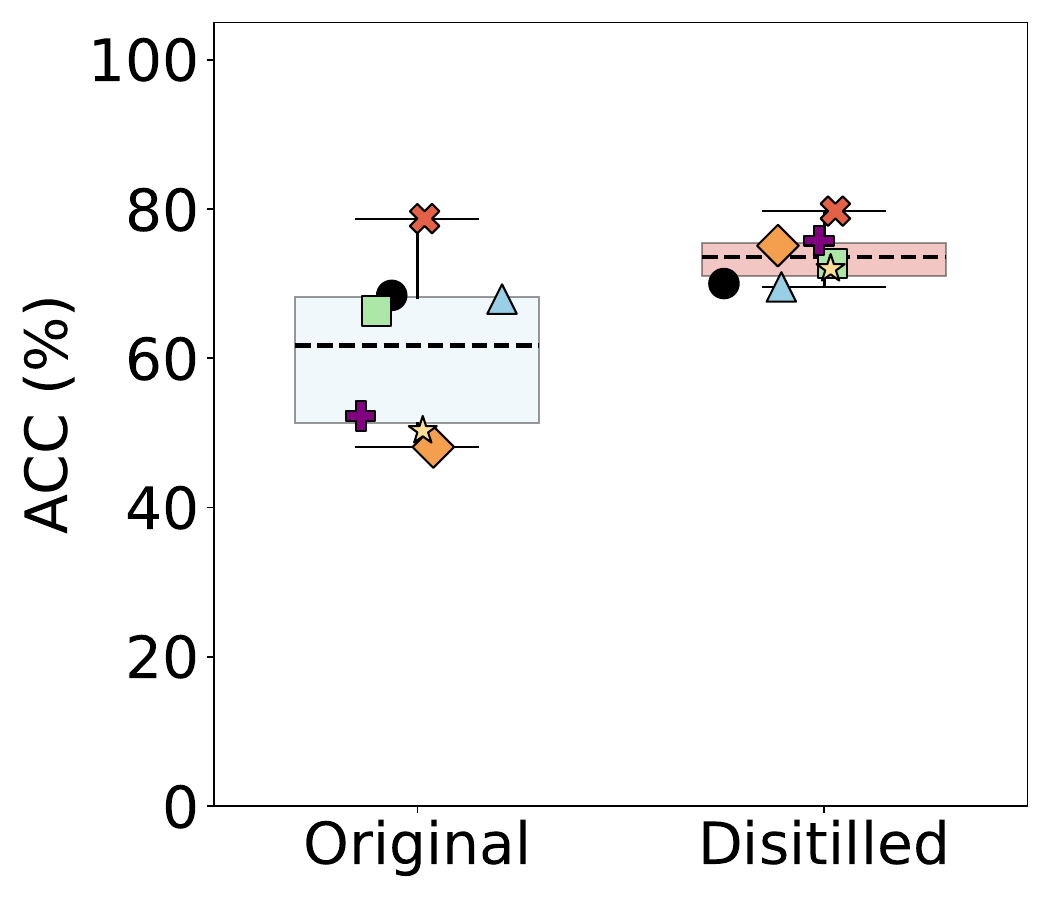}
    \caption{Fashion}
\end{subfigure}
\begin{subfigure}[b]{0.16\textwidth}
    \centering
    \includegraphics[width=\textwidth]{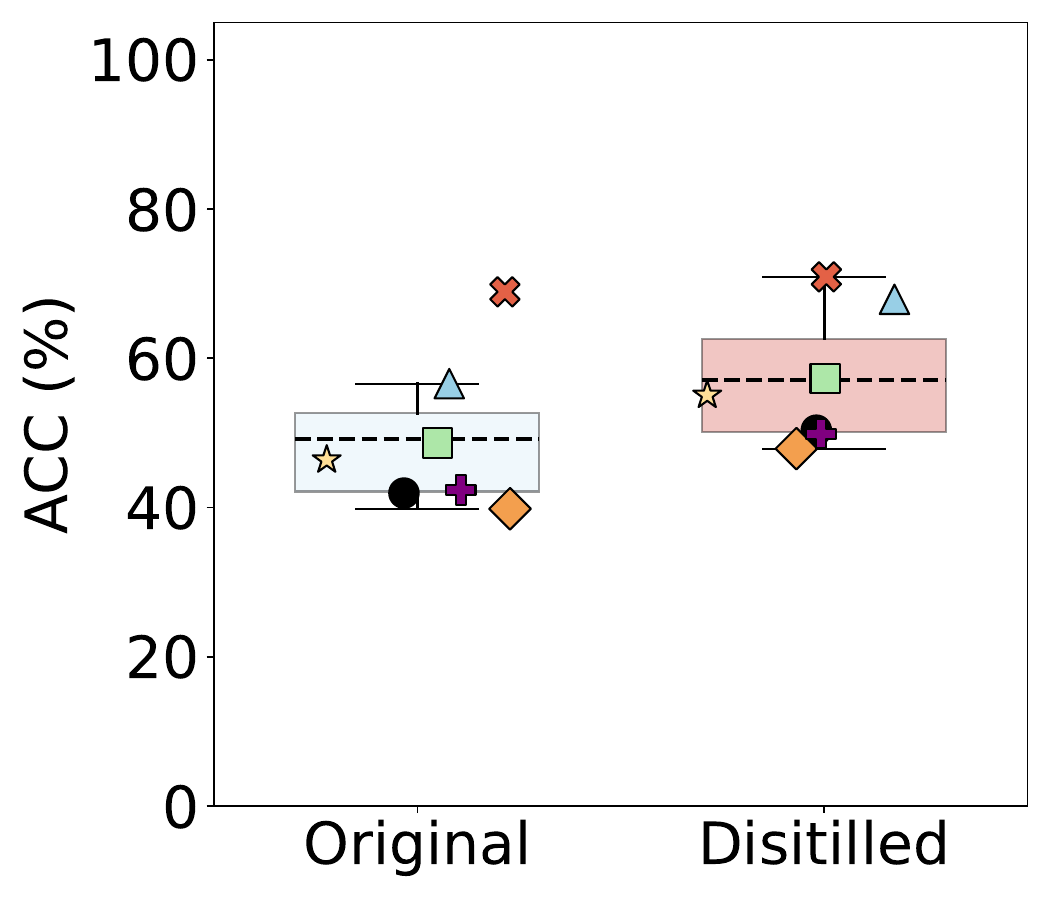}
    \caption{ImageNet-1K}
\end{subfigure}

\begin{subfigure}[b]{\textwidth}
  \centering
  \includegraphics[width=\textwidth]{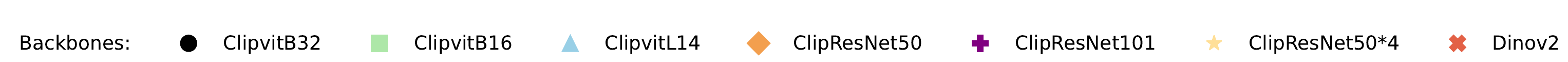} %
\end{subfigure}
\caption{Detailed clustering stability plots for all 18 datasets. Each subplot shows the distribution of clustering accuracy across 7 backbones for Original features and Distilled features on clustering assignment. }
\label{fig:appendix_stability}
\end{figure*}

Figure~\ref{fig:appendix_stability} displays the clustering accuracy distributions obtained across seven different foundation models for each of the 18 datasets, comparing the original embeddings against the FSRC-distilled embeddings.

Overall, the accuracy distributions of FSRC-distilled embeddings are consistently tighter than those of the original embeddings across most datasets, indicating superior robustness to the choice of the pre-trained backbone. For datasets with simpler class structures, such as \textit{CIFAR-10}, both feature types yield high and stable accuracy; however, FSRC exhibits a narrower variance, achieving more reliable performance across models.

On fine-grained datasets like \textit{OxfordPets} and \textit{ImageNet-Dogs}, the benefit of FSRC is more pronounced. While results based on original embeddings show considerable scatter across different backbones, FSRC maintains a higher median accuracy with a more concentrated distribution, demonstrating effective compensation for the weaknesses of individual foundation models.

Remote sensing datasets, including \textit{EuroSAT} and \textit{RESISC45}, also show clear improvement. FSRC yields higher median accuracy and a more compact distribution than the original features, confirming its ability to integrate complementary cues for consistent performance.

On challenging large-scale datasets such as \textit{UCF101} and \textit{ImageNet}, FSRC not only improves median clustering accuracy but also markedly reduces inter-backbone variance. This consistent pattern across diverse data domains underscores that the feature representations learned by FSRC are less sensitive to the specifics of the underlying foundation model, highlighting the generalizability and robustness of our fusion approach.

\section{Clustering Accuracy on 18 Datasets Across 7 Backbones}
\label{H}

\begin{figure*}[t]
\centering
\begin{subfigure}[b]{0.16\textwidth}
  \includegraphics[width=\textwidth]{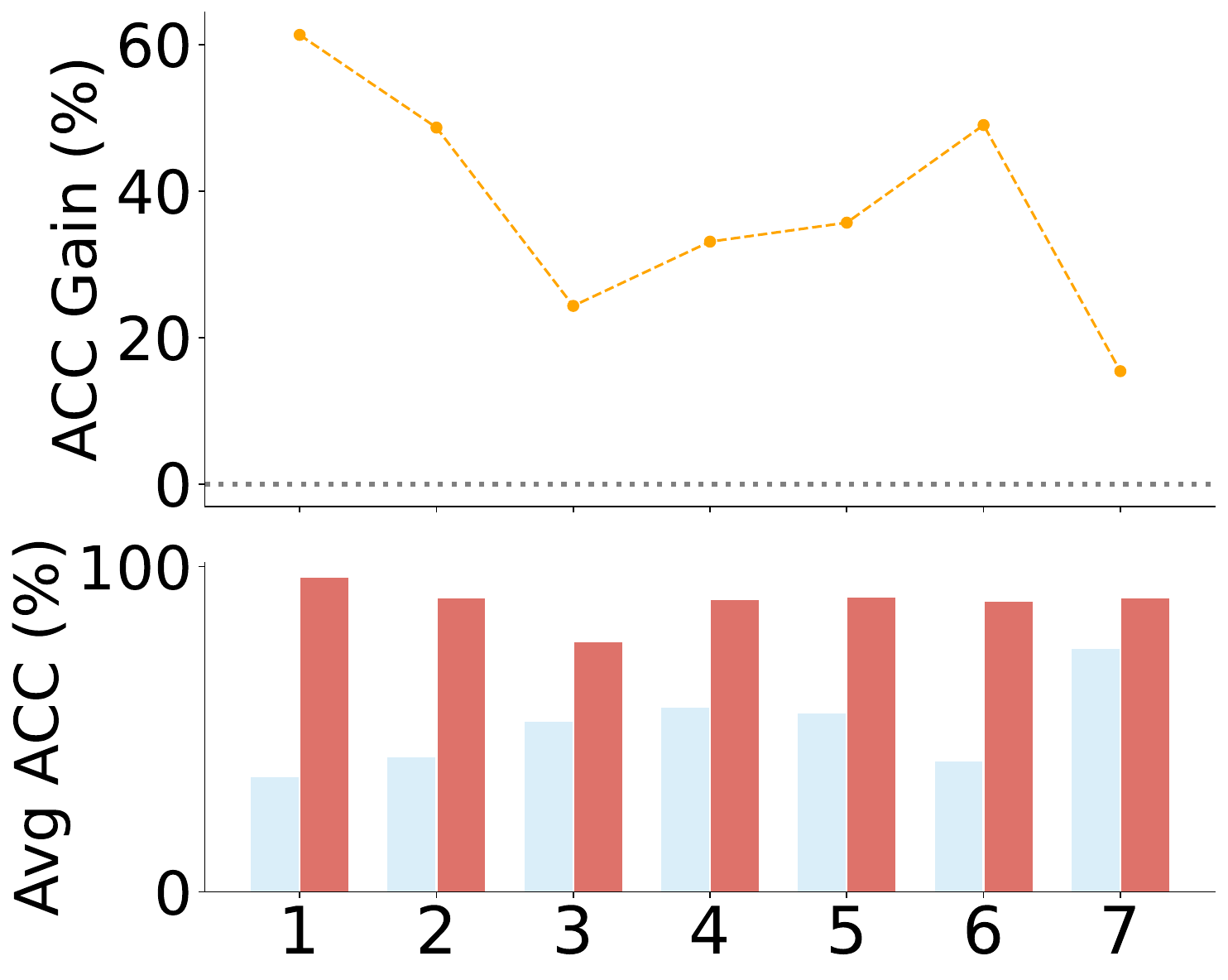}
  \caption{MNIST}
\end{subfigure}\hfill
\begin{subfigure}[b]{0.16\textwidth}
  \includegraphics[width=\textwidth]{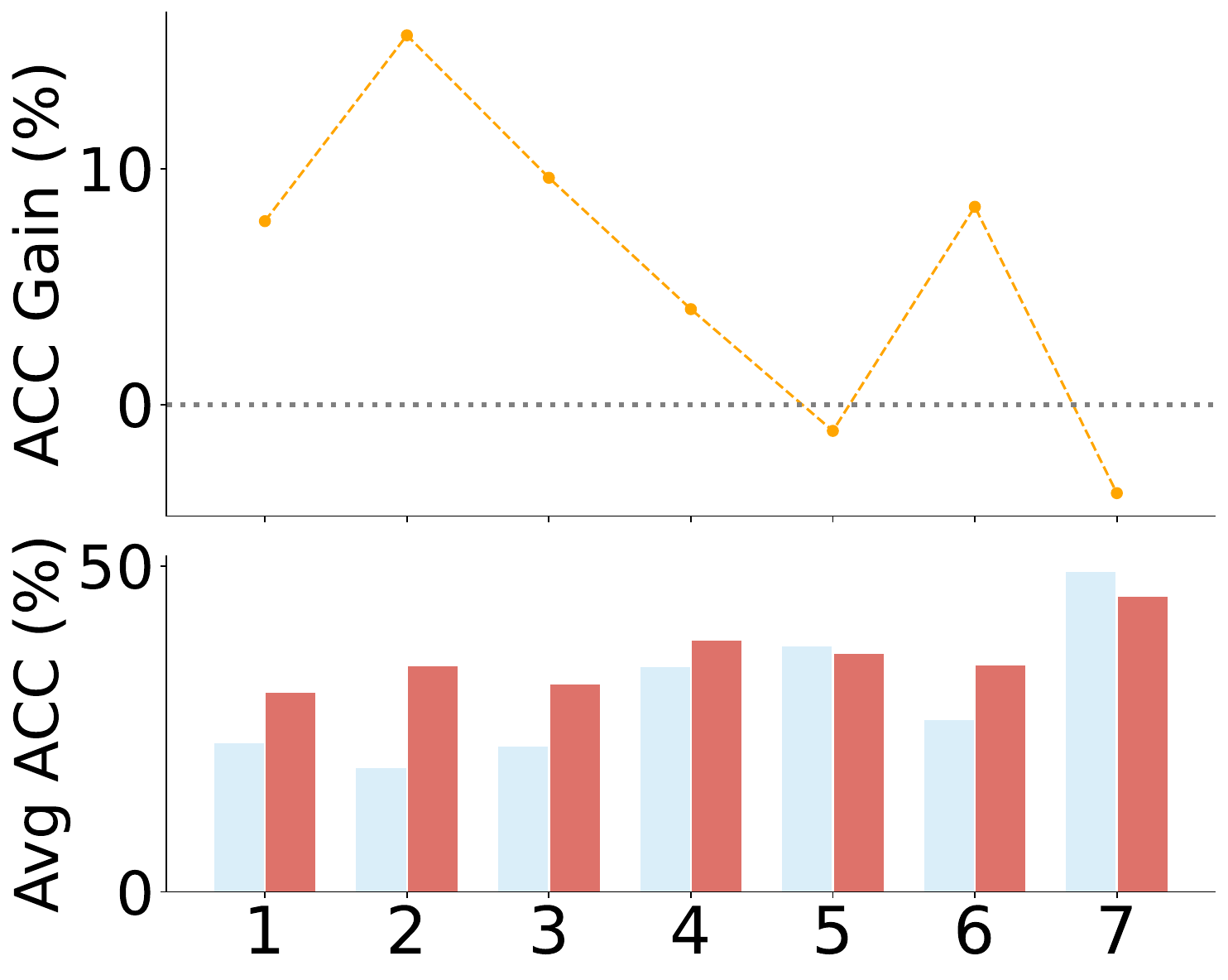}
  \caption{GTSRB}
\end{subfigure}\hfill
\begin{subfigure}[b]{0.16\textwidth}
  \includegraphics[width=\textwidth]{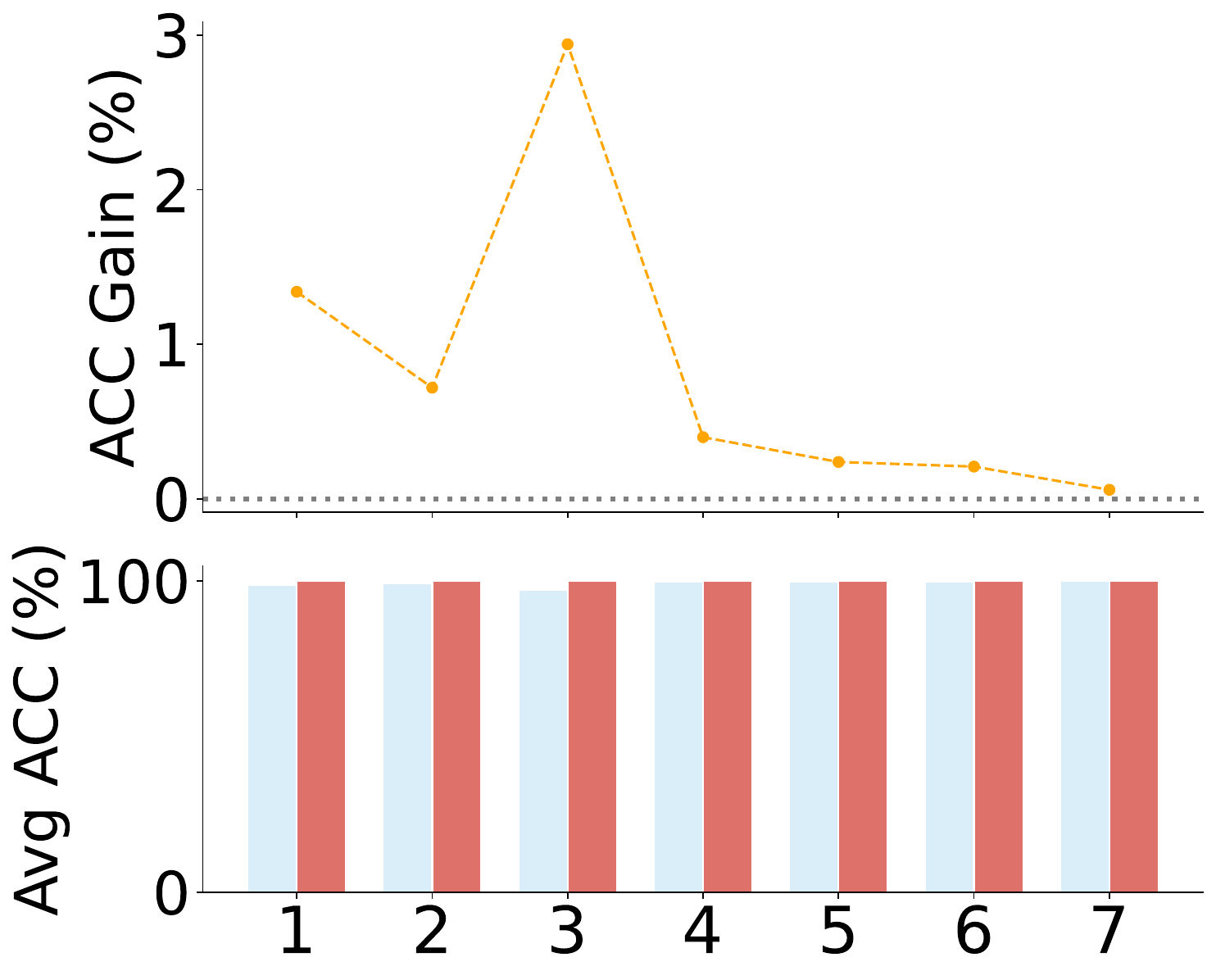}
  \caption{ImageNet-10}
\end{subfigure}
\begin{subfigure}[b]{0.16\textwidth}
  \includegraphics[width=\textwidth]{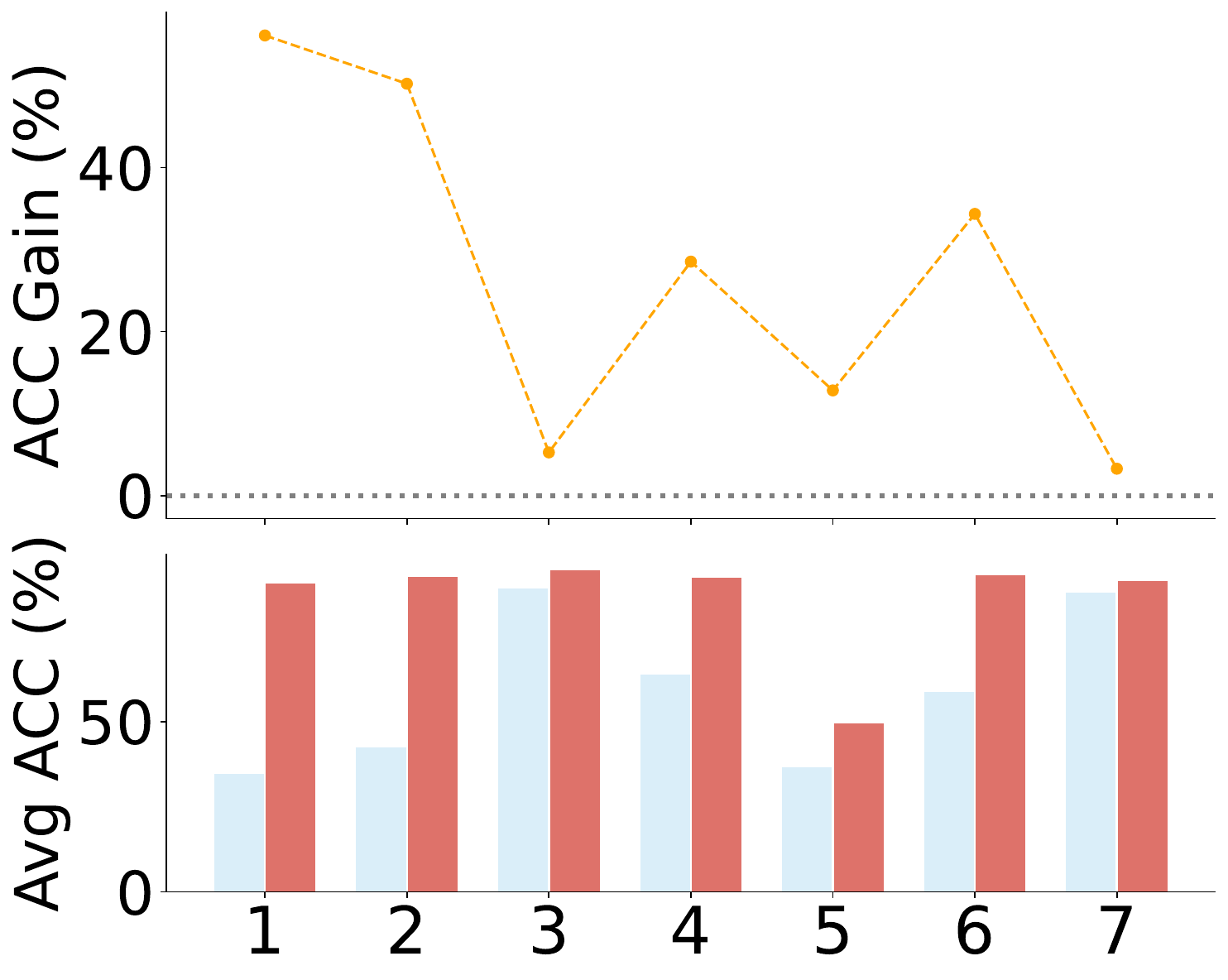}
  \caption{OxfordPets}
\end{subfigure}\hfill
\begin{subfigure}[b]{0.16\textwidth}
  \includegraphics[width=\textwidth]{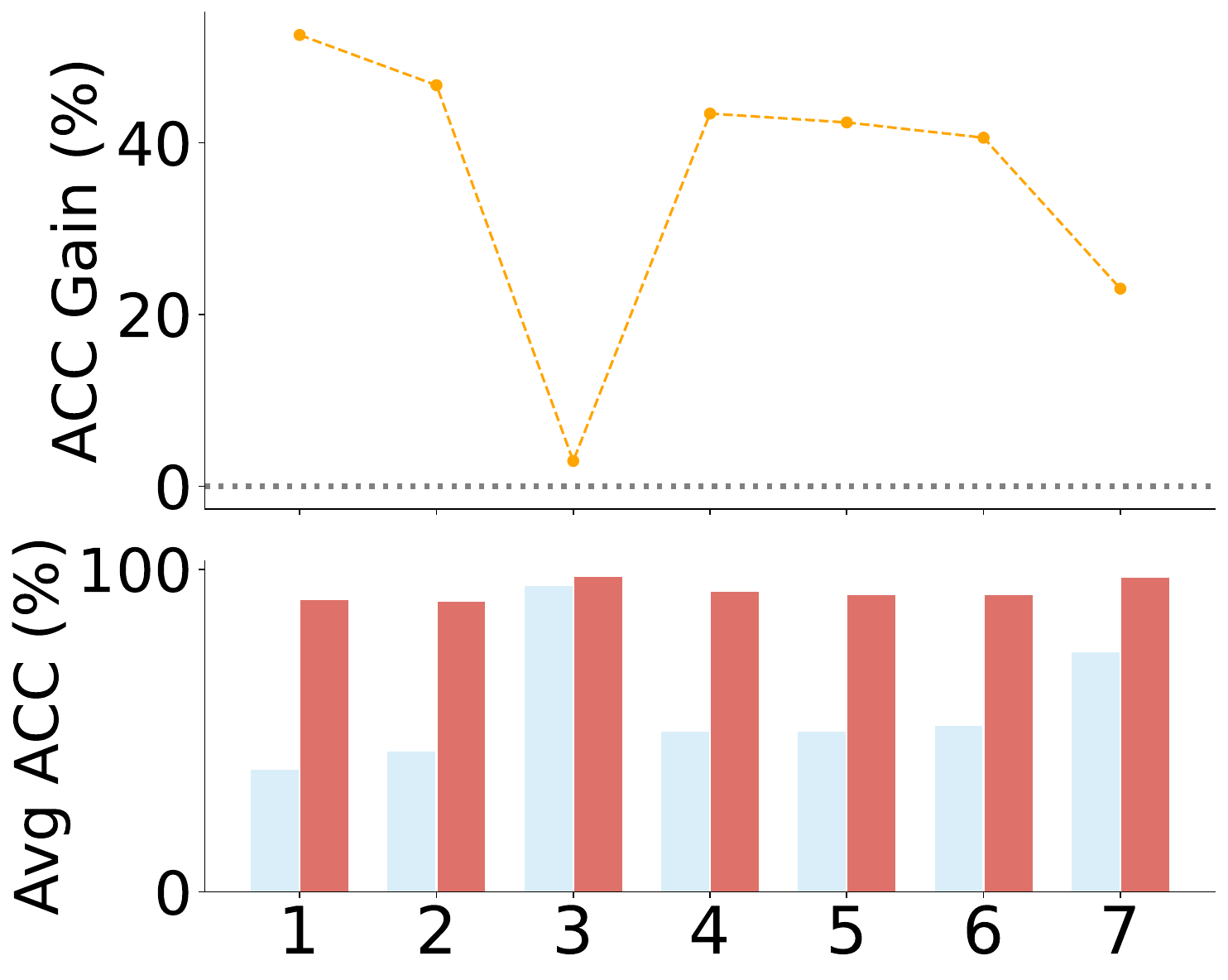}
  \caption{ImageNet-Dogs}
\end{subfigure}\hfill
\begin{subfigure}[b]{0.16\textwidth}
  \includegraphics[width=\textwidth]{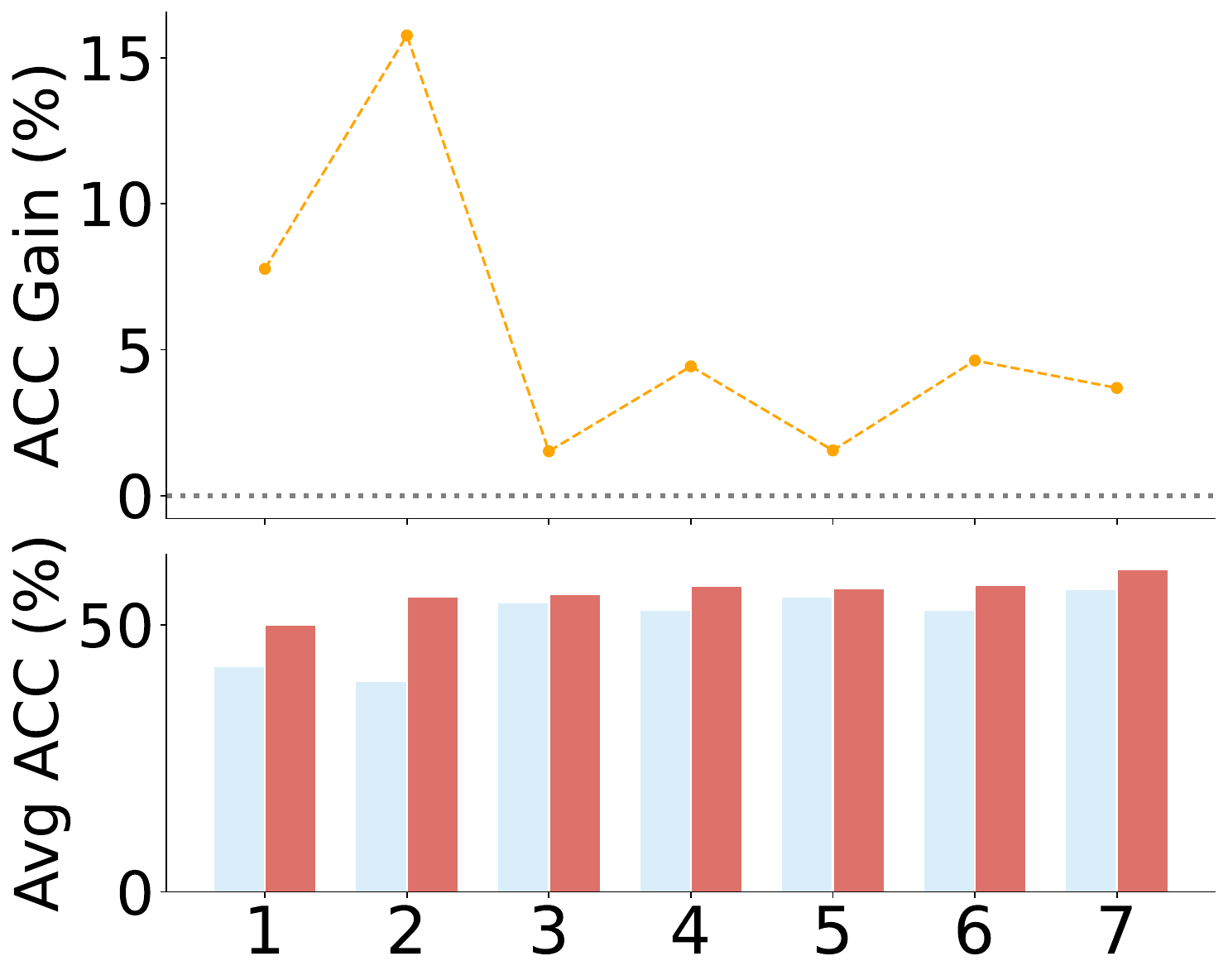}
  \caption{DTD}
\end{subfigure}

\begin{subfigure}[b]{0.16\textwidth}
  \includegraphics[width=\textwidth]{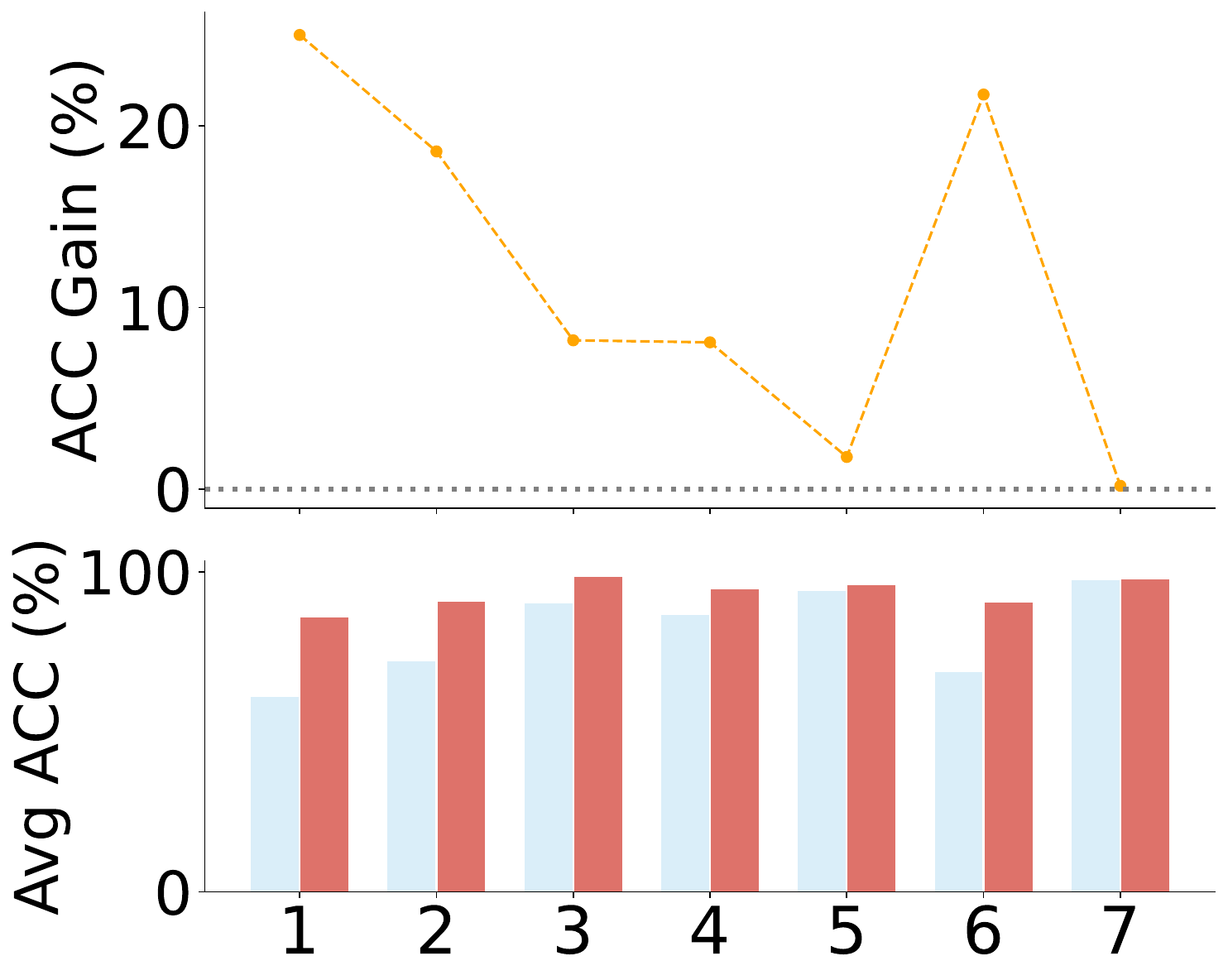}
  \caption{CIFAR-10}
\end{subfigure}\hfill
\begin{subfigure}[b]{0.16\textwidth}
  \includegraphics[width=\textwidth]{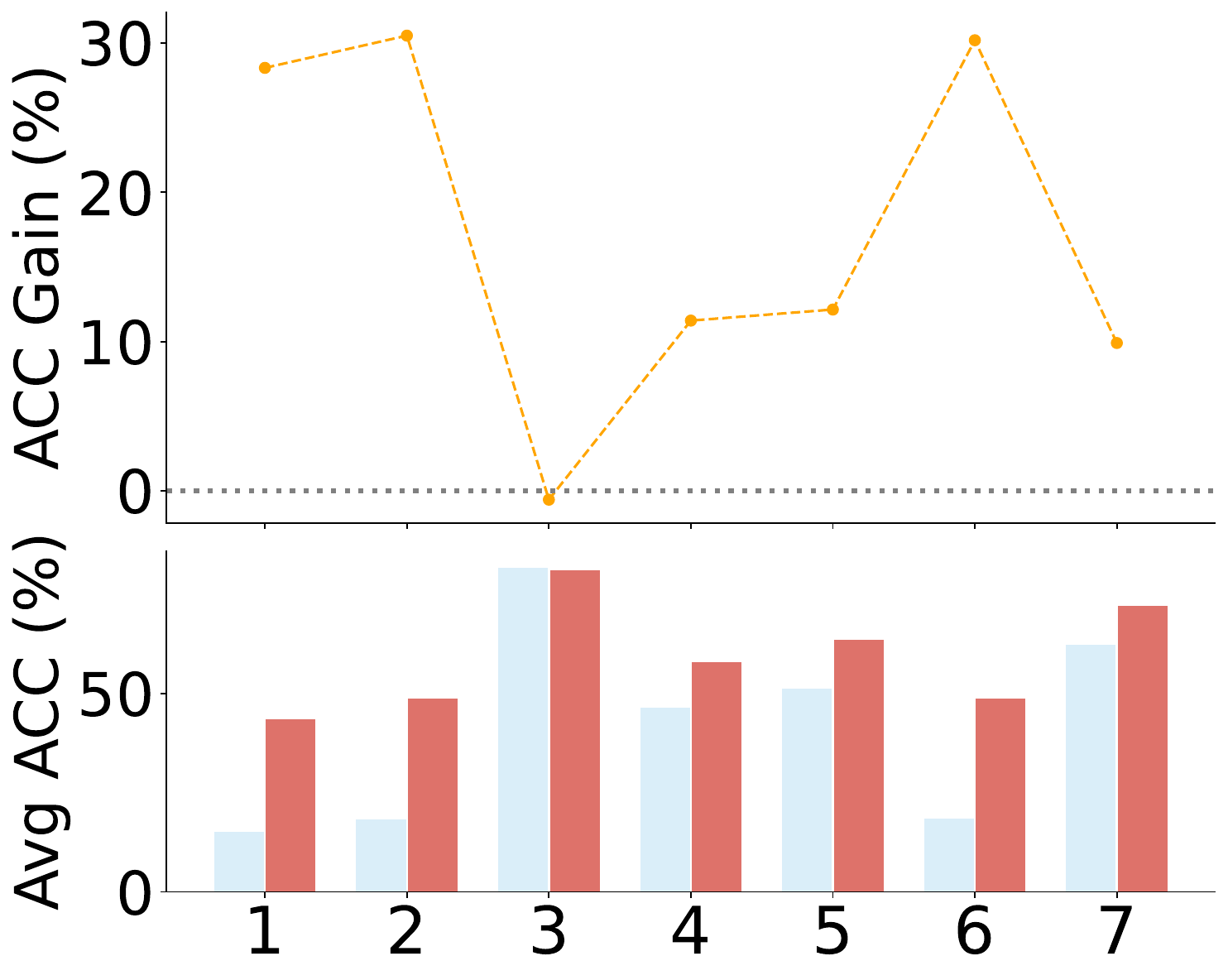}
  \caption{CIFAR-100}
\end{subfigure}\hfill
\begin{subfigure}[b]{0.16\textwidth}
  \includegraphics[width=\textwidth]{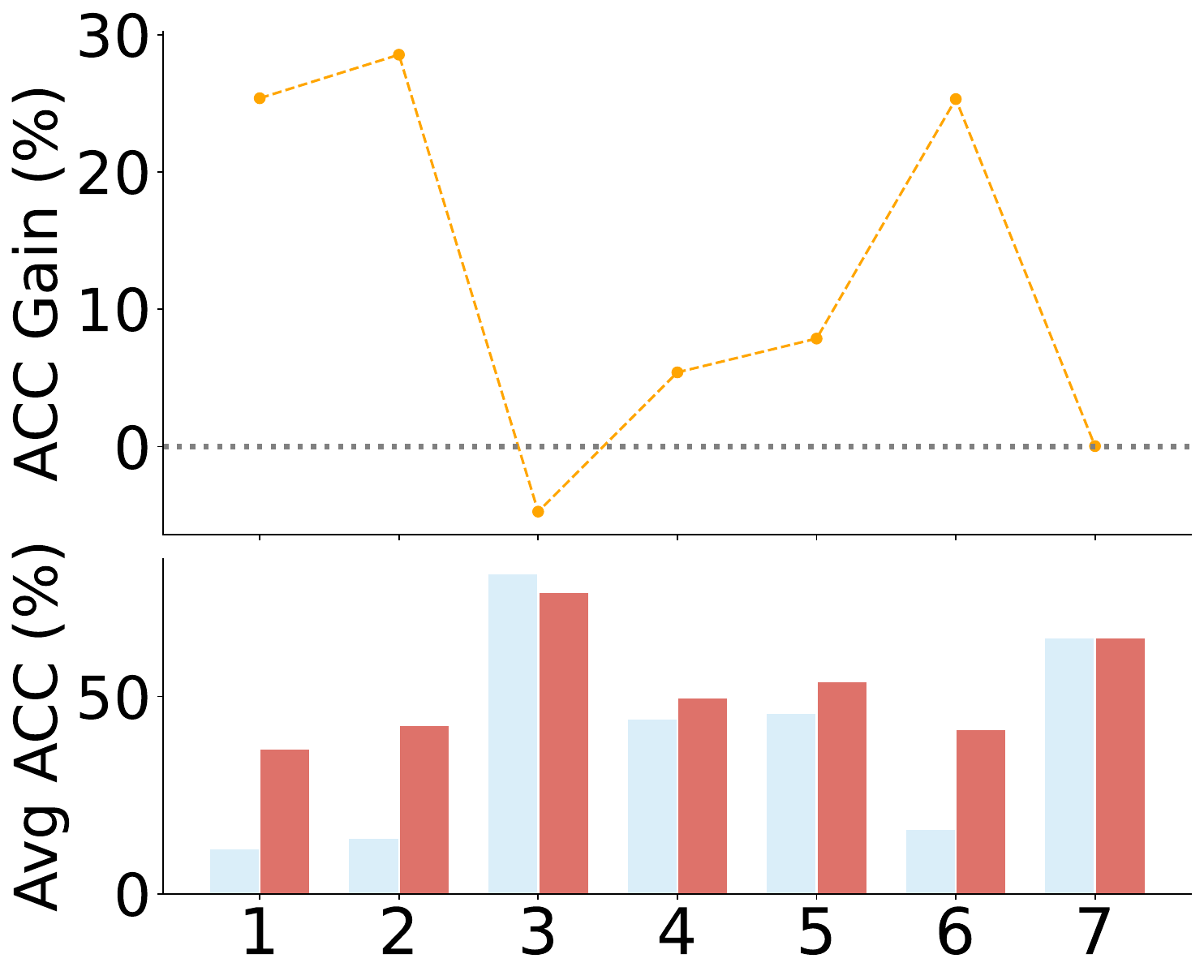}
  \caption{TinyImageNet}
\end{subfigure}
\begin{subfigure}[b]{0.16\textwidth}
  \includegraphics[width=\textwidth]{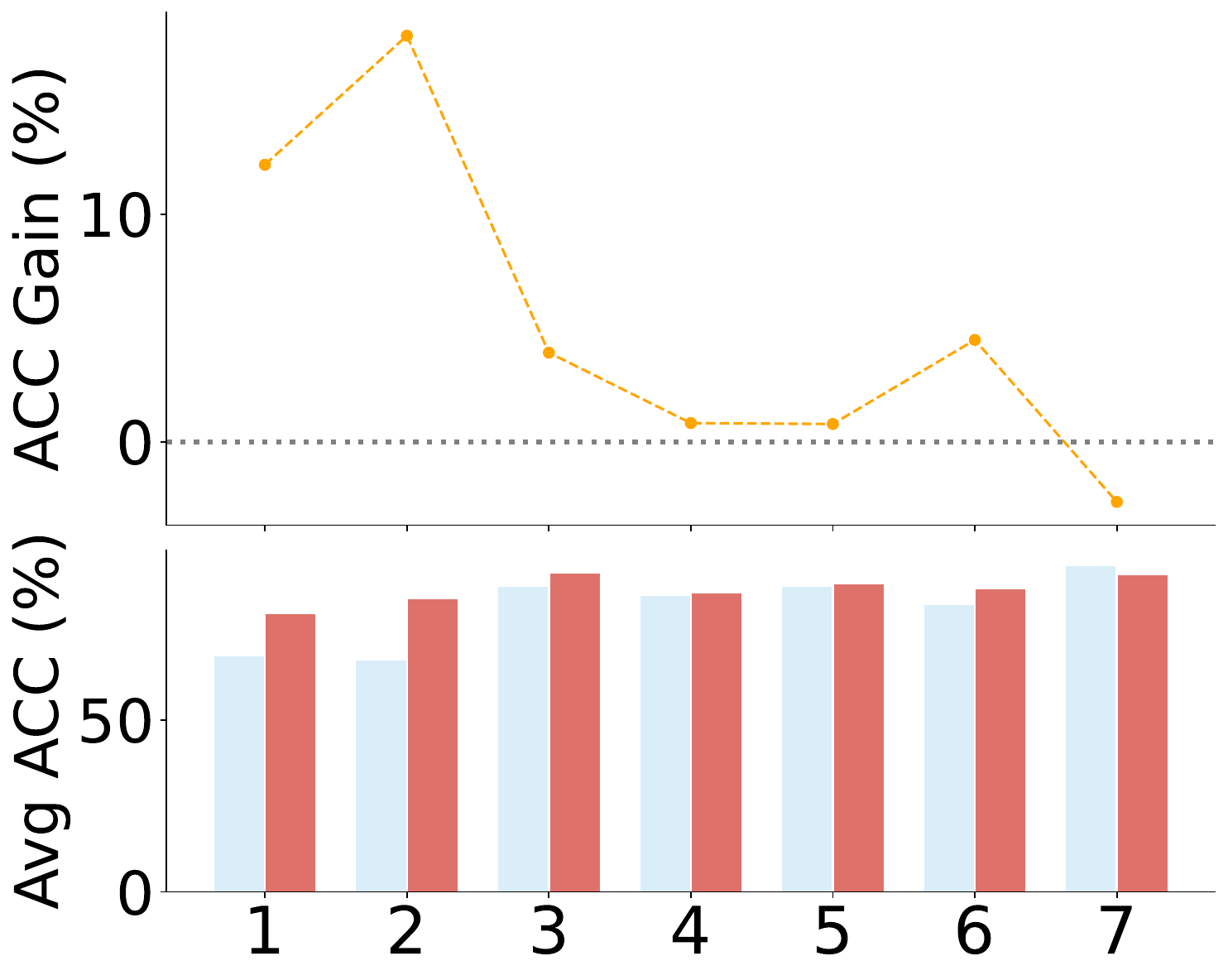}
  \caption{Caltech101}
\end{subfigure}\hfill
\begin{subfigure}[b]{0.16\textwidth}
  \includegraphics[width=\textwidth]{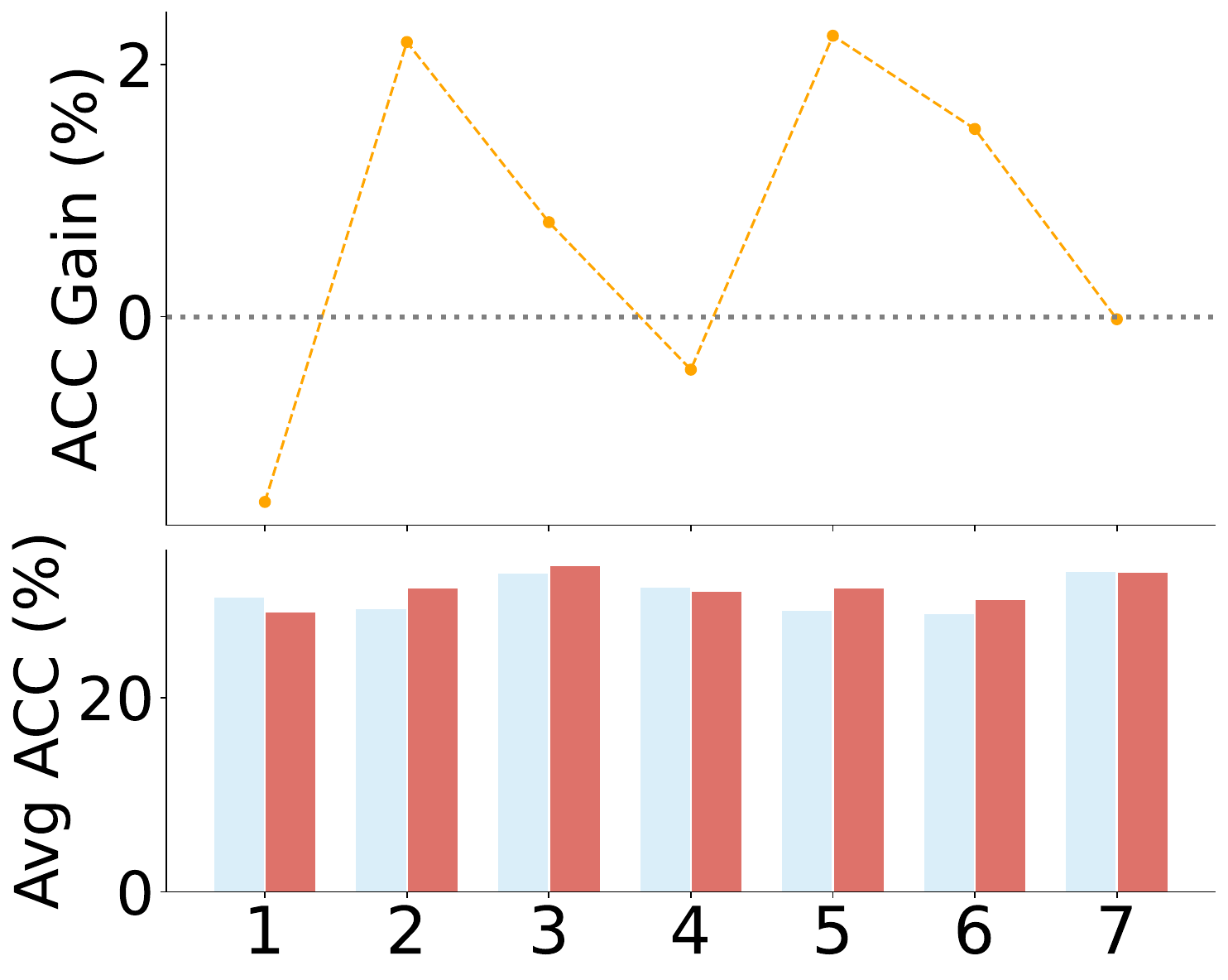}
  \caption{FER2013}
\end{subfigure}\hfill
\begin{subfigure}[b]{0.16\textwidth}
  \includegraphics[width=\textwidth]{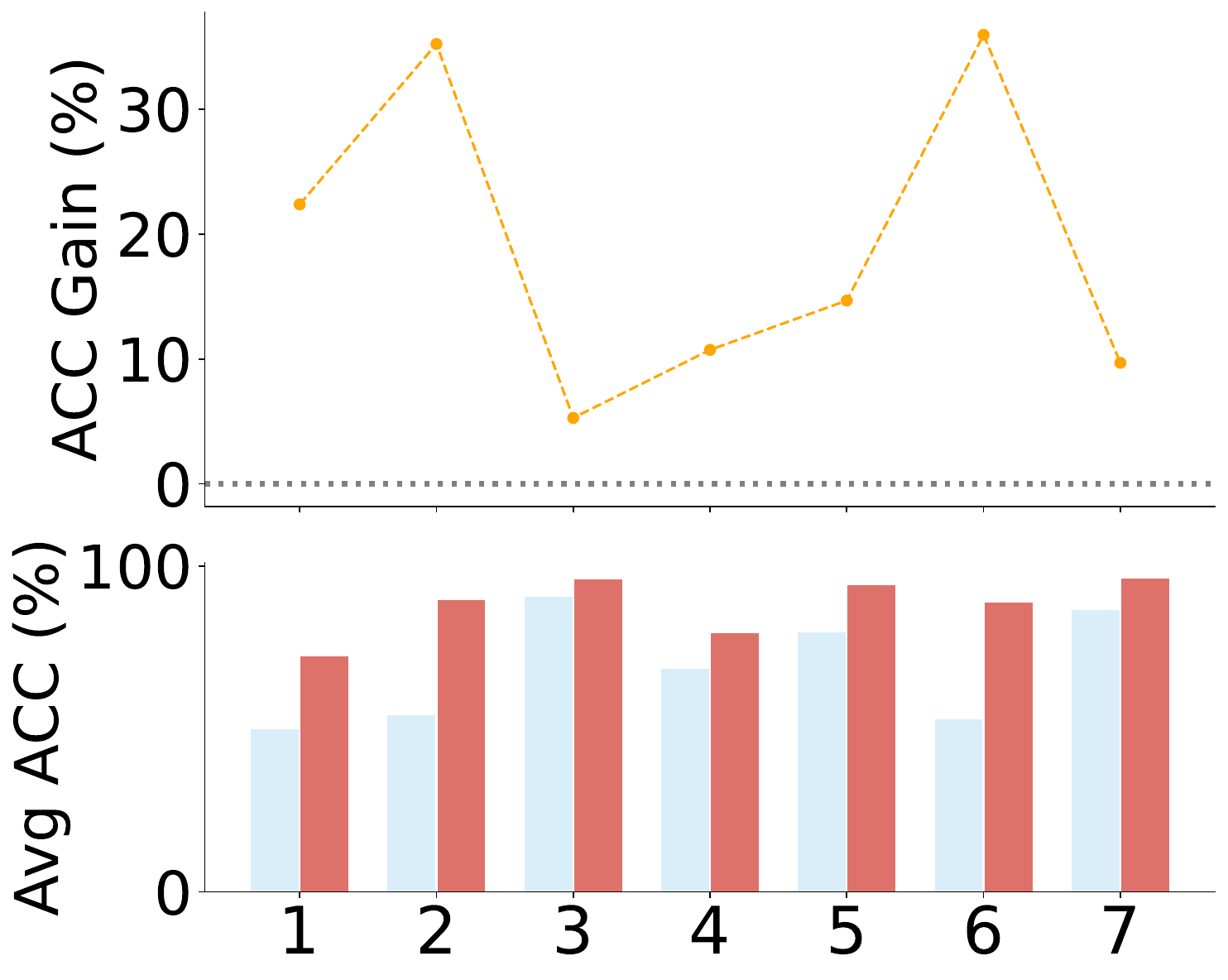}
  \caption{EuroSAT}
\end{subfigure}

\begin{subfigure}[b]{0.16\textwidth}
  \includegraphics[width=\textwidth]{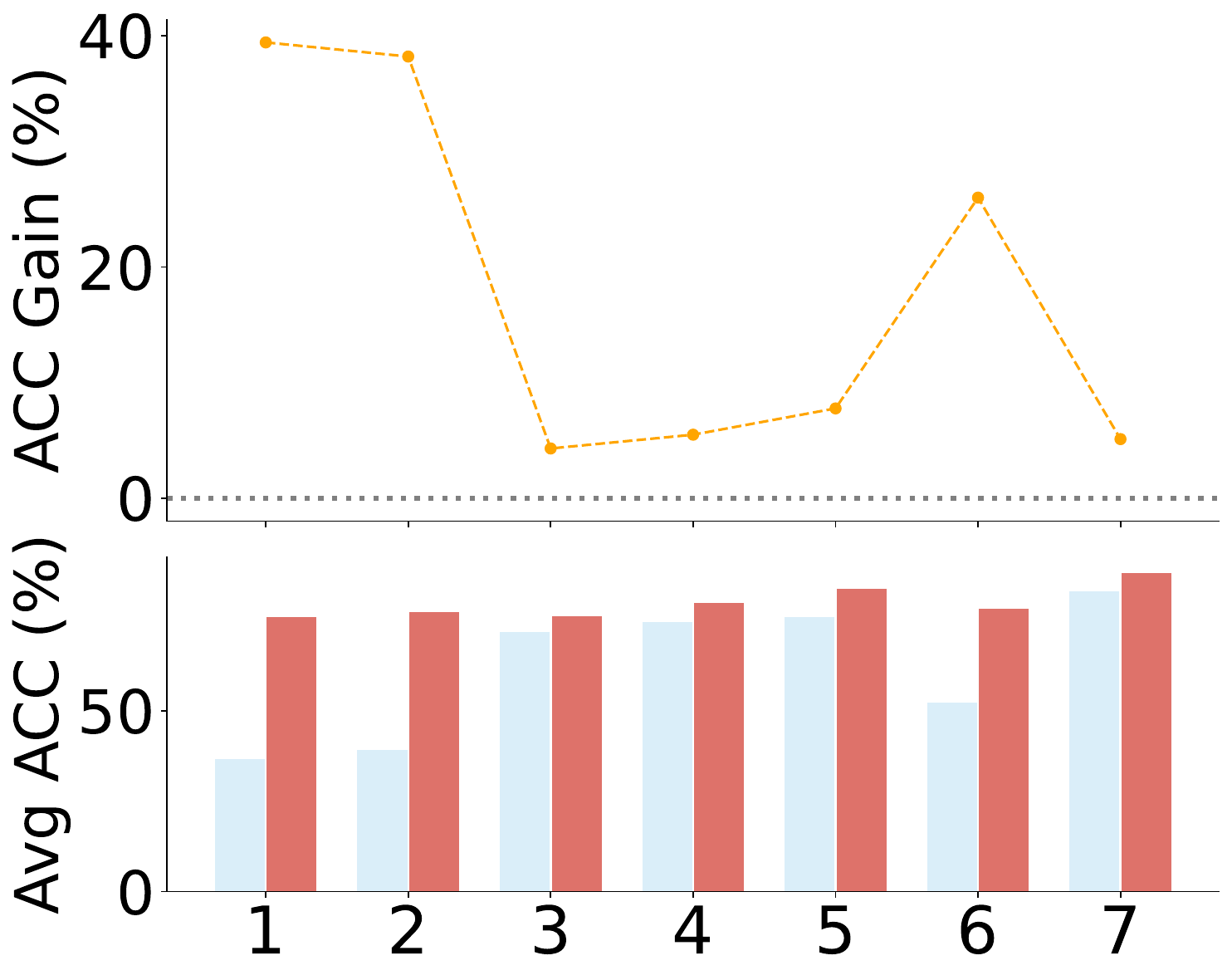}
  \caption{RESISC45}
\end{subfigure}\hfill
\begin{subfigure}[b]{0.16\textwidth}
  \includegraphics[width=\textwidth]{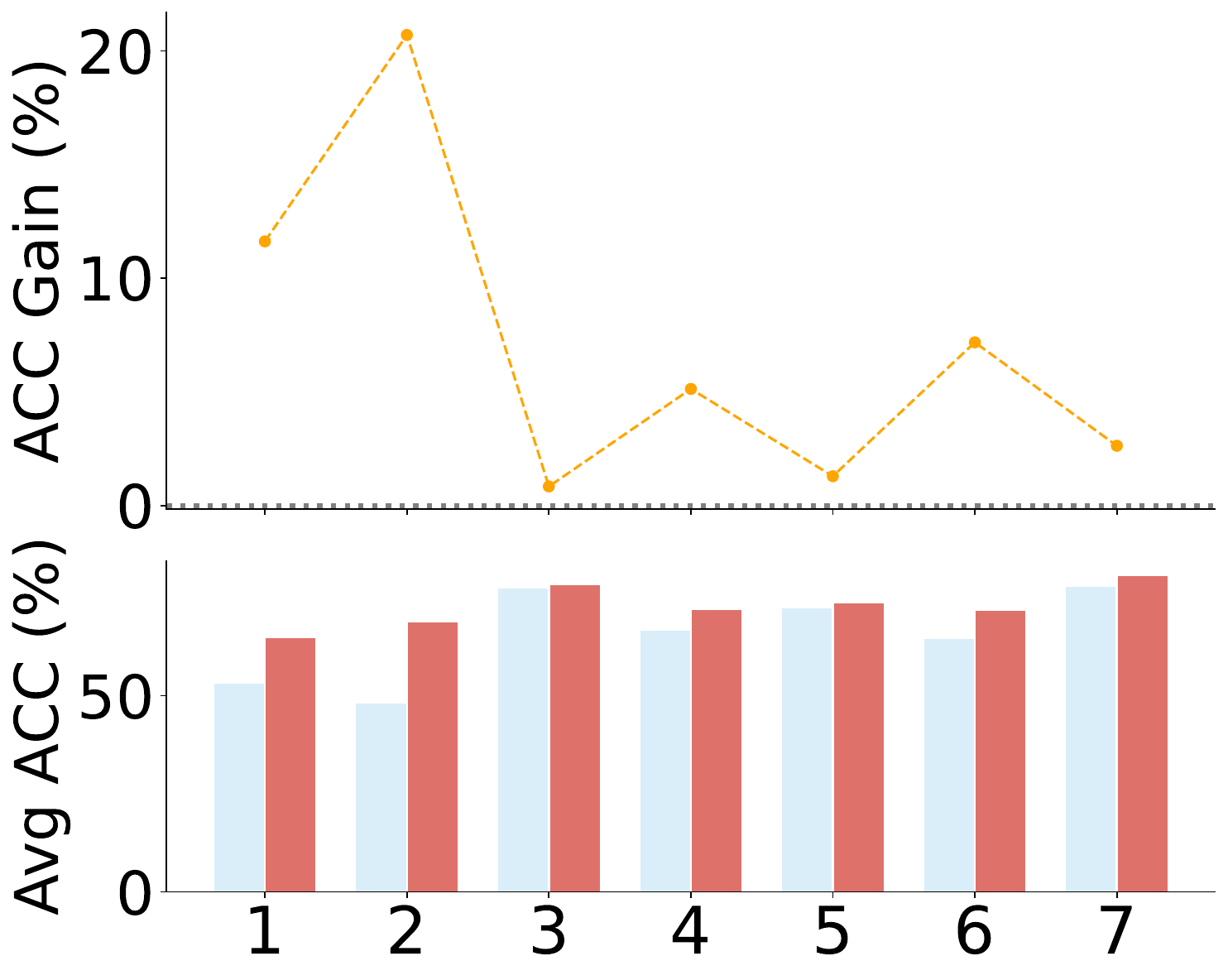}
  \caption{UCF101}
\end{subfigure}\hfill
\begin{subfigure}[b]{0.16\textwidth}
  \includegraphics[width=\textwidth]{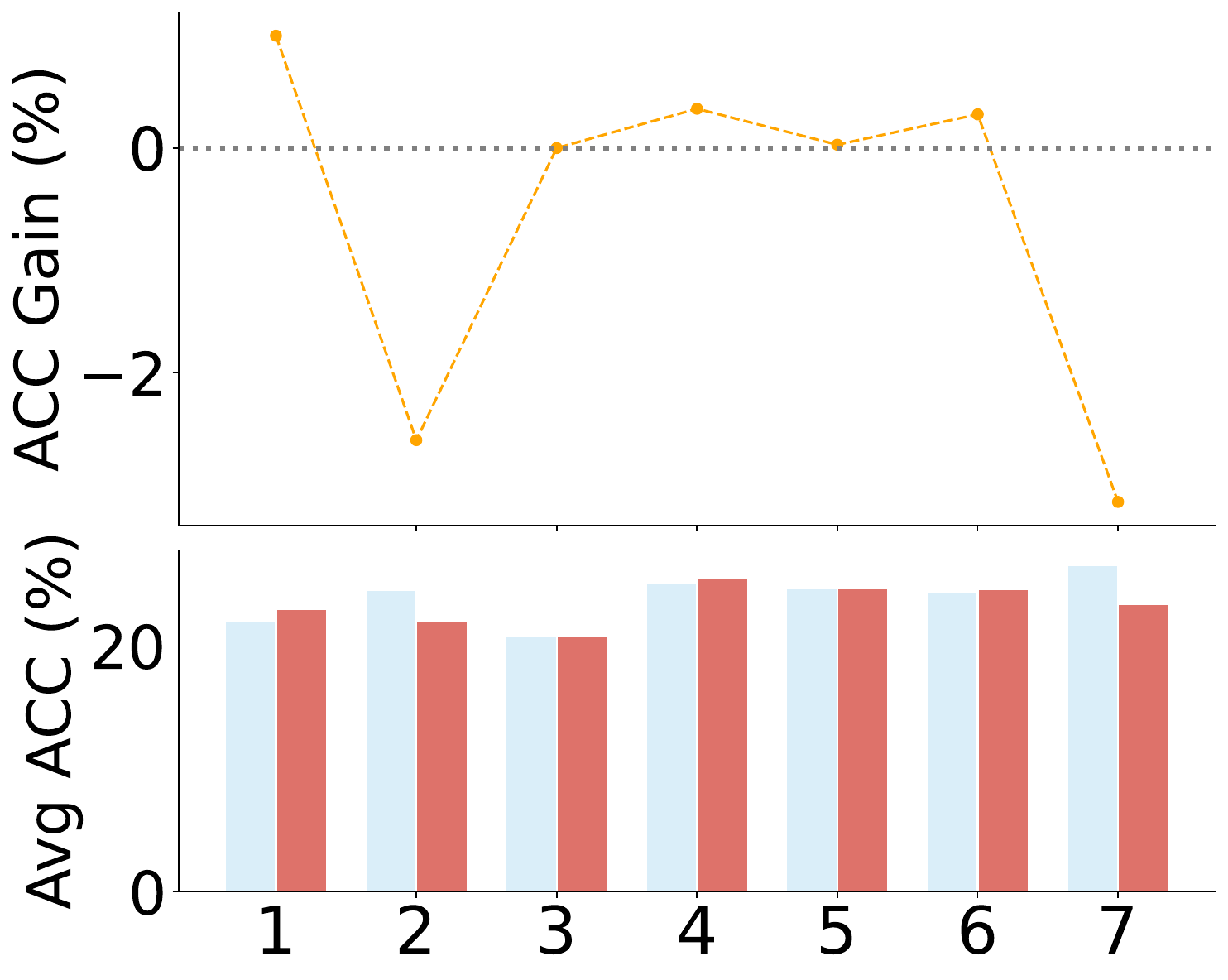}
  \caption{CLEVR}
\end{subfigure}
\begin{subfigure}[b]{0.16\textwidth}
  \includegraphics[width=\textwidth]{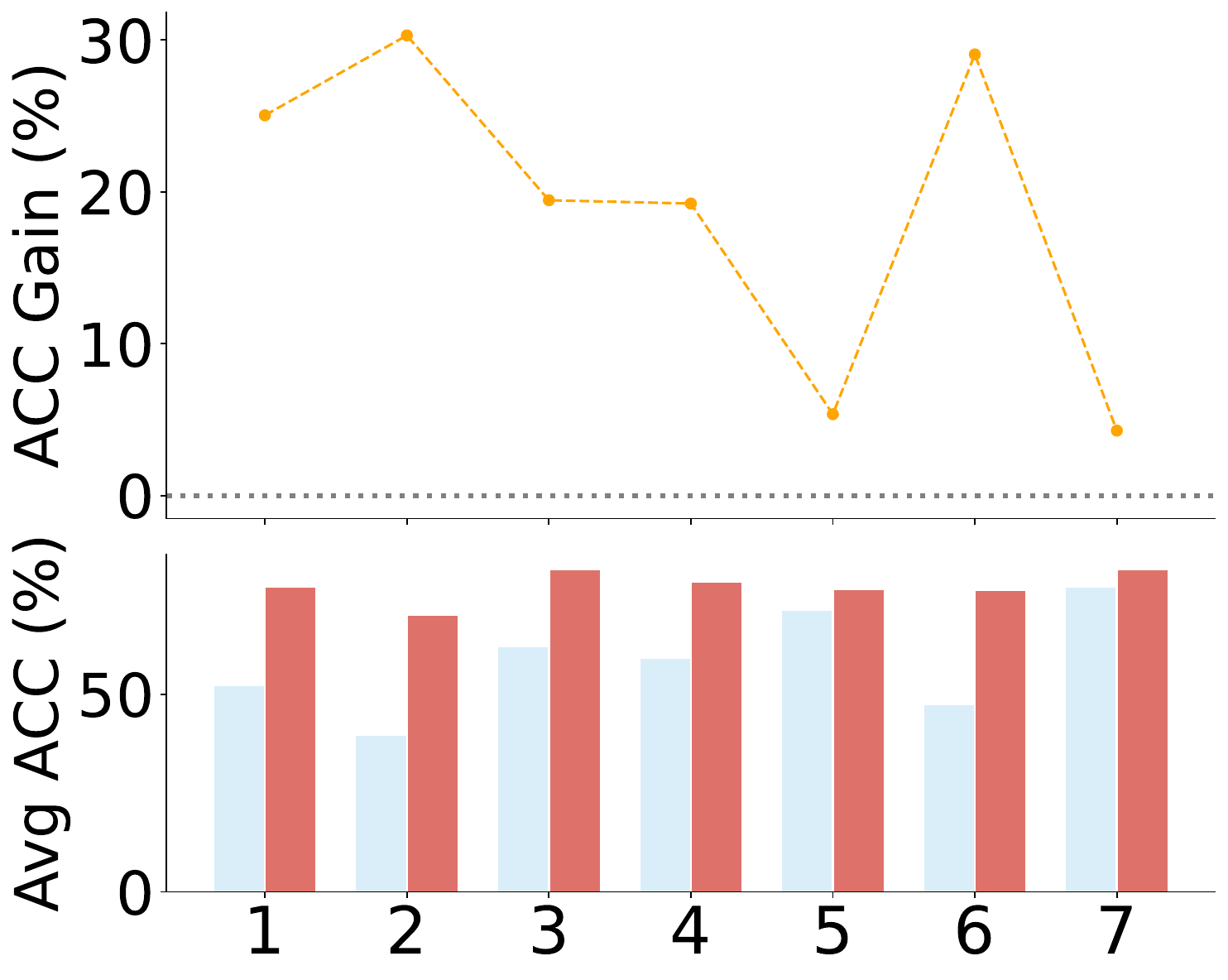}
  \caption{USPS}
\end{subfigure}\hfill
\begin{subfigure}[b]{0.16\textwidth}
  \includegraphics[width=\textwidth]{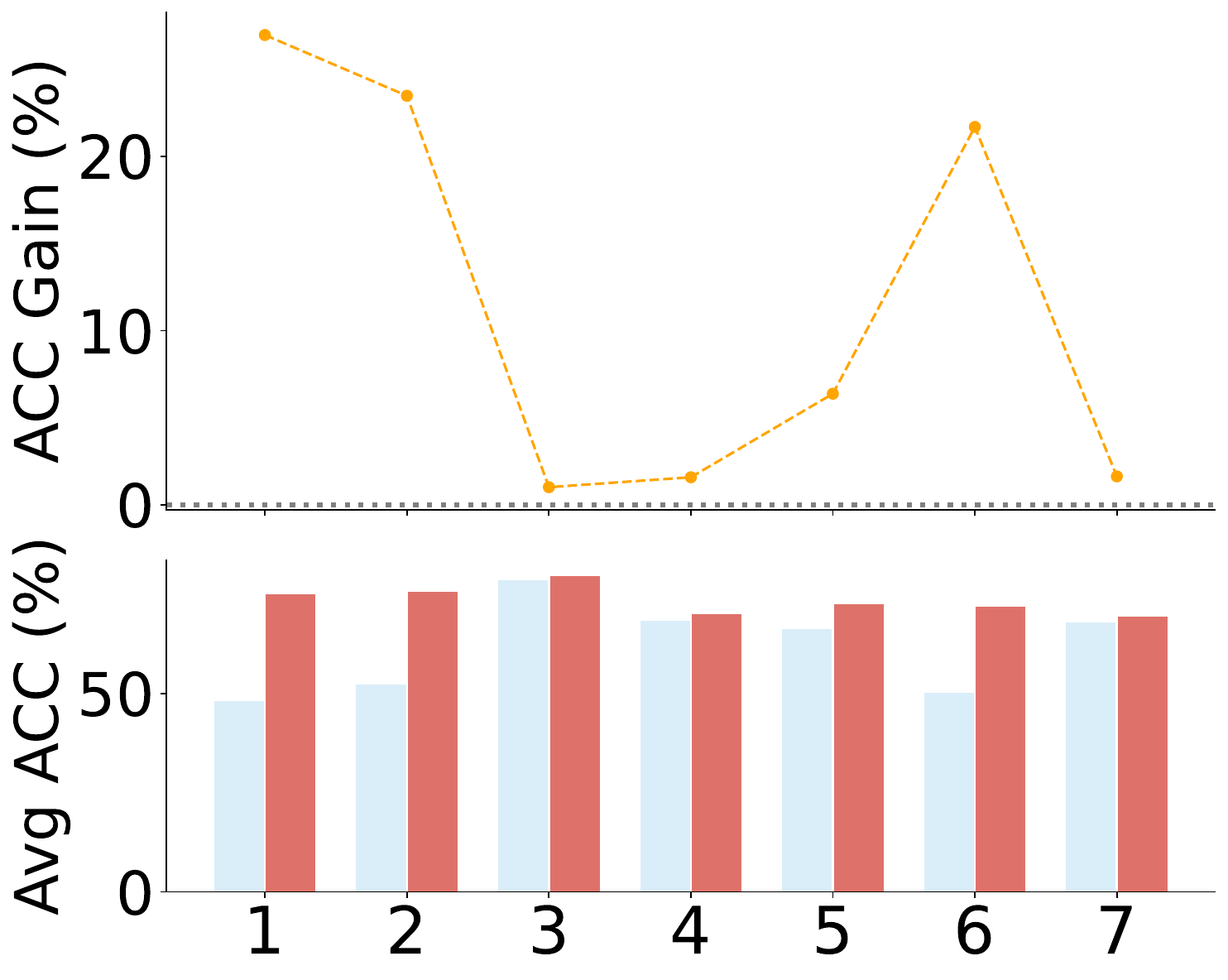}
  \caption{Fashion}
\end{subfigure}\hfill
\begin{subfigure}[b]{0.16\textwidth}
  \includegraphics[width=\textwidth]{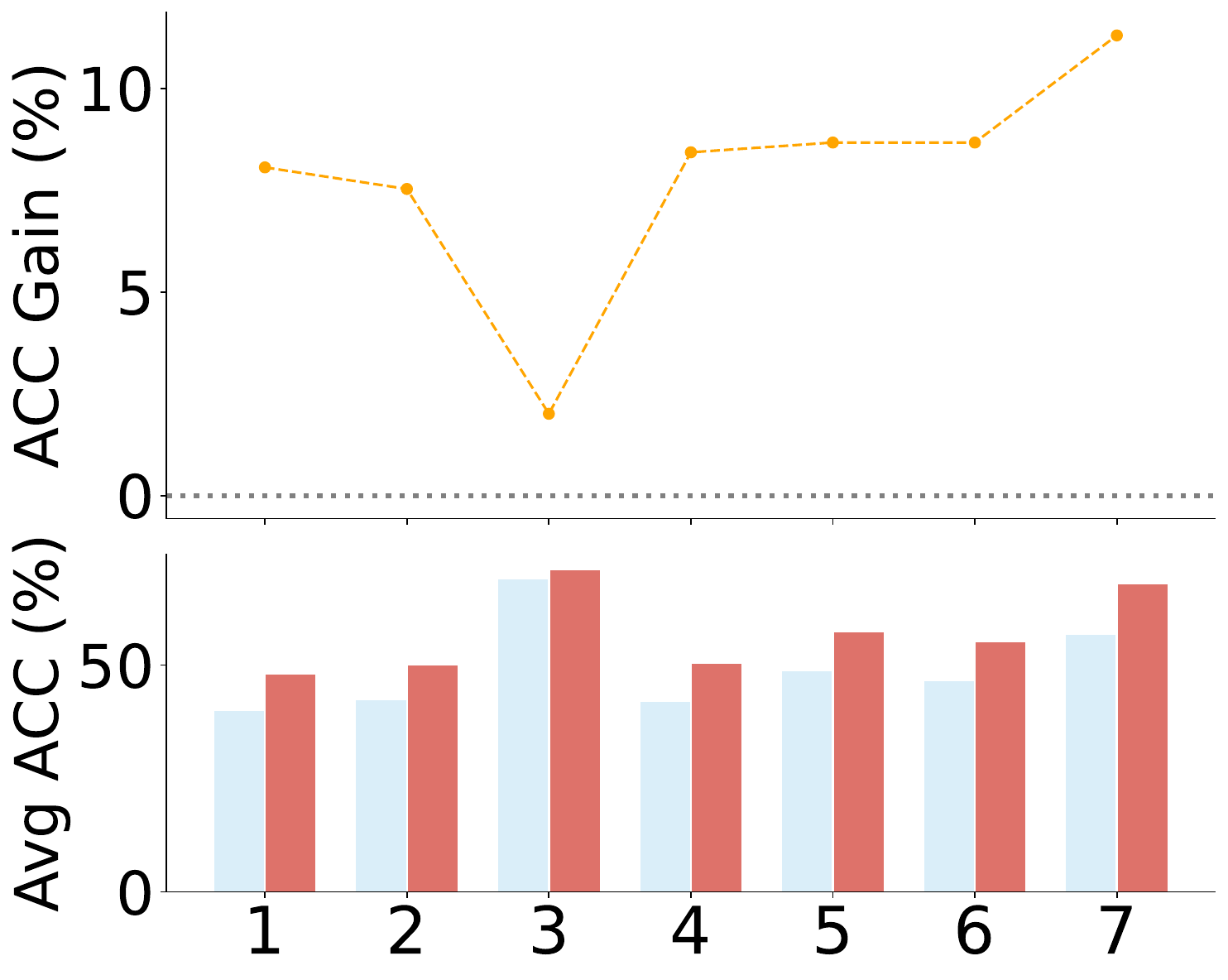}
  \caption{ImageNet-1K}
\end{subfigure}

\begin{subfigure}[b]{\textwidth}
  \centering
  \includegraphics[width=0.35\textwidth]{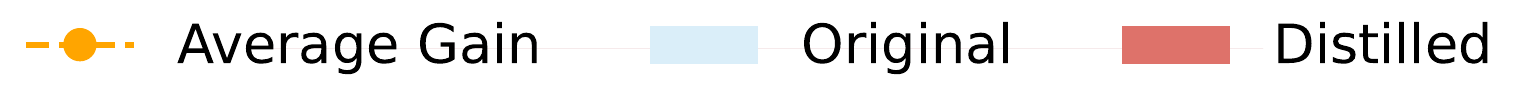} %
\end{subfigure}
\caption{Clustering accuracy of FSRC across 18 datasets using 7 backbone models. Original denotes the original global semantic embeddings, and Distilled denotes the Distilled Representation, which is obtained via mutual distillation between global semantic embeddings and local relational embeddings.}
\label{fig11}
\end{figure*}

Figure~\ref{fig11} presents the clustering accuracy across all 18 datasets and seven foundation models, comparing the original features against the FSRC-distilled features.

Overall, FSRC consistently achieves higher clustering accuracy than the original features across most datasets. The performance improvement of FSRC varies with both the dataset and the backbone. On fine-grained datasets, local relational features substantially enhance discrimination of subtle inter-class differences. For complex, large-scale datasets, the benefit depends on the backbone: fusion refines already strong global features while compensating for weaker or misaligned ones with complementary local information.

\clearpage

\section{Visualization Comparison with Representative Clustering Methods}
\label{I}
\begin{figure*}[t]
  \centering
  \begin{tabular}{ccc}
    \textbf{ACC = 64.5\%} & \textbf{ACC = 82.9\%} & \textbf{ACC = 96.3\%} \\
    \includegraphics[width=0.3\linewidth]{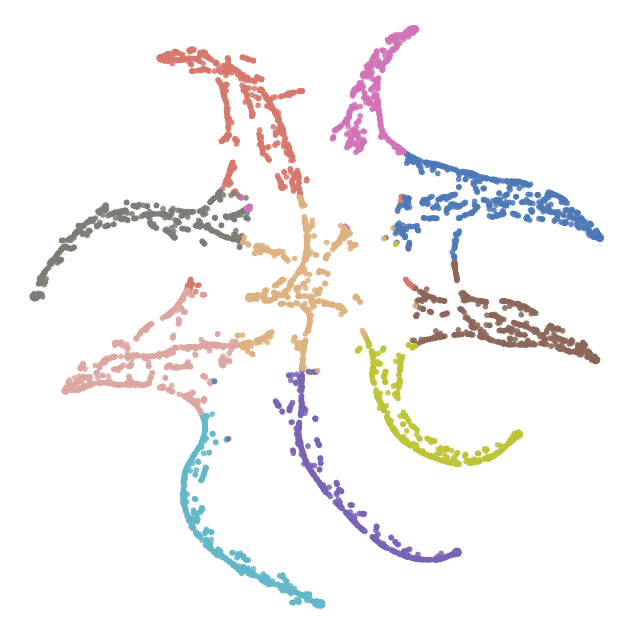} &
    \includegraphics[width=0.3\linewidth]{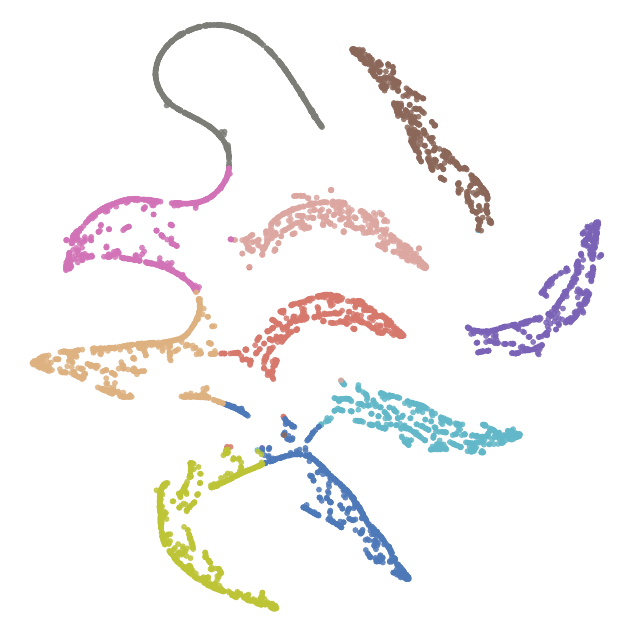} &
    \includegraphics[width=0.3\linewidth]{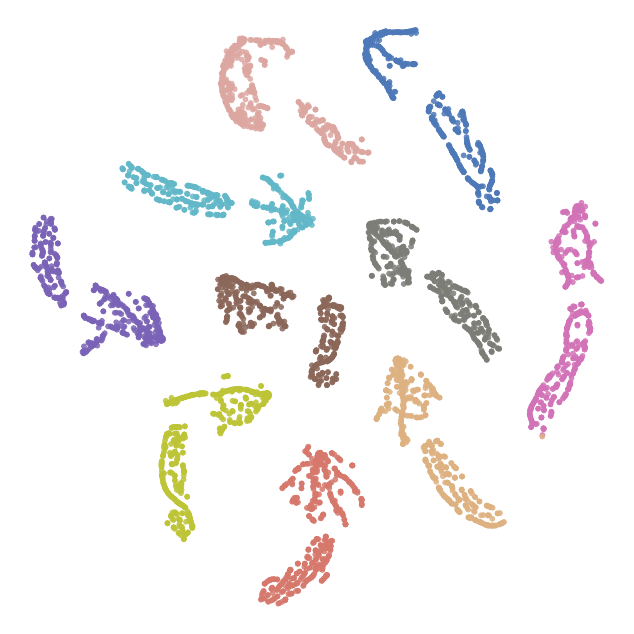} \\
    A. TAC & B. TURTLE & C. FSRC
  \end{tabular}
  \caption{t-SNE visualizations of the EuroSAT dataset learned by TAC, TURTLE, and the proposed FSRC.}
  \label{fig12}
\end{figure*}
To intuitively evaluate the relative advantages of FSRC over competing methods, we further select the representative foundation-model-based clustering methods in Table~1, namely TAC and TURTLE, and conduct a t-SNE visualization study \cite{maaten2008visualizing} on the EuroSAT dataset. 

As shown in Figure~\ref{fig12}, TAC produces relatively scattered clusters with considerable overlap among different categories. TURTLE improves the compactness of several clusters. However, some cluster boundaries remain blurred, and partial overlap can still be observed. FSRC produces substantially more compact and well-separated clusters, exhibiting tighter intra-class distributions and larger inter-class margins than both TAC and TURTLE. FSRC learns more clustering-friendly representations that better capture the intrinsic structure of the dataset. These visual observations are consistent with the quantitative results, where FSRC consistently outperforms TAC and TURTLE.

\section{Training Objective Convergence Analysis}
\label{J}
\begin{figure}[!t]
  \centering
  \subfloat[ImageNet-10\label{fig7a}]{
    \includegraphics[width=0.33\columnwidth]{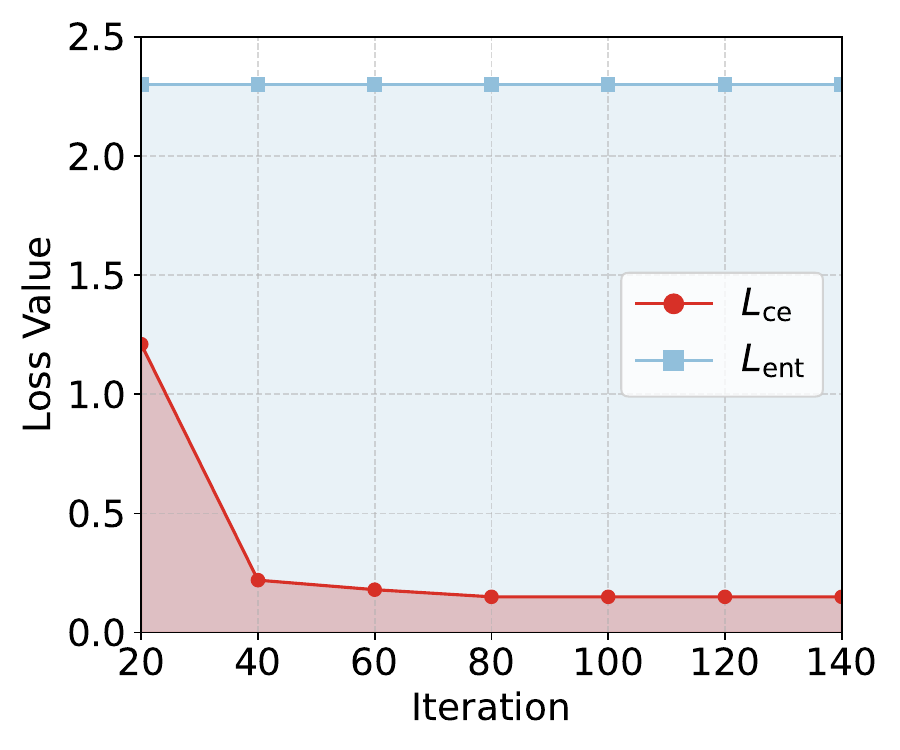}
  }\hfil
  \subfloat[CIFAR-10\label{fig7b}]{
    \includegraphics[width=0.33\columnwidth]{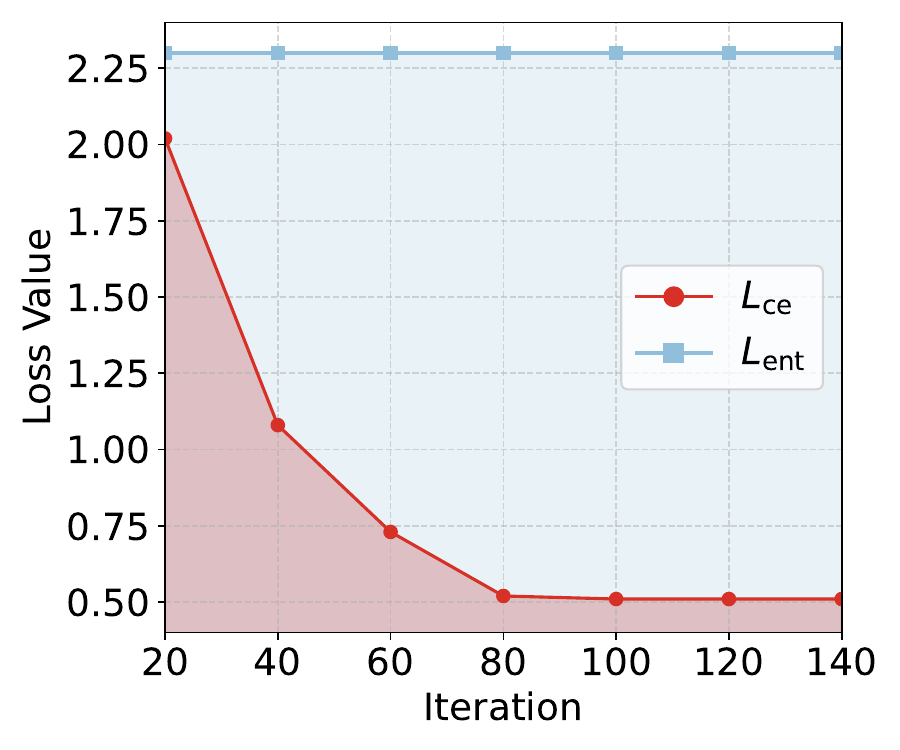}
  }
  \caption{Convergence of the $\mathcal{L}_{\mathrm{ce}}$ and $\mathcal{L}_{\mathrm{ent}}$ on (a) ImageNet-10 and (b) CIFAR-10 datasets.}
  \label{fig13}
\end{figure}
To better understand FSRC's optimization properties and practical reliability, we analyze the convergence of its two loss components, $\mathcal{L}_{\mathrm{ce}}$ and $\mathcal{L}_{\mathrm{ent}}$. We record their values during training and visualize the corresponding trajectories on CIFAR-10 and ImageNet-10. 

As shown in Fig.~\ref{fig13}, $\mathcal{L}_{\mathrm{ce}}$ decreases rapidly in the early stage and then converges to a stable level on both datasets, indicating that FSRC quickly learns discriminative class-specific representations and maintains stable optimization throughout training.
Meanwhile, $\mathcal{L}_{\mathrm{ent}}$ remains relatively stable throughout training, while encouraging balanced cluster assignments and helping prevent clustering collapse. Overall, the consistent behavior of these two objectives demonstrates that FSRC achieves stable convergence and reliable optimization dynamics across different datasets.

\section{Parameter Analysis}\label{K0}

\begin{figure}[t]
  \centering
  \includegraphics[width=0.66\linewidth]{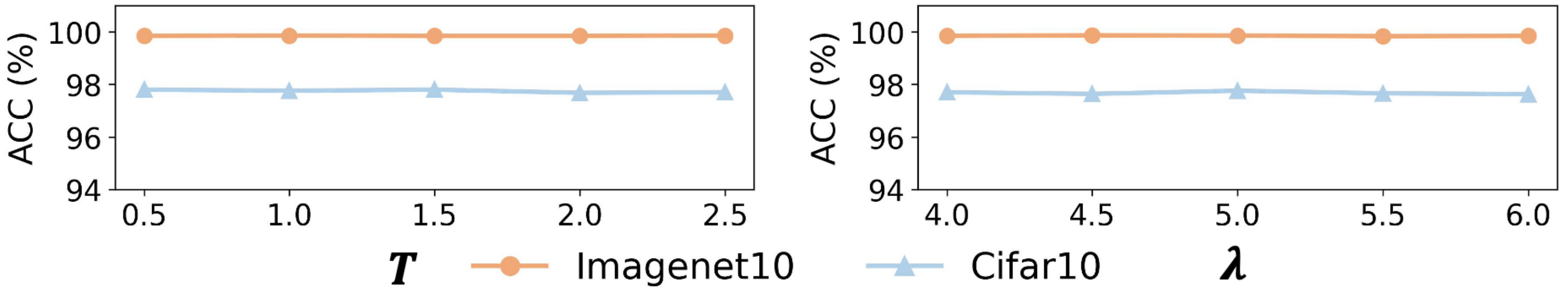}
  \caption{
Parameter analysis on  CIFAR-10 and ImageNet-10 datasets.
}
  \label{fig9}
\end{figure}

We conduct a parameter analysis of FSRC using the CLIP ViT-L/14 backbone on ImageNet-10 and CIFAR-10 datasets. We systematically vary the entropy regularization weight \(\lambda\) and the distillation temperature \(T\), while keeping all other hyperparameters fixed. As shown in Figure~\ref{fig9}, ACC remains highly stable across a wide range of parameter values. These results demonstrate that FSRC is robust to variations in hyperparameters and does not require extensive fine-tuning.

\section{Additional Semantic Compactness Results}
\label{K}
\begin{table}[t]
\centering
\caption{Silhouette Coefficient(SC) across different backbone architectures.  \underline{Underlined} numbers indicate the higher value within each Original/Distilled pair for the same architecture. Original denotes the original global semantic embeddings, and Distilled denotes the Distilled Representation, which is obtained via mutual distillation between global semantic embeddings and local relational embeddings.}
\renewcommand{\arraystretch}{1.5}
\resizebox{\textwidth}{!}{
\begin{tabular}{c|cc|cc|cc|cc|cc|cc|cc}
\toprule
 & \multicolumn{2}{c}{ClipResNet50} & \multicolumn{2}{c}{ClipResNet101} & \multicolumn{2}{c}{ClipResNet50*4} & \multicolumn{2}{c}{ClipVITB/32} & \multicolumn{2}{c}{ClipVITB/16} & \multicolumn{2}{c}{DINOv2} & \multicolumn{2}{c}{ClipVITL/14} \\
\cmidrule(lr){2-3} \cmidrule(lr){4-5} \cmidrule(lr){6-7} \cmidrule(lr){8-9}\cmidrule(lr){10-11}\cmidrule(lr){12-13}\cmidrule(lr){14-15}
  & Original \hspace{0.5pt}& Distilled \hspace{1pt}& Original \hspace{0.5pt}& Distilled \hspace{1pt}& Original \hspace{0.5pt}& Distilled \hspace{1pt}& Original \hspace{0.5pt}& Distilled \hspace{1pt}& Original \hspace{0.5pt}& Distilled \hspace{1pt}& Original \hspace{0.5pt}& Distilled \hspace{1pt}& Original \hspace{0.5pt}& Distilled\\

\midrule
MNIST        & 0.11 & \underline{0.43} & 0.11 & \underline{0.38} & 0.09 & \underline{0.41} & 0.12 & \underline{0.44} & 0.11 & \underline{0.47} & 0.09 & \underline{0.38} & 0.19 & \underline{0.49} \\
GTSRB        & -0.04 & \underline{0.03} & -0.02 & \underline{0.03} & -0.03 & \underline{0.04} & -0.01 & \underline{0.07} & 0.00 & \underline{0.07} & -0.01 & \underline{0.03}& 0.10 & \underline{0.15} \\
ImageNet-10  & 0.36 & \underline{0.83} & 0.35 & \underline{0.89} & 0.35 & \underline{0.89} & 0.35 & \underline{0.90} & 0.36 & \underline{0.91} & 0.54 & \underline{0.84}  & 0.34 & \underline{0.90} \\
OxfordPets   & 0.05 & \underline{0.32} & 0.09 & \underline{0.40} & 0.10 & \underline{0.42} & 0.07 & \underline{0.39} & -0.05 & \underline{0.01} & 0.43 & \underline{0.64}& 0.14 & \underline{0.52} \\
ImageNet-Dogs& 0.03 & \underline{0.30} & 0.06 & \underline{0.39} & 0.07 & \underline{0.41} & 0.05 & \underline{0.40} & 0.07 & \underline{0.45} & 0.46 & \underline{0.78}  & 0.10 & \underline{0.53} \\
DTD          & 0.02 & \underline{0.09} & 0.04 & \underline{0.09} & 0.04 & \underline{0.13} & 0.04 & \underline{0.10} & 0.05 & \underline{0.13} & 0.13 & \underline{0.19} & 0.07 & \underline{0.16} \\
CIFAR-10     & 0.07 & \underline{0.24} & 0.11 & \underline{0.33} & 0.11 & \underline{0.33} & 0.16 & \underline{0.45} & 0.18 & \underline{0.49} & 0.20 & \underline{0.62} & 0.16 & \underline{0.48} \\
CIFAR-100    & -0.02 & \underline{0.00} & 0.01 & \underline{0.04} & 0.00 & \underline{0.03} & 0.04 & \underline{0.10} & 0.05 & \underline{0.13} & 0.27 & \underline{0.41}  & 0.05 & \underline{0.17} \\
TinyImageNet & -0.03 & \underline{0.00} & \underline{0.00} & -0.01 & \underline{0.00} & -0.02 & 0.02 & \underline{0.05} & 0.03 & \underline{0.06} & \underline{0.32} & 0.24 & 0.08 & \underline{0.10} \\
Caltech101   & 0.17 & \underline{0.27} & 0.21 & \underline{0.30} & 0.22 & \underline{0.35} & 0.22 & \underline{0.34} & 0.24 & \underline{0.37} & 0.55 & \underline{0.65} & 0.26 & \underline{0.43} \\
FER2013      & -0.03 & \underline{0.00} & -0.04 & \underline{-0.01} & -0.03 & \underline{-0.01} & -0.03 & \underline{-0.01} & -0.03 & \underline{0.00} & -0.01 & \underline{0.00} & -0.03 & \underline{0.00} \\
EuroSAT      & 0.12 & \underline{0.36} & 0.11 & \underline{0.52} & 0.14 & \underline{0.46} & 0.09 & \underline{0.43} & 0.13 & \underline{0.56} & 0.12 & \underline{0.56} & 0.16 & \underline{0.62} \\
RESISC45     & 0.09 & \underline{0.24} & 0.12 & \underline{0.28} & 0.13 & \underline{0.31} & 0.12 & \underline{0.29} & 0.16 & \underline{0.34} & 0.12 & \underline{0.34}  & 0.17 & \underline{0.37} \\
UCF101       & 0.10 & \underline{0.19} & 0.13 & \underline{0.23} & 0.14 & \underline{0.27} & 0.15 & \underline{0.25} & 0.16 & \underline{0.28} & 0.32 & \underline{0.45}  & 0.20 & \underline{0.37} \\
CLEVR        & -0.04 & \underline{-0.04} & \underline{-0.03} & -0.04 & -0.04 & \underline{-0.04} & \underline{-0.03} & -0.04 & -0.04 & \underline{-0.04} & -0.08 & \underline{-0.05} & -0.05 & \underline{-0.04} \\
USPS         & 0.21 & \underline{0.50} & 0.19 & \underline{0.49} & 0.21 & \underline{0.50} & 0.23 & \underline{0.56} & 0.30 & \underline{0.56} & 0.16 & \underline{0.46} & 0.32 & \underline{0.58} \\
Fashion      & 0.09 & \underline{0.32} & 0.13 & \underline{0.33} & 0.11 & \underline{0.30} & 0.15 & \underline{0.33} & 0.14 & \underline{0.35} & 0.15 & \underline{0.43}  & 0.12 & \underline{0.32} \\
ImageNet-1K  & 0.00 & \underline{0.03} & 0.02 & \underline{0.04} & 0.03 & \underline{0.06} & 0.02 & \underline{0.04} & 0.04 & \underline{0.07} & \underline{0.31} & 0.28 & 0.08 & \underline{0.10} \\
\hline
\textbf{AVG.} & 0.07 & \underline{0.23} & 0.09 & \underline{0.26} & 0.09 & \underline{0.27} & 0.10 & \underline{0.28} & 0.11 & \underline{0.29} & 0.23 & \underline{0.40} & 0.14 & \underline{0.34} \\
\bottomrule
\end{tabular}}
\label{tab8}
\end{table}

To further evaluate the clustering friendliness of the distilled features, we conduct additional experiments using a label-free metric, the Silhouette Coefficient (SC), across seven different backbone architectures. In this experiment, SC is computed using cosine distance and the cluster assignments predicted by K-Means on the original image features, without relying on ground-truth labels. A higher SC indicates that samples within the same predicted cluster are more compact and better separated from samples in other clusters, reflecting a stronger clustering structure in the embedding space.

As shown in Table~\ref {tab8}, the distilled representations generally achieve higher SC values than the original embeddings across most datasets and backbones, demonstrating that local relations improve cluster compactness and separability. The improvement is especially evident on datasets such as MNIST, ImageNet-10, ImageNet-Dogs, CIFAR-10, and EuroSAT. For more challenging datasets such as FER2013 and CLEVR, the SC values remain close to zero or slightly negative, suggesting that class structure is harder to form in these settings. Overall, these results confirm that the distilled features are more clustering-friendly and better organized than the original feature space.

\section{Effectiveness Analysis of Local Relational Graph Construction}
\label{N}

To evaluate the effectiveness of local relational graph construction, we compare our spatial 8-neighborhood graph with a fully connected non-local graph while keeping the remaining FSRC components unchanged. The local relational graph preserves spatial connectivity while reducing the number of edges by approximately 7.5 times.

As shown in Table~\ref{tab:graph_ablation}, the local relational graph consistently outperforms the non-local alternative across all three clustering metrics, improving ACC by 1.3 and 1.1 points on OxfordPets and ImageNet-Dogs, respectively. These results demonstrate that local relational graph construction reduces relational modeling overhead while better capturing local relational information beneficial for clustering.

\begin{table}[t]
\centering
\caption{Comparison of local relational and non-local graphs on OxfordPets and ImageNet-Dogs.}
\label{tab:graph_ablation}
\renewcommand{\arraystretch}{1.15}
\setlength{\tabcolsep}{7pt}
\begin{tabular}{llccc}
\toprule
Dataset & Graph & ACC & NMI & ARI \\
\midrule
\multirow{2}{*}{OxfordPets}
& Non-local & 91.2 & 91.9 & 84.8 \\
& Local Relational (Ours) & \textbf{92.5} & \textbf{92.9} & \textbf{86.9} \\
\midrule
\multirow{2}{*}{ImageNet-Dogs}
& Non-local & 90.5 & 89.7 & 88.0 \\
& Local Relational (Ours) & \textbf{91.6} & \textbf{90.6} & \textbf{88.8} \\
\bottomrule
\end{tabular}
\end{table}

\section{Statistical Stability Analysis}
\label{M}

To evaluate the statistical stability of FSRC, we conduct 10 independent runs with different random seeds on three representative datasets: ImageNet-10, ImageNet-Dogs, and CIFAR-10. As shown in Table~\ref{tab:statistical_stability}, FSRC exhibits highly consistent performance on ImageNet-10 and CIFAR-10, with negligible variations across different random seeds. On ImageNet-Dogs, the results show greater variability, reflecting the increased difficulty of clustering fine-grained visual categories. Overall, these results demonstrate that FSRC maintains stable clustering performance across different random seeds.

\begin{table}[t]
\centering
\caption{Statistical stability of FSRC across 10 random seeds. Results are reported as mean $\pm$ standard deviation (\%).}
\label{tab:statistical_stability}
\renewcommand{\arraystretch}{1.15}
\setlength{\tabcolsep}{5pt}
\begin{tabular}{lccc}
\toprule
Dataset & ACC & NMI & ARI \\
\midrule
ImageNet-10
& $99.9 \pm 0.00$
& $99.5 \pm 0.00$
& $99.7 \pm 0.00$ \\

ImageNet-Dogs
& $90.0 \pm 3.43$
& $88.6 \pm 2.11$
& $84.9 \pm 3.83$ \\

CIFAR-10
& $97.8 \pm 0.05$
& $93.9 \pm 0.09$
& $94.3 \pm 0.11$ \\
\bottomrule
\end{tabular}
\end{table}

\section{Computational Cost Analysis}
\label{O}

We further evaluate the computational scalability of FSRC, using direct clustering on CLIP ViT-L/14 features as the single-backbone baseline. Extracting global semantic features requires 155.60 GFLOPs per image, while the additional local relation encoder introduces only 16.71 GFLOPs per image. On ImageNet-10, the total feature-extraction cost increases from 2.02 to 2.20 PFLOPs, while on CIFAR-10, it increases from 9.34 to 10.17 PFLOPs. The peak GPU memory usage increases only from 3212 MB to approximately 3416 MB. Regarding inference efficiency, the complete inference times of FSRC and the single-backbone baseline are 223.92 s and 216.29 s on ImageNet-10, respectively, and 174.25 s and 157.33 s on CIFAR-10, respectively. Overall, FSRC introduces only modest additional computational, memory, and inference-time overhead despite incorporating the local relation encoder.

\end{document}